\documentclass{l4dc2024}

\usepackage{hyperref, nicefrac}
\usepackage{booktabs, subcaption, comment, nicefrac}

\newcommand{\algo}{{\small \sc\textsf{SynCMA}}}
\newcommand{\framework}{{\small \sc\textsf{InvIGO}}}

\title[An Invariant Information Geometric Method]{An Invariant Information Geometric Method for High-Dimensional Online Optimization}
\usepackage{times}

 \coltauthor{\Name{Zhengfei Zhang} \Email{zf-zhang20@mails.tsinghua.edu.cn}\\
  \Name{Yunyue Wei} \Email{weiyy20@mails.tsinghua.edu.cn}\\
  \Name{Yanan Sui} \Email{ysui@tsinghua.edu.cn}\\
  \addr Tsinghua University, Beijing, China}

\begin{document}

 \RestyleAlgo{ruled}
\maketitle

\newtheorem{assumption}{Assumption}

\begin{abstract}%
    Sample efficiency is crucial in optimization, particularly in black-box scenarios characterized by expensive evaluations and zeroth-order feedback. When computing resources are plentiful, Bayesian optimization is often favored over evolution strategies. In this paper, we introduce a full invariance oriented evolution strategies algorithm, derived from its corresponding framework, that effectively rivals the leading Bayesian optimization method in tasks with dimensions at the upper limit of Bayesian capability.
    Specifically, we first build the framework~\framework~that fully incorporates historical information while retaining the full invariant and computational complexity. 
    We then exemplify~\framework~on multi-dimensional Gaussian, which gives an invariant and scalable optimizer~\algo~. The theoretical behavior and advantages of our algorithm over other Gaussian-based evolution strategies are further analyzed. 
    Finally, We benchmark~\algo~against leading algorithms in Bayesian optimization and evolution strategies on various high dimension tasks, including Mujoco locomotion tasks, rover planning task and synthetic functions. In all scenarios, \algo~demonstrates great competence, if not dominance, over other algorithms in sample efficiency, showing the underdeveloped potential of property oriented evolution strategies.
\end{abstract}


\section{Introduction}

Without access to gradient information, many real-world continuous-space optimization problems rely on zeroth-order evaluations. These function evaluations are often costly and become less useful over time as the environment changes. Therefore, such tasks are better approached as online optimization problems with zeroth-order feedback and an ignorant initial. An ideal optimizer should have high sample efficiency with reasonable computational complexity.

In this sense, Bayesian optimization (BO) is often the favored choice because it has empirically better sample efficiency in various machine learning scenarios\citep{frazier2018tutorial, shahriari2015taking}. 
However, this success is limited to low-dimensional problems due to the cubic computational complexity of its surrogate model\citep{rasmussen2003gaussian, eriksson2019scalable}.
Researchers developed versatile scalable variants of Bayesian optimization\citep{binois2022survey}, extending the dominance dimension of Bayesian optimization up to hundreds. In this paper, we use the name of high dimension to denote dimensions ranging from dozens to hundreds, and the name of online optimization to include both Bayesian optimization and evolution strategy.

Evolution strategy has computational complexity that is independent of sample size, which makes it a general method to apply. Over years' development, some theoretical frameworks and guidance are developed\citep{akimoto2014comparison, akimoto2012theoretical}, through which covariance matrix adaptation evolution strategies(CMA-ES) \citep{hansen2016cma} and its variants\citep{akimoto2020diagonal, abdolmaleki2017deriving} stand out. They are the current leading family of algorithms and reach a balance between sample efficiency and computational cost. However, the lack of a solid theoretical foundation greatly hinders their development despite of many efforts invested\citep{arnold2010active,brockhoff2012effect,shirakawa2018sample,nishida2018psa,ba2016distributed,akimoto2016online}. The potential of CMA family and even evolution strategies seems to be far from being explored. Instead of developing a solid theory, we want to explore this potential from a property oriented perspective. In specific, we wonder, \textit{can a fully invariant oriented evolution strategies algorithm have great competence against Bayesian optimization in high dimensional tasks?}

The invariant orientation is motivated by general optimization problems where gradient information is available. In this scenario, it is widely known that the performance of leading first-order optimizers such as AdaGrad \citep{duchi2011adaptive} and Adam \citep{kingma2014adam} are highly dependent on the curvature of the optimization objective. Since the curvature depends on the parameterization of the model, parameterization invariant optimizers are thus considered as promising ways. Natural gradient \citep{amari1998natural} and further efforts \citep{song2018accelerating, transtrum2012geodesic} towards practical invariant methods concentrate on exploiting first or higher order structure in parameter space to accelerate or strengthen invariant for optimizers. However, only limited progress has been made. When gradient information is not available, sampling is used in Information Geometric Optimization(IGO)\citep{ollivier2017information} to estimate the natural gradient for specific parametric distributions. Similarly, geodesic modification is also explored \citep{bensadon2015black} but the practical invariant capability is limited as in the general case. 

\paragraph{Our contribution.} We build the first invariant optimizer framework~\framework~for online optimization with ignorant initial and zeroth-order feedback. Which adopts an approximation to the objective in IGO to allow everywhere differentiability, and a line search strategy to completely and scalably incorporate historical information. When further exemplified with multi-dimensional Gaussian as in CMA family, the derived practical optimizer~\algo~inherits all properties of~\framework~and has fewer hyperparameters comparing with other CMA optimizers. It is also the first time that historical information is stably incorporated for both mean and covariance parameters. In experiments that benchmark on high dimensional realistic tasks, that Bayesian optimizers usually dominant, and synthetic tasks,~\algo~demonstrates great competence over other optimizers in sample efficiency. 

\section{An Invariant Optimizer Family with an Approximate Objective}

Considering the online optimization problem where a black-box function $f$ needs to be optimized, and the optimizer is initially ignorant with only zeroth-order feedback available. \begin{equation}
    x^* = \mathrm{argmin}_{x \in \mathbb{R}^n}~f(x)
\end{equation}
A global parametric sample distribution $\theta\mapsto p_{\theta}$ is often used to relax the original optimization problem into an optimization problem on the parameter space $\Theta$, with a substitutional fitness function $g_{f, \theta}(x)$ to represent how good a sample $x \in \mathbb{R}^n$ is.  \begin{equation}
   \theta^* = \mathrm{argmin}_{\theta}~\mathbb{E}_{p_{\theta}}[g_{f, \theta}(x)] \label{utmost_question}
\end{equation} 
The expression of $g$ is determined manually from the set of integrable functions depending on original objective $f$ and the current point $\theta \in \Theta$. The substitution here actually generalizes the problem as specifically, (1) if $f$ has good properties, then it is naturally to set $g_{f, \theta} = f$; (2) if $g$ is related with $\theta$, then the problem fits the time-varying environment setting that many online optimizers pursue.

\subsection{Natural Gradient Flow with Zeroth-order Feedback}

To solve equation (\ref{utmost_question}) on the parameter space $\Theta$, a natural gradient flow is often the primary choice. \begin{equation}
    \frac{\mathrm{d}\theta}{\mathrm{d}t} = -  \tilde{\triangledown}_{\theta} \mathbb{E}_{p_{\theta}}[g_{f, \theta^t}(x)] 
\end{equation} Here $\theta^t$ denotes the current $\theta$ with static nature that gradient should not apply on. The usage of natural gradient keep this ODE invariant under smooth bijective transformation of parameter space. Its vanilla discrete version, i.e. natural gradient descent algorithm goes to, \begin{equation}
    \theta^{t + 1} = \theta^t - h \tilde\triangledown_{\theta}|_{\theta = \theta^t} L_{\theta^t}(\theta)\label{ngd_alg}
\end{equation} 
Here $h$ denotes the learning rate, and the loss function is defined as $L_{\theta^t}(\theta) \equiv \mathbb{E}_{p_{\theta}}[g_{f, \theta^t}(x)]$. This definition of loss function can be easily extended to contain the common definition of loss function in deep learning by extending the sample distribution $p_{\theta}$ to include neural network, although in this paper we focus on parametric distribution families.

To practically compute the natural gradient in black-box setting, IGO set a sampling method for certain distribution families, \begin{align}
    \tilde{\triangledown}_{\theta}|_{\theta = \theta^t} L_{\theta^t}(\theta)
    =&~ I^{-1}(\theta^t) \int g_{f, \theta^t}(x) \frac{\partial \ln{p_{\theta}(x)}}{\partial \theta}|_{\theta = \theta^t} p_{\theta^t}(\mathrm{d}x) \label{eq_complexity_definition}
\end{align}  
In IGO, the natural gradient is only available at point $\theta^t$, different choices of distribution family $\Theta$ provide different optimization methods in the form of (\ref{ngd_alg}). We thus define the IGO complexity to measure the computational complexity of other natural gradient based optimizers.

\begin{assumption}
    For a given $x \in \mathbb{R}^n$ and $\theta \in \Theta$, $I^{-1}(\theta) \frac{\partial \ln{p_{\theta}(x)}}{\partial \theta}$ cost finite $\mathcal{O}(H)$ time to compute. 
    \label{assum_availability}
\end{assumption}

\begin{definition}[IGO complexity]
    When assumption \ref{assum_availability}~holds, the IGO complexity $\mathcal{O}(HN)$ is the computational complexity for single step updates when applying IGO to natural gradient method, i.e. to compute equation (\ref{eq_complexity_definition}) with $N$ samples. 
\label{igo_complexity}
\end{definition}

When discretizing with a given learning rate, errors with respect to the invariant property occur, which may accumulate to drastically change the trajectory. In gradient accessible setting with general lost function, the best invariant error achieved \citep{song2018accelerating} is $2$-nd order invariant, representing the decrease of the error between the optimizer and some exactly invariant trajectories is $\mathcal{O}(h^2)$. Similar error order is achieved in the content of black-box setting for certain parametric distribution \citep{bensadon2015black}. There are some attempts to illustrate a fully invariant optimizer, such as IGO-ML in \citep{ollivier2017information}, but they are not practically invented. In short, there is no practical algorithm that is fully invariant or even have a better error order.

\begin{definition}[Invariant property]
    Let $\theta$ be the parameter of an optimizer using model $p_{\theta}$ and $\varphi(\theta)$ be an smooth bijective transformation of $\theta$ of the same optimizer using model $p'_{\varphi(\theta)} = p_{\theta}$. Let $\theta^t$ be the optimization trajectory when optimizing objective $f$, parameterized by $\theta$ and initialized at $\theta^0$. And $\varphi^t$ the optimization trajectory when optimizing objective $f$, parameterized by $\varphi$ and initialized at $\varphi^0 = \varphi(\theta^0)$. We claim that the optimizer is invariant if $\forall t \in \mathbb{N}, \varphi^t = \varphi(\theta^t)$.
\end{definition}

\subsection{Optimizing with the Approximate Objective}

Given that $g_{f, \theta^t}$ is manually selected and $\forall b \in \mathbb{R}, \triangledown_{\theta} E_{p_{\theta}}[g_{f, \theta^t}(x)] = \triangledown_{\theta} E_{p_{\theta}}[g_{f, \theta^t}(x) + b]$, we assume $g_{f, \theta^t}$ to be non-negative without loss of generality. Let reweighted distribution $q_{\theta}(x) \equiv \frac{p_{\theta}(x)g_{f, \theta^t}(x)}{L_{\theta^t}(\theta)}$, then we can decompose $\log L_{\theta^t}(\theta)$ as follow, \begin{align}
    \log{\frac{L_{\theta^t}(\theta)}{L_{\theta^t}(\theta^t)}} = D_{KL}(q_{\theta^t}||q_{\theta}) + D_{KL}(q_{\theta^t}||p_{\theta^t}) - D_{KL}(q_{\theta^t}||p_{\theta}) \label{decomposition}
\end{align}
Inspired from this decomposition, we claim $D_{KL}(q_{\theta^t}||p_{\theta})$ a good objective approximating $L_{\theta^t}(\theta)$. All the proofs and detailed derivations are set in Appendix\footnote{Please refer to https://github.com/Anoxxx/SynCMA-official for appendix and source codes.} \ref{proof_derivation}.

\begin{proposition}
    The KL-divergence $D_{KL}(q_{\theta^t}||p_{\theta})$ is a substitution for $L_{\theta^t}(\theta)$ with the following properties. \begin{enumerate}
        \item The (natural) gradients for $\log L_{\theta^t}(\theta)$ and $ -D_{KL}(q_{\theta^t}||p_{\theta})$ coincide at current point $\theta^t$, further $\forall \theta \equiv (\theta^t + \delta \theta) \in \Theta$, $\triangledown_{\theta} \log L_{\theta^t}(\theta) = -\triangledown_{\theta}D_{KL}(q_{\theta^t}||p_{\theta}) + O(\delta \theta)$.
        \item Under Assumption~\ref{assum_availability}~, computing natural gradient of $D_{KL}(q_{\theta^t}||p_{\theta})$ at any point $\theta \in \Theta$ costs the IGO complexity $O(HN)$. While objective $L_{\theta^t}(\theta)$ in IGO is only available to be differentiated at point $\theta^t$.
    \end{enumerate}
    \label{thm_approx}
\end{proposition}

Combining with the above two properties of $D_{KL}(q_{\theta^t}||p_{\theta})$, it is natural to consider a step size constraint update for $\theta^{t + 1}$ when optimizing $D_{KL}(q_{\theta^t}||p_{\theta})$. The specific choice of the step size constraint comes from the definition of natural gradient,\begin{align}
    \tilde{\triangledown}|_{\theta = \theta^t} L_{\theta^t}(\theta) \propto \lim_{\epsilon \to 0^{+}} \frac{1}{\epsilon} \mathrm{argmax}_{\delta \theta ~s.t.~ D_{KL}(p_{\theta^t}||p_{\theta^t + \delta \theta}) \le \nicefrac{\epsilon^2}{2}} D_{KL}(q_{\theta^t}||p_{\theta^t + \delta \theta}) \label{def_ng}
\end{align}

Now our optimization problem for each time step is,\begin{align}
    \theta^{t + 1}_* =~ \mathrm{argmax}_{\theta} D_{KL}(q_{\theta^t}||p_{\theta}) \label{problem} ~~s.t.~D_{KL}(p_{\theta^t}||p_{\theta}) \le \nicefrac{\epsilon^2}{2} 
\end{align}

It is notable that the optimization problem (\ref{problem}) is approximately solving $\frac{\mathrm{d}\theta}{\mathrm{d}t} = -s(\theta) \tilde{\triangledown} \log L_{\theta^t}(\theta)$.
Here $s(\theta) \equiv \frac{1}{\epsilon} ||\tilde{\triangledown}|_{\theta = \theta^t} D_{KL}(q_{\theta^t}||p_{\theta}) ||$ corresponds to the implicit learning rate of~\framework. As there is no explicit learning rate, this implicit learning rate hints an overall learning rate that is proportion to $D_{KL}(p_{\theta^t}||p_{\theta})^{-0.5}$. The efficiency of such dependency of the overall learning rate with the KL-divergence between adjacent distributions is widely verified in natural gradient based optimization methods such as K-FAC~\citep{ba2016distributed} and reinforcement learning algorithms such as ACKTR~\citep{wu2017scalable}. Our formulation is justified thereby.
\subsection{Invariantly Incorporating Historical Information}

When only local information is used in each iteration, historical information is helpful for the optimization, even if the environment is constantly changing over time. We thus modify objective $D_{KL}(q_{\theta^t}||p_{\theta})$ to incorporate historical information. Here $T$ denotes the horizon and the widely used exponential decay is applied with decay parameter $\lambda \in [0, 1)$, $\lambda^0$ is regarded as $1$ in default. \begin{align}
    \theta^{t + 1}_* =~ \mathrm{argmax}_{\theta} \sum_{\tau = 0}^{T} \lambda^{\tau} D_{KL}(q_{\theta^{t - \tau}}||p_{\theta}) \label{historical_problem}
    ~~s.t.~D_{KL}(p_{\theta^t}||p_{\theta}) \le \nicefrac{\epsilon^2}{2}
\end{align}

Although problem (\ref{historical_problem}) can be solved with strong duality and additional convex optimization, applying a simple natural Lagrange condition will yields more room for accessible invariant with proper computational cost. Here we denote $G^t(\theta) \equiv \sum_{\tau = 0}^{T} \lambda^{\tau} D_{KL}(q_{\theta^{t - \tau}}||p_{\theta})$, \begin{equation}
    \tilde\triangledown_{\theta}|_{\theta = \theta^{t + 1}} (-G^t(\theta) + \eta(\nicefrac{\epsilon^2}{2} - D_{KL}(p_{\theta^t}||p_{\theta}))) = 0
    \label{nat_lag2}
\end{equation}

We name this algorithm family from iteratively solving (\ref{nat_lag2}) for different choice of parametric distribution family $\Theta$ as \framework. Further, when $T$ is set to be large, we can always replace $G^t(\theta)$ with a self-evolved term $M^t(\theta)$ that retain the gradient information. In practice, it suffice to evolve only $\tilde{\triangledown}_{\theta} M^t(\theta)$. \begin{align}
    \tilde{\triangledown}_{\theta} G^t(\theta) = -\tilde{\triangledown}_{\theta} M^t(\theta) +\tilde{\triangledown}_{\theta} D_{KL}(q_{\theta^t}||p_{\theta})
    \label{lag3}
\end{align}

\begin{assumption}
    The chosen fitness function $g_{f, \theta^t}(x)$ and the Lagrange multiplier $\eta$ are independent from the parameterization of $\theta$.  \label{assum_eta_g}
\end{assumption}

\begin{theorem}[Invariant for \framework]
    When assumption~\ref{assum_availability}, \ref{assum_eta_g}~hold and the decay weight $\lambda$ is independent from the parameterization of $\theta$, optimizers in \framework~are invariant and the single step computational cost is $\mathcal{O}(\min(HNT, HN + K))$ where $\mathcal{O}(K)$ denotes the cost to compute $\tilde{\triangledown}_{\theta} M^t(\theta)$.\label{thm_historical}
\end{theorem}

\section{Exemplifying with Multi-dimensional Gaussian}

We choose multi-dimensional Gaussian as our candidate distribution family $\Theta$ given its wide applications. To start with, we claim the computational accessibility of multi-dimensional Gaussian for assumption~\ref{assum_availability}, 
\begin{proposition}[Theorem 4.1 in \cite{akimoto2012theoretical}]\label{aki}
    Suppose $\theta_m$ and $\theta_c$ are $n$- and $n(n + 1)/2$-dimensional column representing mean and covariance respectively. Then $\partial m / \partial \theta_m$ and $\partial vec(C)/ \partial \theta_c$ are invertible at $\theta \in \Theta$ and, \begin{align}
        I_m^{-1}(\theta)\frac{\partial \ln{P_{\theta}(x)}}{\partial \theta_m} &= (\frac{\partial m}{\partial \theta_m})^{-1} (x - m) \\
        I_c^{-1}(\theta)\frac{\partial \ln{P_{\theta}(x)}}{\partial \theta_c} &= (\frac{\partial vec(C)}{\partial \theta_c})^{-1} vec((x - m)(x - m)^T - C)
    \end{align} 
\end{proposition}
Then we propose our choices of the fitness function $g_{f, \theta^t}(x)$ and the Lagrange multiplier $\eta$ to satisfy assumption~\ref{assum_eta_g}~while being as simple as possible. We denote the fitness function $g_{f, \theta^t}(x)$ as the level function that reflect the probability to sample a better value from $p_{\theta^t}$. This is the exact choice used in standard CMA-ES \cite{hansen2016cma} . In time step $t$, $N$ samples $\{x_i^t\}$ are drawn from $p_{\theta^t}$ and we further denote $\hat{w}_i^t \equiv \frac{g_{f, \theta^{t}}(x_i^t)}{\sum_i g_{f, \theta^{t}}(x_i^t)}$ as the normalized fitness for sample $x_i^t$. 

According to proposition~\ref{aki}, parameter $\theta = (\theta_m, \theta_c) \mapsto \mathcal{N}(m, C)$ with $\theta_m \in \mathbb{R}^n$ and $\theta_c \in \mathbb{R}^{n(n + 1) / 2}$ representing mean and covariance respectively. We can thus split the Lagrange multiplier into $\eta = (\eta_m, \eta_c)$ in~\framework~without violating the invariant property. For better comparisons with CMA family optimizers, we adopt this split to directly use the default values in the fine-tuned version of CMA-ES and keep them constant. The assumption~\ref{assum_eta_g}~is satisfied therefore.

We use the parameterization $\theta = (m, C)$ for simplification sake through this section. Different parameterizations that meet the conditions in proposition~\ref{aki}~will conduct different practical optimizers by following this section with minor modifications. The performance should be the same up to the transformation due to the invariant property.


\subsection{An Invariant Optimizer with Historical Information : \algo}

We directly apply~\framework~and a maximum time horizon $T = t - 1$. By choosing such infinite horizon, the historical information is maximally used. To reduce the computational costs to the same as IGO, i.e. scalable to $\mathcal{O}(HN)$ for single step update, we design $M^t(\theta)$ as follow, \begin{align}
    \tilde{\triangledown}_{m} M^t(\theta) =&~ \lambda_0(s^t_m + m^t - m)\label{assume_m}\\
    \label{assume_c}\tilde{\triangledown}_{c} M^t(\theta) =&~\lambda_0((s^t_c + m^t - m)(s^t_c + m^t - m)^T - C) + Q^t_1 + Q^t_2 \circ m + Q^t_3 mm^T\nonumber
\end{align}

Here scalars $\lambda_0, Q_1^t \in \mathbb{R}$ and vectors $s^t_m, s^t_c, Q^t_2, Q^t_3 \in \mathbb{R}^n$, with $\circ$ is used to denote $v_1 \circ v_2 \equiv v_1 v_2^T + v_2 v_1^T$ for two vectors $v_1, v_2 \in \mathbb{R}^n$. For simplicity needs, we denote $d_i^t \equiv x_i^t - m^t, d_w^t \equiv \sum_i \hat{w}_i^t d_i^t, \hat{d}_w^t \equiv d_w^t + m^t$ to represent statistics in a single generation, and $\hat{s}_m^{t - 1} \equiv s^{t - 1}_m + m^{t - 1}$, $\hat{s}_c^{t - 1} \equiv s^{t - 1}_c + m^{t - 1}$ to represent elements for history. Corresponding updates for hyperparameter $\lambda_0 \in \mathbb{R}$ and self-evolved terms $s^t_m, s^t_c, Q^t_2, Q^t_3 \in \mathbb{R}^n$ that initially zero are shown below. \begin{align}
    \lambda =&~ \nicefrac{\lambda_0}{\lambda_0 + 1}\\
    s^t_m + m^t =&~ \lambda \hat{s}_m^{t - 1} + (1 - \lambda) \hat{d}_w^{t - 1}\\
    s^t_c + m^t =&~ \sqrt{\lambda} \hat{s}_c^{t - 1} + \sqrt{1 - \lambda} \hat{d}_w^{t - 1}\\
    Q^t_1 =&~ \lambda Q^{t - 1}_1 + \lambda \sum \hat{w}_i (d_i^{t - 1} - d_w^{t - 1})(d_i^{t - 1} - d_w^{t - 1})^T - \lambda_0 \sqrt{\lambda} \sqrt{1 - \lambda} \hat{d}_w^{t - 1} \circ \hat{s}_c^{t - 1} \\
    Q^t_2 =&~ \lambda Q^{t - 1}_2 - \lambda_0(\sqrt{\lambda} + \sqrt{1 - \lambda} - 2) (\sqrt{\lambda} * \hat{s}_c^{t - 1}+ \sqrt{1 - \lambda} * \hat{d}_w^{t - 1}) \\
    Q^t_3 =&~ \lambda Q^{t - 1}_3 - \lambda_0(\sqrt{\lambda} - 1)(\sqrt{1 - \lambda} - 1)
\end{align}
We now arrive at the single step update for next parameter $\theta^{t + 1} = (m^{t + 1}, C^{t + 1})$. The resulting algorithm is named as~\algo~to emphasize another prominent characterization, the synchronous update nature, as discussed in section \ref{theoretical_comparison}, besides invariance. The final updates in single iteration with $z_m = \eta_m + \lambda_0 + 1, z_c = \eta_c + \lambda_0 + 1, \beta^t = \frac{1}{z_m}(d_w^t + \lambda_0 s^t_m)$ for brevity sake is shown below. \begin{align}
    m^{t + 1} =&~ m^t + \beta^t\label{mU}\\
    C^{t + 1} =&~ \frac{\eta_c}{z_c}(C^t + \beta^t(\beta^t)^T) + \frac{\lambda_0}{z_c}(s^t_c - \beta^t)(s^t_c - \beta^t)^T\label{cU} \\&+ \frac{1}{z_c} (\sum_i \hat{w}_i(d_i^t - \beta^t)(d_i^t - \beta^t)^T + Q^t_1 + Q^t_2 \circ m^t + Q^t_3 m^t(m^t)^T)\nonumber
\end{align}

\subsection{Theoretical Comparison with Other CMA Optimizers} \label{theoretical_comparison}

\paragraph{Theoretical advantages.} We here list four main theoretical advantages of~\algo~over other CMA optimizers : it is invariant, it has less hyperparamters, it fully incorporates historical information, and it is synchronous.

The invariant and historical information fully incorporated properties are grant by induced framework~\framework~. In the previous literature, no optimizer ever incorporates historical information into the mean update with a stable or invariant procedure. 

There are only three hyperparamters $\lambda_0, \eta_m, \eta_c$ for \algo, these hyperparameters all show up in all other proposed optimizers. Besides, other optimizers such as CMA-ES also have additional hyperparamters such as the weights and the initial value of learning rate $\sigma$ in the parameterization of $\theta = (m, \sigma^2\Sigma)$, and the weights for evolving historical information term.

The synchronous property from which~\algo~is named as a result of the fact that~\framework~is built on IGO, which treats the current distribution $\theta$ as a single point in space to update, i.e. the updates for mean and covariance are intertwined. For other CMA algorithms, the updates are performed sequentially in each generation, e.g. $m^{t + 1} = U_m(m^t, (\sigma^t)^2 \Sigma^t)$, $\Sigma^{t + 1} = U_c(m^{t + 1}, (\sigma^t)^2 \Sigma^t)$, etc. Moreover, to strictly follow the proposition~\ref{aki}~, updates need to be intertwined as~\algo~does.

\paragraph{Connection to CMA-ES.} It is also worth making the connection between~\algo~and the CMA family algorithms. Where the approximations used correspond to the four theoretical advantages of~\algo mentioned above.

\begin{proposition}
    When (1) the historical information is partially used for covariance, i.e. $\tilde{\triangledown}_{m} M^t(\theta) = 0$ and $\tilde{\triangledown}_{c} M^t(\theta) = \lambda_0((s^t_c + m^t - m)(s^t_c + m^t - m)^T - C)$. (2) all the higher order terms, when assuming $\eta_c \approx z_c \gg 1, z_m \gg 1$, are discarded. ~\algo~coincide with CMA-ES up to an external learning rate difference. \label{connection}
\end{proposition}

\section{Experiments} \label{exp} 

In this section, we intensively evaluate \algo~with other baselines in Mujoco locomotion tasks, rover planning task and synthetic functions. The criteria are chosen in the context of online optimization, focusing on full optimization procedures in the natural axis and sample efficiency when achieve a near global value. All optimization procedures are plotted with the shaded area bounded by quantiles and the solid line denoting the median performance over all trails. 

Baselines are chosen in a structured way. First, random search (RS) \citep{bergstra2012random}~is chosen as the overall baseline. Then, two black-box optimizers, differential evolution (DE) \citep{storn1997differential} and simulated annealing (SA) \citep{bouttier2019convergence} are chosen. Among the CMA optimizers, we choose CMA-ES and two of its leading variants DD-CMA\citep{akimoto2020diagonal} and TR-CMA-ES\citep{abdolmaleki2017deriving} for detailed comparison. Finally, the Bayesian optimization method TuRBO \citep{eriksson2019scalable} is used as the state-of-the-art baseline for BO. Parameters $\eta_m, \eta_c$ of \algo~are set to constant that match the initial settings of the corresponding parameters in CMA-ES. $\lambda_0$ corresponds to a combination of several parameters in CMA-ES so we simply test with the constant value $\lambda_0 = 2$, which corresponds to the approximate counterpart in CMA-ES. We use this value throughout the main paper while there are better performances of \algo~with different $\lambda_0$ as shown in ablation studies in Appendix \ref{appendix_ablation}. The initial distribution for all Gaussian based optimizers are identity matrix.

TR-CMA-ES is based on its original paper version implemented in Matlab due to precision problems in Python, and therefore we exclude TR-CMA-ES in the Mujoco locomotion and rover planning tasks as they are based on specific Python libraries. All other baselines are implemented with their fine-tuned version available online \citep{duan2022pypop7, balandat2020botorch}. See Appendix \ref{appendix_exp} for details.

\subsection{Mujoco Locomotion Task}

We first evaluate \algo~and other baselines on the widely tested Mujoco locomotion tasks \citep{todorov2012mujoco}, which are popular benchmarks for Bayesian optimization and reinforcement learning algorithms. To run sampling-based optimizers on Mujoco,  we refer to \citep{mania2018simple} and optimize a linear policy: $\textbf{a} = \textbf{Ws}$, where $\textbf{a}$ is the agent action and $\textbf{s}$ is the environment state. The parameter matrix $\textbf{W}$ are continuous and in the range of $[-1, 1]$. Among all 6 tasks, we dismiss the overly high dimensional task Humanoid(6392d) and test all other 5 tasks with batch size $N = 100$. Two results are shown here in figure \ref{fig:ant}, \ref{fig:half}~with more results in Appendix \ref{appendix_mujoco}. While TuRBO dominates other baselines, \algo~outperforms TuRBO in 2 tasks and remains competitive with TuRBO for the other 3 tasks.
\begin{figure}[!htb]
    \centering
    \subfigure[Ant(888d)]{
        \includegraphics[width = .3\linewidth]{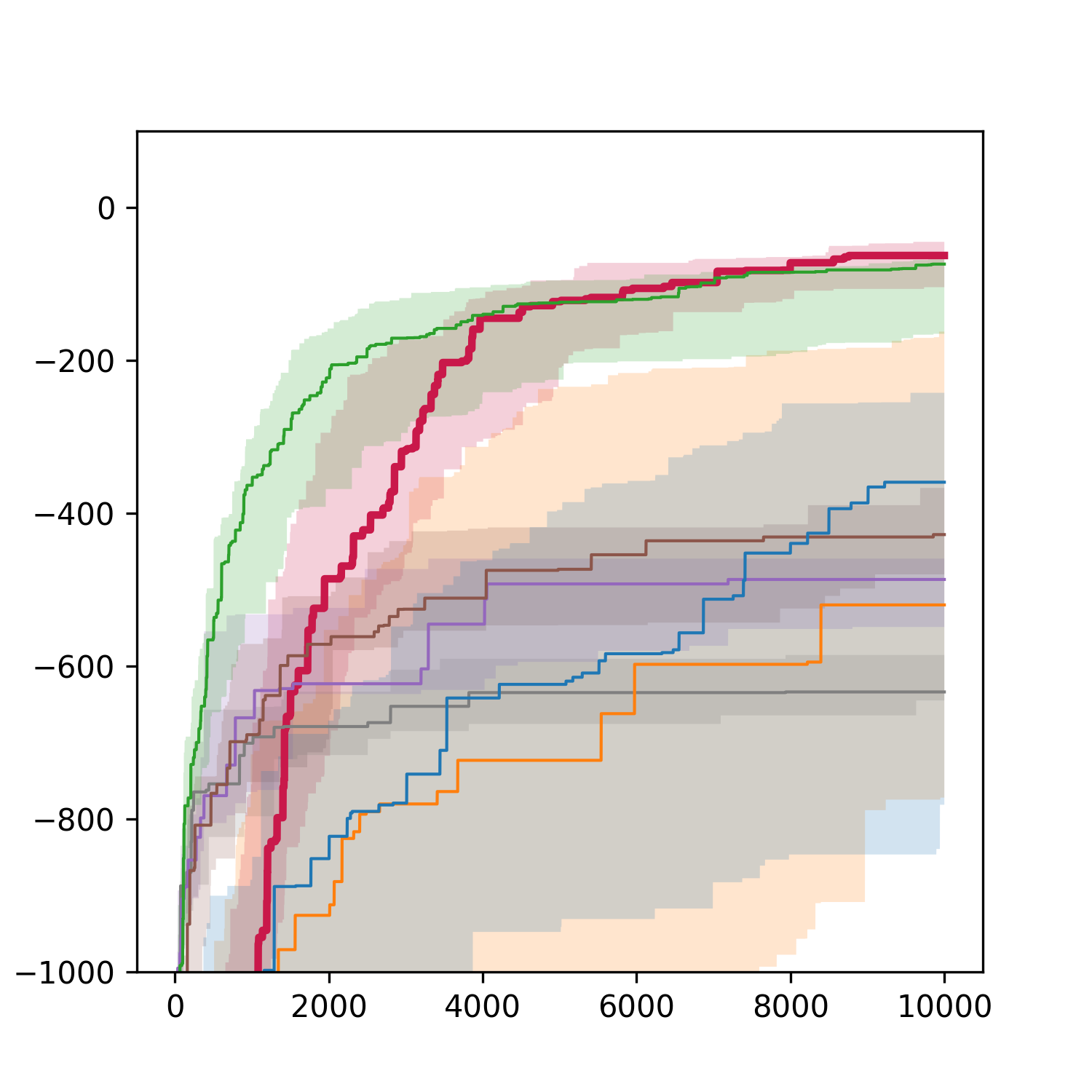}
        \label{fig:ant}
    }
    \subfigure[HalfCheetah(102d)]{
        \includegraphics[width = .3\linewidth]{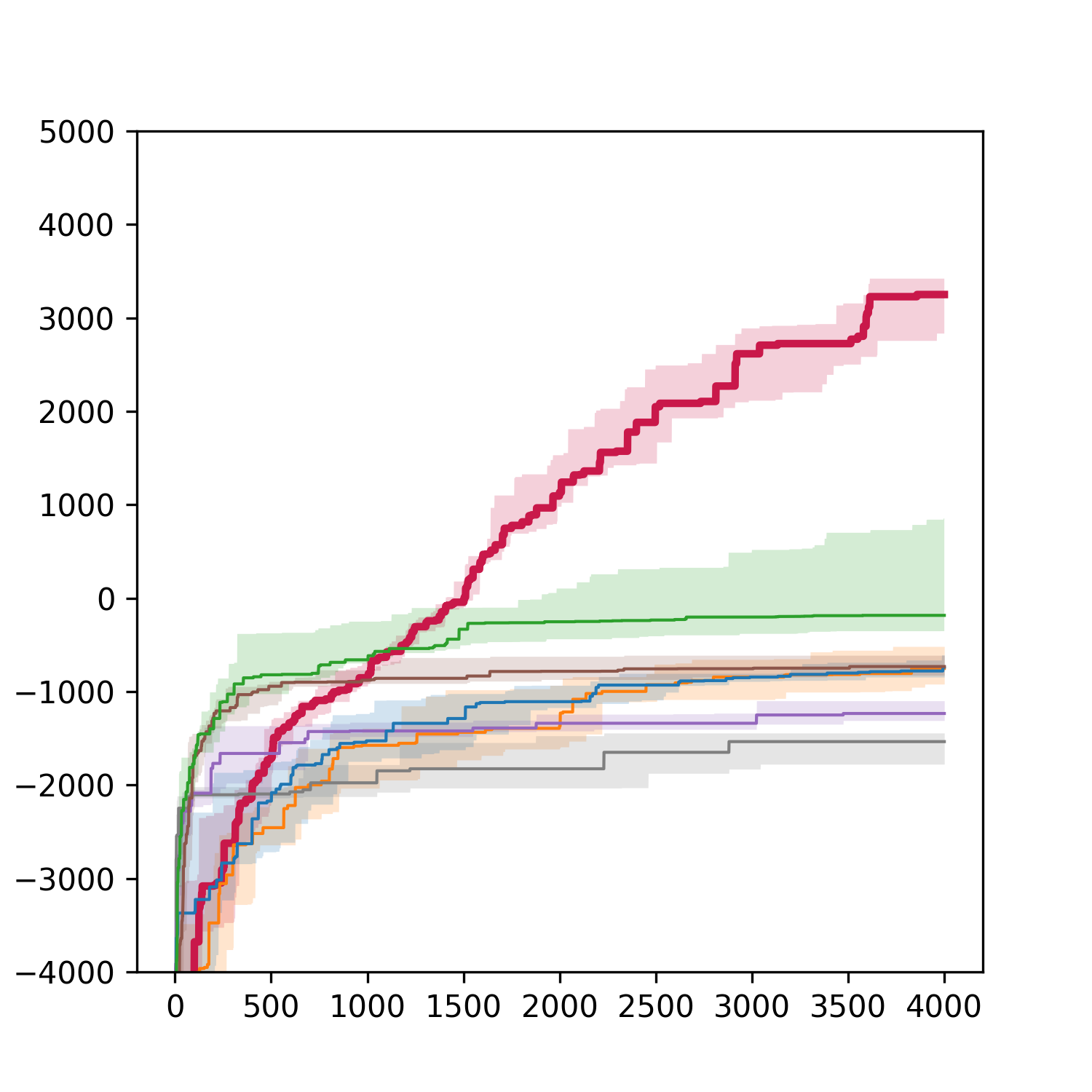}
        \label{fig:half}
    }
    \subfigure[Rover(60d)]{
        \includegraphics[width = .3\linewidth]{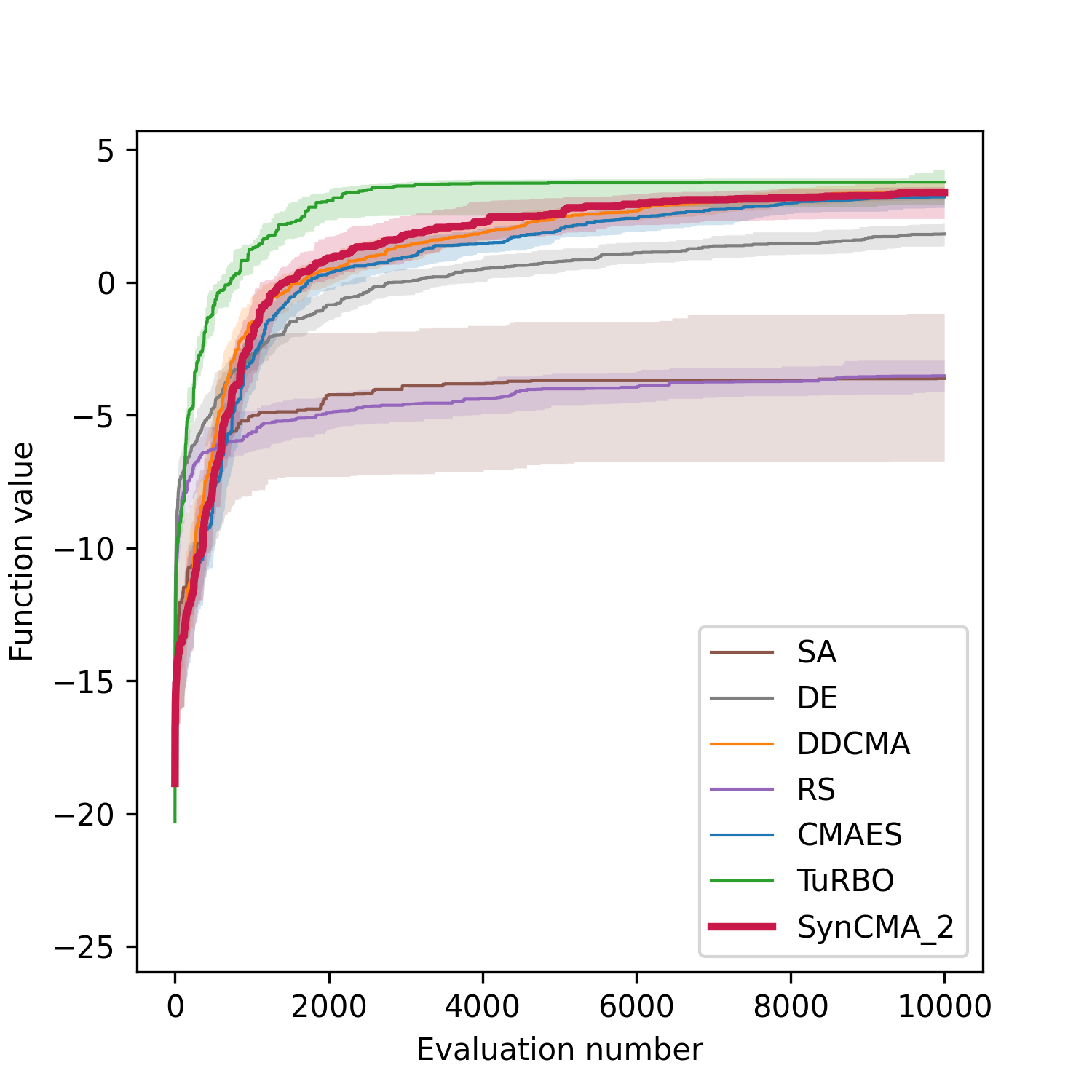}
    }
    \caption{Optimization procedure for two high dimensional Mujoco locomotion tasks over 10 trials and rover planning task over 100 trails. Index of \algo~indicate $\lambda_0$.}
\end{figure}
\subsection{Rover Planning Task}

To further explore the empirical performance of \algo~in a realistic setting, we consider the rover trajectory optimization task, where a start position $s$ and a goal position $g$ are defined in the 2D plane, as well as a cost function $c(x)$ over the state space. The trajectories are described by a set of points to which a B-spline is fitted and the cost function is computed. The whole state space is $x\in [0,1]^{60}$ and we make the batch size $N = 2n = 120$. A reward function to be optimized is defined to be non-smooth, discontinuous, and concave over the first two and last two dimensions of the state. The result in figure \ref{fig:rover}~shows that \algo~still exhibits competitive performance over other baselines.

\subsection{Synthetic Function}

We select 10 commonly used synthetic functions with dimension $n$ arbitrarily set. This is the traditional test bed for black-box optimization and specific evolution strategies studies. These functions, including different characteristics such as multi-model, ill-conditioned and ill-scaled, are scaled to a global minimum value $0$ with shifted domain. 
The batch size is $N = 2n$ and the evaluation limit is the same for all optimizers except TuRBO, where the budget is fixed at 5,000 evaluations due to memory limitations. 

The full experiments are run with different dimensions of $n = \{32, 64, 128\}$ and the results are presented in two ways under the same evaluation budget: near global optimum performance and the whole optimization procedure. Some results are presented here and please refer to Appendix \ref{appendix_syn}~for the full experimental results.

\begin{figure}[!htb]
    \centering
    \subfigure[Discus]{
        \includegraphics[width = .23\linewidth]{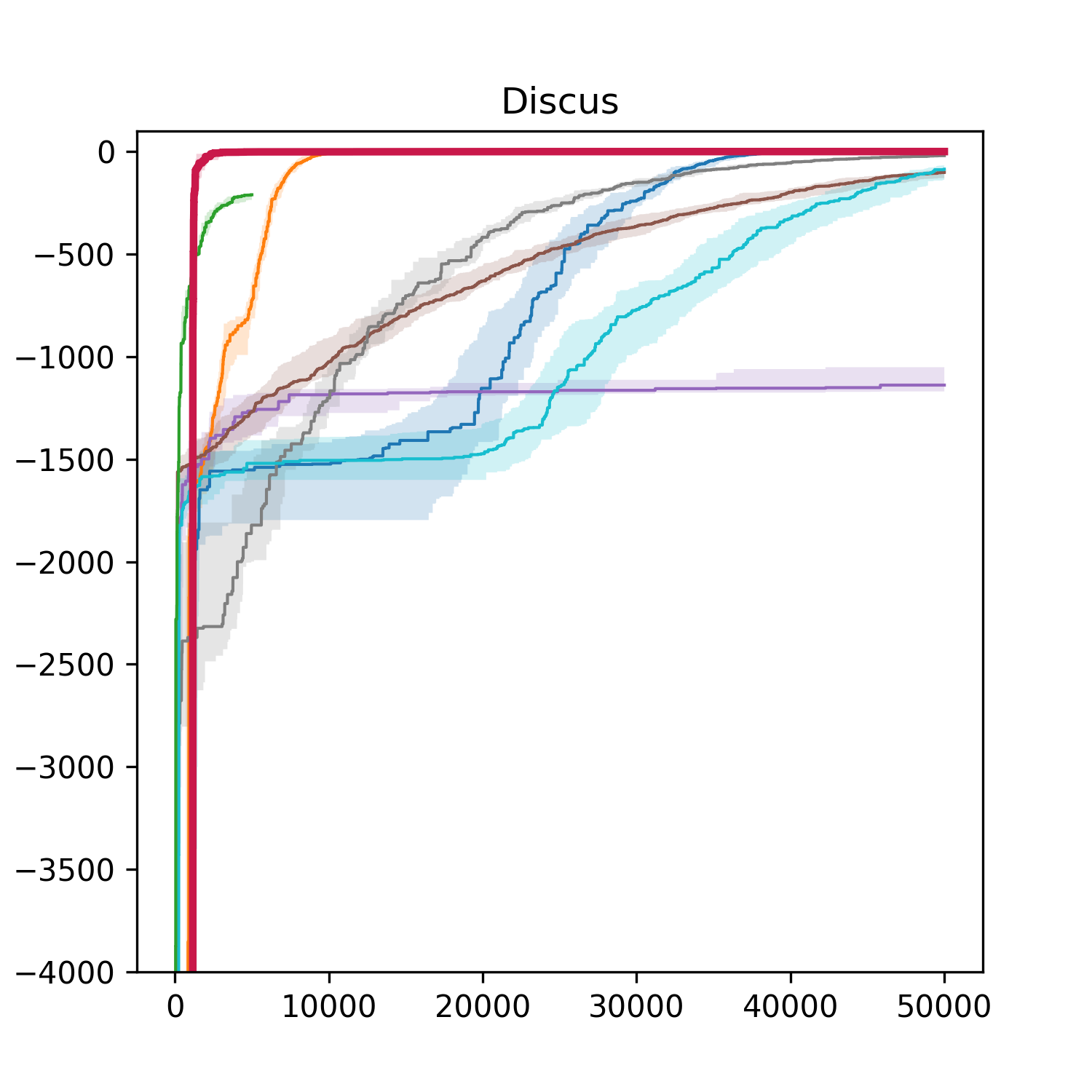}
    }
    \subfigure[Rastrigin]{
        \includegraphics[width = .23\linewidth]{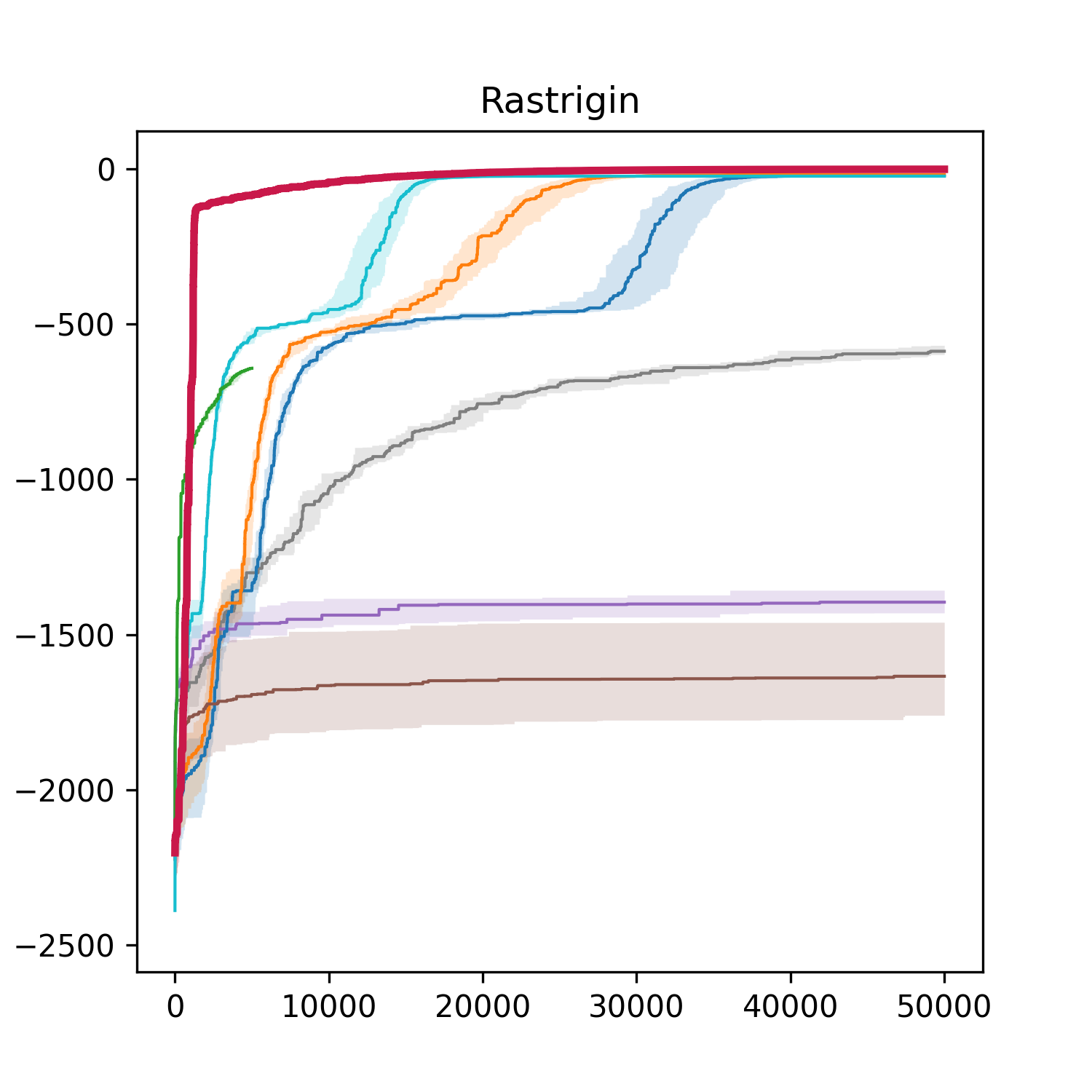}
    }
    \subfigure[LevyMontalvo]{
    \includegraphics[width = .23\linewidth]{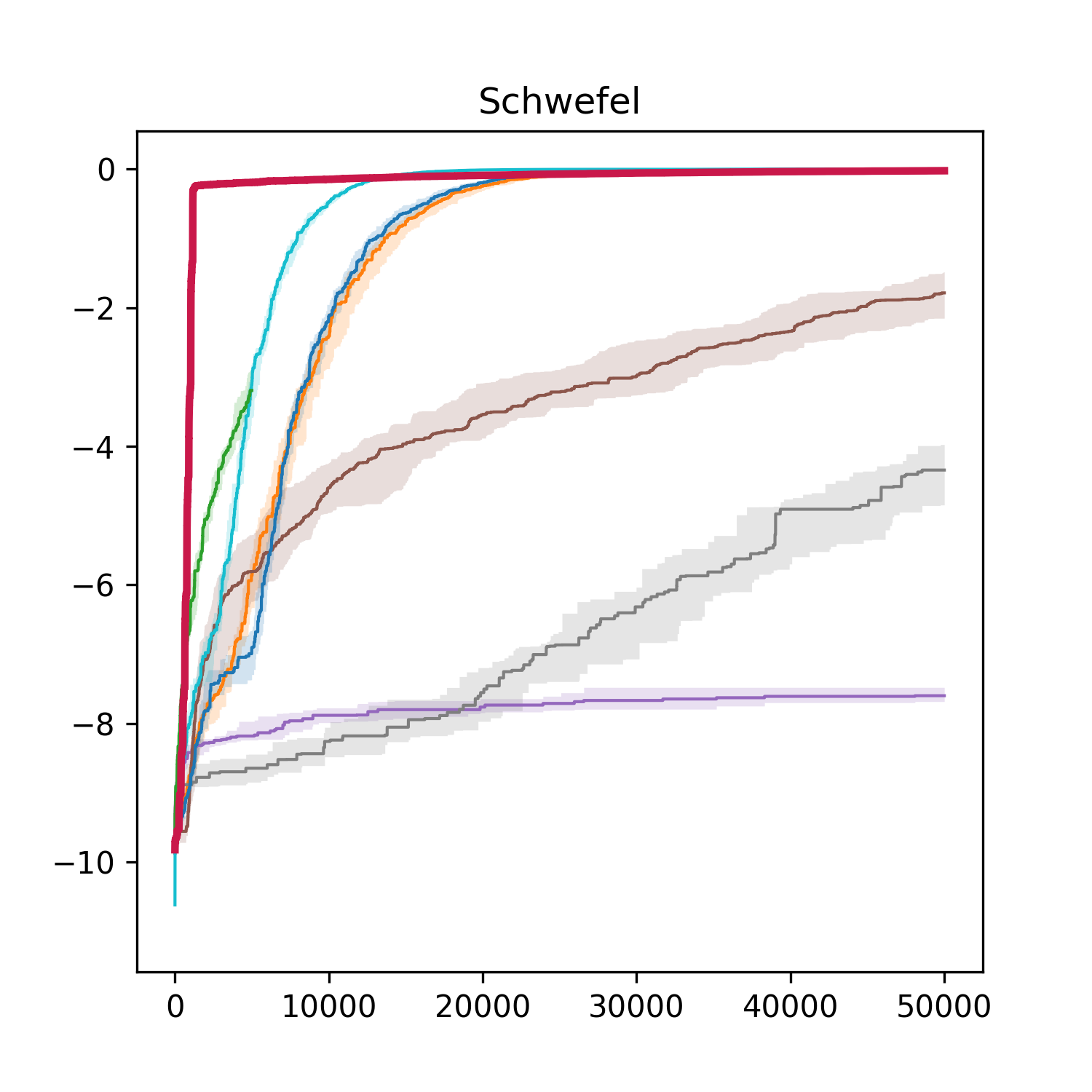}
    }
    \subfigure[Rosenbrock]{
        \includegraphics[width = .23\linewidth]{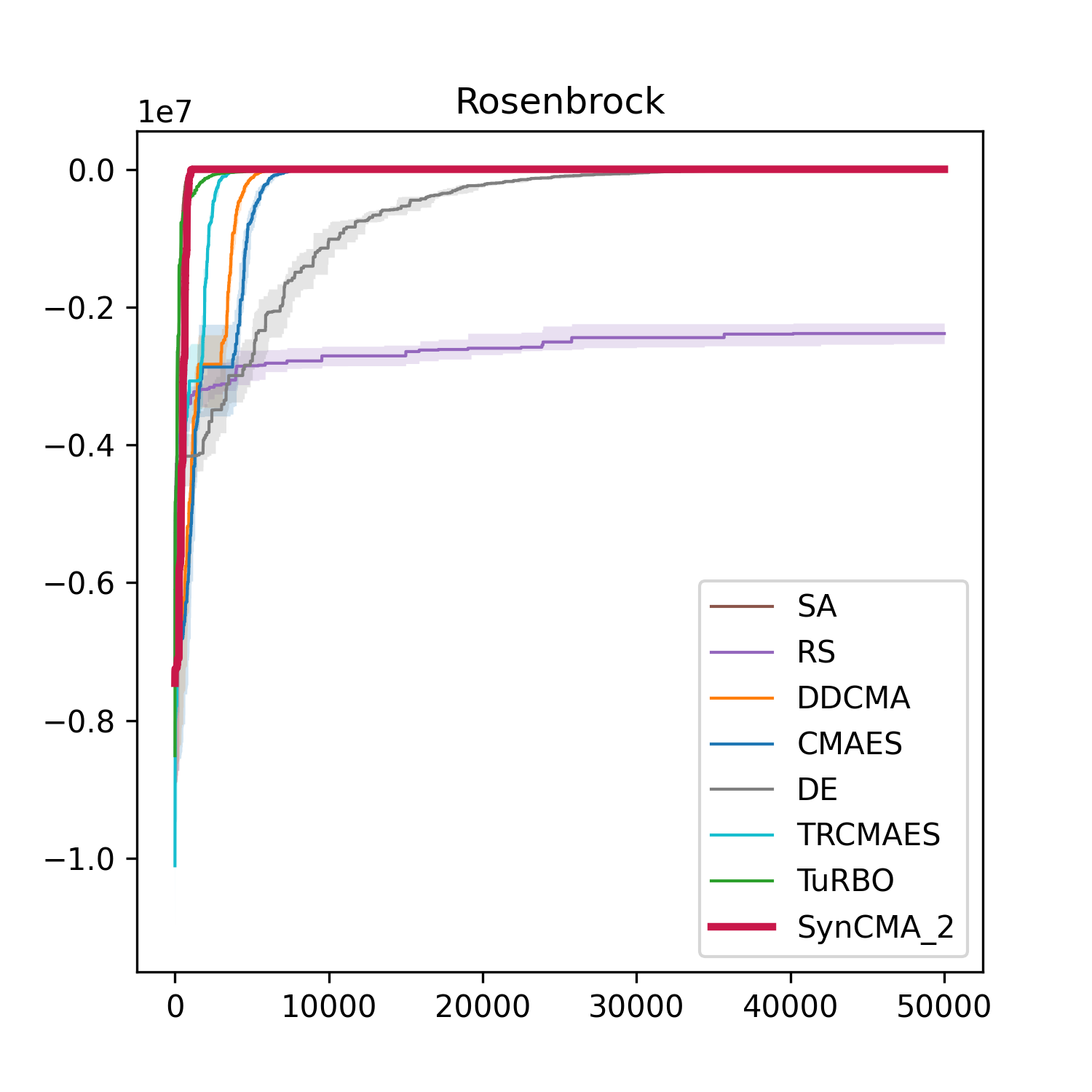}
    }
    \caption{Optimization procedure in 4 typical synthetic functions with dimension $n = 64$ over 20 trails considering all optimizers. Index of \algo~indicate $\lambda_0$.}
    \label{fig:syn}
\end{figure}

\begin{table}[th]
\caption{Near global optimum performance on 64d synthetic functions(lower is better) over 20 trials with budget of 50000 evaluations, TuRBO is excluded due to memory limitation. Numbers in brackets indicates the median evaluation number needed for optimizers to achieve value better than 0.5.}
    \label{tab:dim}
\resizebox{\textwidth}{!}{%
\begin{tabular}{cccccccc}
        \toprule 
        Optimizer & Sphere & Discus & Schwefel & DiffPowers & LevyMontalvo & Rastrigin & Ackley \\ 
        \midrule
        SA & 0.6 & 100.7 & 0.1(2650) & 37.5 & 6.8 & 1634.7 & 13.6 \\ 
        RS & 793.9 & 1138.8 & 29902.6 & 2369.4 & 14.4 & 1395.9 & 11.8 \\ 
        DE & 7.2 & 20.0 & 25.6 & 63.6 & 0.4(49700) & 586.9 & 3.1 \\ 
        DDCMA & 0.0(10047) & \textbf{0.0(12748)} & 0.0(16820) & 0.0(10164) & 0.0(9087) & 17.9 & 0.0(11757) \\ 
        CMAES & 0.0(13825) & 0.0(43265) & 0.0(16134) & 0.0(18503) & 0.0(9901) & 21.4 & 0.0(15620) \\ 
        TRCMAES & 0.0(7185) & 85.9 & 0.0(9905) & 0.0(12040) & 0.0(5515) & 22.4 & 0.0(7869) \\ 
        \text{\algo (\textbf{Ours})} & \textbf{0.0(3938)} & 0.0(18820) & \textbf{0.0(1157)} & \textbf{0.0(1158)} & \textbf{0.0(2318)} & \textbf{0.2(42696)} & \textbf{0.0(7567)} \\ 
        \bottomrule
    \end{tabular}
  }
\end{table}
According to table \ref{tab:dim}~where TuRBO is excluded as it is unable to scale to this budget, \algo~demonstrate both superior optimization capability and efficiency over others. While other optimizers are less efficient and fail to optimize high-conditional number multi-model function Rastrigin, ill-scaled function Discus and others. Further, full optimization procedures including TuRBO with maximum budget under storage limit are partially shown in figure \ref{fig:syn}, \algo~still outperforms others including TuRBO after first several hundreds evaluation from 32 to 128 dimension, demonstrating the capability of such optimizer derived from an invariant framework.

\subsection{Ablation Study}

The weight for historical information $\lambda_0$ is a parameter that substitutes a combination of several parameters in CMA optimizers, and is set constantly as $\lambda_0 = 2$. We thus study the sensitivity on this parameter for constant setting here. All of previous experiments are repeated for $\lambda_0 \in [0, 4]$, with results for $\lambda_0 = \{0, 1, 2, 4\}$ shown in Appendix \ref{appendix_ablation}, from which we summarize several observations within range $[0, 4]$ here.

\textbf{Sensitivity}. When \algo~includes historical information, i.e. $\lambda_0 > 0$, \algo~consistently shows competitive performance.

\textbf{Function Landscape}. When their exists a fundamental subspace that covers the structure of the problem, as in Rastrigin, a higher $\lambda_0$ yields better performance and efficiency. Otherwise, as in LevyMontalvo, a higher $\lambda_0$ might be detrimental.

\textbf{Dimensionality}. Observed from tasks in Mujoco, synthetic functions, and rover planning, a higher dimension generally requires a higher $\lambda_0$.

\section{Limitations}
There is still much to explore in both framework~\framework~and optimizer~\algo~. For the framework, we use a line search for each step to yield rooms for invariant and complexity, and it may need more theoretical results to characterize its behavior. Moreover, it currently only works for certain parametric distribution families that IGO set. As we are based on the approximation of the IGO objective, it is possible to generalize to a broader family of models such as neural networks. Additionally, the utilized line search hamper a rigorous theoretical analysis, which may of single interest to design per step subroutine while keep the framework invariant and scalable.

For the algorithm, we set parameters $\eta_m, \eta_c$ and $\lambda_0$ constants in the experiments, which might make~\algo~overshoot in the final stage of optimization when it is close to the optimum. However, they can absolutely be invariant variables with evolving rules for a better overall optimization performance. 

\section{Conclusion}

We present an invariant optimizer framework~\framework~that fully incorporates historical information. When exemplified with multi-dimensional Gaussian, our framework derives a invariant optimizer \algo~that retains the computational complexity as in information geometric optimization. With a straightforward invariant oriented motivation, \algo~shows competitive performance in both realistic and synthetic scenarios against leading Bayesian and evolution strategies optimizers. 

We highlight its performance on high dimensional realistic problem as it shows the potential of a property oriented evolution strategies optimizer against Bayesian optimization optimizers. And we also defend the importance of fully invariant in optimization. Moreover, we believe the property oriented perspective is more approachable than inventing a rigorous theory that illustrates and improves current methods.

\newpage

\bibliography{my}

\newpage
\appendix
\section*{Appendices : An Invariant Information Geometric Method for High-dimensional Online Optimization}

\setcounter{section}{0}  

\renewcommand\thesection{\Alph{section}}  

\section{Proofs and Derivations}\label{proof_derivation}

\subsection{Proof for Theorem~\ref{thm_approx}}

\begin{proof}
    To prove the former part, it is suffice to notice, \begin{align}
        \triangledown_{\theta} \log L_{\theta^t}(\theta) - (-\triangledown_{\theta}D_{KL}(q_{\theta^t}||p_{\theta})) &~= \triangledown_{\theta} D_{KL}(q_{\theta^t}||q_{\theta})\\
        &~= \triangledown_{\theta} (\frac{1}{2}\sum I_{ij}^{q_{\theta^t}}(\theta^t)\delta\theta_i \delta \theta_j + O(\delta \theta^3))
    \end{align} 

    To prove the latter part, we first apply natural gradient to $L_{\theta^t}(\theta)$, \begin{align}
        \tilde{\triangledown}_{\theta} L_{\theta^t}(\theta) =&~ \tilde{\triangledown}_{\theta} \int g_{f, \theta^t}(x) p_{\theta}(x) \mathrm{d}x \\
        =&~ I^{-1}(\theta) \int g_{f, \theta^t}(x) \frac{\partial \ln{p_{\theta}(x)}}{\partial \theta} \frac{p_{\theta}(x)}{p_{\theta^t}(x)} p_{\theta^t}(\mathrm{d}x)
    \end{align}  In contrast, applying natural gradient to $D_{KL}(q_{\theta^t}||p_{\theta})$, \begin{align}
        \tilde{\triangledown}_{\theta} D_{KL}(q_{\theta^t}||p_{\theta}) =&~ 
        -I^{-1}(\theta) \int \frac{g_{f, \theta^t}(x)}{E_{p_{\theta^t}}[g_{f, \theta^t}(x)]} \frac{\partial \ln{p_{\theta}(x)}}{\partial \theta} p_{\theta^t}(\mathrm{d}x)
    \end{align} Thus under assumption~\ref{assum_availability}~, only $D_{KL}(q_{\theta^t}||p_{\theta})$ is available to take natural gradient on any point while the computational cost remain the same 
\end{proof}

\subsection{Proof for Theorem~\ref{thm_historical}}

\begin{proof}
    Similar to $\theta$, we define $q'_{\varphi^t}(x) = \frac{p'_{\varphi^t}(x)g_{f, \varphi^t}(x)}{\mathbb{E}_{p'_{\varphi^t}}[g_{f, \varphi^t}(x)]}$. Initially, we have $q'_{\varphi^0}(x) = q_{\theta^0}(x)$ from given conditions. So we assume for the current time $t$, $q'_{\varphi^t}(x) = q_{\theta^t}(x)$ holds as well. Then from equation (\ref{nat_lag2}), we have \begin{align}
        &\tilde{\triangledown}_{\varphi}|_{\varphi = \varphi^{t + 1}}(-\sum_{\tau = 0}^T \lambda^\tau D_{KL}(q'_{\varphi^{t - \tau}}||p'_{\varphi}) + \eta(\nicefrac{\epsilon^2}{2} - D_{KL}(p'_{\varphi^t}||p'_{\varphi}))) \\
        =&~ \tilde\triangledown_{\theta}|_{\theta = \varphi^{-1}(\varphi^{t + 1})} (-\sum_{\tau = 0}^{T} \lambda^{\tau} D_{KL}(q_{\theta^{t - \tau}}||p_{\theta}) + \eta(\nicefrac{\epsilon^2}{2} - D_{KL}(p_{\theta^t}||p_{\theta}))) \cdot (\frac{\partial \theta}{\partial \varphi}|_{\theta = \varphi^{-1}(\varphi^{t + 1})})^3 \\
        =&~ 0
    \end{align}
    Thus we have $\varphi^{t + 1} = \varphi(\theta^{t + 1})$. From mathematical infuction, our proof finished.
\end{proof}

\subsection{Derivations of~\algo}

The derivations consist of two parts. The first part is to substitute optimization objevtive $G^t(\theta) = \sum_{\tau = 0}^{t - 1} \lambda^{\tau} D_{KL}(q_{\theta^{t - \tau}}||p_{\theta})$ with a self-evolved term $M^t(\theta)$ so that the gradient information is completely preserved as equation (\ref{lag3}) while the computational costs reduce to the same as IGO, i.e. the cost of computing $\tilde{\triangledown}_{\theta}|_{\theta = \theta^t} L_{\theta^t}(\theta)$. The second part is to derive analytical updates for $\theta^{t + 1} = (m^{t + 1}, C^{t + 1})$ from solving equation (\ref{nat_lag2}). 

\subsubsection{Subsititution of $G^t(\theta)$}

We start from our target of preserving historical information, \begin{align}
    \tilde{\triangledown}_{\theta} G^t(\theta) = -\tilde{\triangledown}_{\theta} M^t(\theta) +\tilde{\triangledown}_{\theta} D_{KL}(q_{\theta^t}||p_{\theta})
\end{align}

We assume $M^t(\theta)$ has the form that present in the main paper, with scalars $\lambda_0, Q_1^t \in \mathbb{R}$ and vectors $s^t_m, s^t_c, Q^t_2, Q^t_3 \in \mathbb{R}^n$ all start from zero. We use $\circ$ to denote $v_1 \circ v_2 \equiv v_1 v_2^T + v_2 v_1^T$ for two vectors $v_1, v_2 \in \mathbb{R}^n$. \begin{align}
    \tilde{\triangledown}_{m} M^t(\theta) =&~ \lambda_0(s^t_m + m^t - m)\\
    \tilde{\triangledown}_{c} M^t(\theta) =&~\lambda_0((s^t_c + m^t - m)(s^t_c + m^t - m)^T - C)\\ +&~Q^t_1 + Q^t_2 \circ m + Q^t_3 mm^T\nonumber
\end{align}

Notice that $G^t(\theta) = \lambda G^{t - 1} + D_{KL}(q_{\theta^t}||p_{\theta})$, therefore we have, \begin{equation}
    \tilde{\triangledown}_{\theta}M^t(\theta) = \lambda (\tilde{\triangledown}_{\theta}M^{t - 1}(\theta) - \tilde{\triangledown}_{\theta}D_{KL}(q_{\theta^{t - 1}}||p_{\theta})) \label{derive_M}
\end{equation}

\paragraph{Mean Solution}. Apply assumption (\ref{assume_m}) for $\tilde{\triangledown}_{m} M^t(\theta)$ yields, \begin{align}
    \lambda_0(s^t_m + m^t - m) = \lambda\lambda_0(s^{t - 1}_m + m^{t - 1} - m) + \lambda \sum_i \hat{w}_i(x^{t - 1}_i - m)
\end{align}

Which straightly gives the only solutions on $\lambda$ and $s^t_m$, \begin{align}
    \lambda =&~ \frac{\lambda_0}{\lambda_0 + 1}\\
    s^t_m + m^t =&~ \lambda s_m^{t - 1} + (1 - \lambda) d_w^{t - 1} + m^{t - 1}
\end{align}

\paragraph{Covariance Maxtrix Solution}Apply assumption (\ref{assume_c}) for $\tilde{\triangledown}_{c} M^t(\theta)$ while denoting $Q^t(m) = Q^t_1 + Q^t_2 \circ m + Q^t_3 mm^T$ yields, \begin{align}
    &~\lambda_0((s^t_c + m^t - m)(s^t_c + m^t - m)^T - C) + Q^t(m)\label{supneedc}\\
    =&~ \lambda\lambda_0((s^{t - 1}_c + m^{t - 1} - m)(s^{t - 1}_c + m^{t - 1} - m)^T - C) + \lambda Q^{t - 1}(m) \nonumber\\&~+ \lambda(\sum_i \hat{w}_i (x^{t - 1}_i - m)(x^{t - 1}_i - m)^T - C)\nonumber
\end{align}

To cancel out $C$ on both side, same value of $lambda = \nicefrac{\lambda_0}{\lambda_0 + 1}$ is derived. Further, we assume the form of update on $s^t_c$ is similar to $s^t_m$ with parameter $\alpha$ and $zeta$ undetermined yet, \begin{equation}
    s^t_c + m^t = \zeta(s^{t - 1}_c + m^{t - 1}) + \alpha(d^{t - 1}_w + m^{t - 1})
\end{equation}

Thus we arrive at, \begin{align}
    (s^t_c + m^t - m)(s^t_c + m^t - m)^T =&~ \zeta^2(s^{t - 1}_c + m^{t - 1} - m)(s^{t - 1}_c + m^{t - 1} - m)^T \label{sup_c}\\
    &~+\alpha^2(d^{t - 1}_w + m^{t - 1} - m)(d^{t - 1}_w + m^{t - 1} - m)^T \nonumber\\
    &~+ (\alpha + \zeta - 2)m \circ (\zeta(s^{t - 1}_c + m^{t - 1}) + \alpha(d^{t - 1}_w + m^{t - 1}))\nonumber\\
    &~+ (\alpha - 1)(\zeta - 1)mm^T \nonumber\\
    &~+ \alpha \zeta (d^{t - 1}_w + m^{t - 1}) \circ (s^{t - 1}_c + m^{t - 1})\nonumber
\end{align}

Notice that for the second term of Eq.\ref{sup_c}, we have,\begin{align}
    &~(d^{t - 1}_w + m^{t - 1} - m)(d^{t - 1}_w + m^{t - 1} - m)^T\nonumber\\ =&~ \mathbb{E}_{q_{\theta^{t - 1}}}[x - m]\mathbb{E}_{q_{\theta^{t - 1}}}[x - m]^T\\
    =&~ \mathbb{E}_{q_{\theta^{t - 1}}}[(x - \mathbb{E}_{q_{\theta^{t - 1}}}[x] + \mathbb{E}_{q_{\theta^{t - 1}}}[x] - m)(x - \mathbb{E}_{q_{\theta^{t - 1}}}[x] + \mathbb{E}_{q_{\theta^{t - 1}}}[x] - m)^T]\\
    &~- \mathbb{E}_{q_{\theta^{t - 1}}}[(x - \mathbb{E}_{q_{\theta^{t - 1}}}[x])(x - \mathbb{E}_{q_{\theta^{t - 1}}}[x])^T]\nonumber\\
    =&~ \mathbb{E}_{q_{\theta^{t - 1}}}[(x - m)(x - m)^T] - \mathbb{E}_{q_{\theta^{t - 1}}}[(x - \mathbb{E}_{q_{\theta^{t - 1}}}[x])(x - \mathbb{E}_{q_{\theta^{t - 1}}}[x])^T]\\
    =&~ \sum_i \hat{w}_i (x^{t - 1}_i - m)(x^{t - 1}_i - m)^T - \sum_i \hat{w}_i (d^{t - 1}_i - d^{t - 1}_w)(d^{t - 1}_i - d^{t - 1}_w)^T
\end{align}
Thus, to meet Eq.\ref{supneedc}, we arrive at updates for $s^t_c, Q^t_1, Q^t_2, Q^t_3$ with $\zeta = \sqrt{\lambda}, \alpha = \sqrt{1 - \lambda}$, \begin{align}
    s^t_c + m^t =&~ \zeta s_c^{t - 1} + \alpha d_w^{t - 1} + (\zeta + \alpha) m_{t - 1}\\
    Q^t_1 =&~ \lambda Q^{t - 1}_1 + \lambda \sum \hat{w}_i (d_i^{t - 1} - d_w^{t - 1})(d_i^{t - 1} - d_w^{t - 1})^T  \\
    &~- \lambda_0 \alpha \zeta (d^{t - 1}_w + m^{t - 1}) \circ (s^{t - 1}_c + m^{t - 1})\nonumber \\
    Q^t_2 =&~ \lambda Q^{t - 1}_2 - \lambda_0 (\zeta + \alpha - 2) (\zeta * (s^{t - 1}_c + m^{t - 1})+ \alpha * (d_w^{t - 1} + m^{t - 1}))\\
    Q^t_3 =&~ \lambda Q^{t - 1}_3 - \lambda_0 (\zeta - 1)(\alpha - 1)
\end{align} 

\subsubsection{Derivation of $\theta^{t + 1}$}

Substitute the $G^t(\theta)$ in the natural Lagrange condition (\ref{nat_lag2}) gives equation, \begin{align}
    0 =&~ \tilde\triangledown_{\theta}|_{\theta = \theta^{t + 1}} (-G^t(\theta) + \eta(\nicefrac{\epsilon^2}{2} - D_{KL}(p_{\theta^t}||p_{\theta})))\\
    =&~\tilde\triangledown_{\theta}|_{\theta = \theta^{t + 1}} (M^t(\theta) - D_{KL}(q_{\theta^t}||p_{\theta}) - [\eta_m, \eta_c]^T D_{KL}(p_{\theta^t}||p_{\theta}) )
\end{align}

Apply proposition~\ref{aki}~on above equation, we now arrive at equations for $\theta^{t + 1}$, \begin{align}
    \sum_i (\hat{w}^t_i + \frac{\eta_m}{N})(x^t_i - m^{t + 1}) + \tilde{\triangledown}_{m} M^t(\theta) &= 0 \label{lag_m}\\
    \sum_i (\hat{w}^t_i + \frac{\eta_c}{N})((x^t_i - m^{t + 1})(x^t_i - m^{t + 1})^T - C^{t + 1}) + \tilde{\triangledown}_{c} M^t(\theta) &= 0 \label{lag_c}
\end{align}

Solving equations above straightly give updates with $z_m = \eta_m + \lambda_0 + 1, z_c = \eta_c + \lambda_0 + 1, \beta^t = \frac{1}{z_m}(d_w^t + \lambda_0 s^t_m)$ for brevity sake. \begin{align}
    m^{t + 1} =&~ m^t + \beta^t\\
    C^{t + 1} =&~ \frac{\eta_c}{z_c}(C^t + \beta^t(\beta^t)^T) + \frac{\lambda_0}{z_c}(s^t_c - \beta^t)(s^t_c - \beta^t)^T \\&+ \frac{1}{z_c} (\sum_i \hat{w}_i(d_i^t - \beta^t)(d_i^t - \beta^t)^T + Q^t_1 + Q^t_2 \circ m^t + Q^t_3 m^t(m^t)^T)\nonumber
\end{align}

\subsection{Proof and discussions on Proposition~\ref{connection}}

\label{proof_prop_connection}

\begin{proof}
    We apply the first condition to equations (\ref{lag_m}, \ref{lag_c}), which gives the updates for $\theta^{t + 1}$ with $D_w^t = \sum_i \hat{w}_i^t (x_i^t - m^t)(x_i^t - m^t)^T$, \begin{align}
    m^{t + 1} =&~ m^t + \frac{1}{z_m}d_w^t\\
    C^{t + 1} =&~ \frac{\eta_c}{z_c}C^t + \frac{1}{z_c}D_w^t + \frac{\lambda_0}{z_c}s_c^t(s_c^t)^T\\
    &~- \frac{\lambda_0}{z_cz_m}(d_w^t \circ (s_c^t)) - \frac{1}{z_cz_m}(2 - \frac{z_c}{z_m})d_w^t(d_w^t)^T\nonumber
\end{align}

Then under the second condition, the last two terms in the update of $C^{t + 1}$ should be discarded. The rest part coincide with CMA-ES up to an external learning rate difference.
\end{proof}

\section{Experiments} \label{appendix_exp}

We implement \algo~using PyPop7 \citep{duan2022pypop7}\footnote{\url{https://github.com/Evolutionary-Intelligence/pypop}}. All baselines except TuRBO and TR-CMA-ES are also implemented with this library. Except for the hyperparameters mentioned in the main paper, all evolutionary based algorithms use default options. In \algo, we use constant learning rate $\sigma = 0.1$ in order to match the initial learning rate in other CMA optimizers to sample with $\mathcal{N}(m, \sigma \Sigma)$. As CMA optimizers also use $\sigma$ as the update fraction for mean, we set $\eta_m$ such that $\nicefrac{1}{z_m} = 0.1$ accordingly. $\eta_c$ is set two times of the corresponding value in CMA-ES, as several tunning techniques used in fine-tuned version of CMA-ES may let its actual corresponding value larger.
For TuRBO, we refer to the implementation of BoTorch and replicate the original algorithm. We set the trust region number as one and keep all hyperparameter same as the original implementation.\footnote{\url{https://botorch.org/tutorials/turbo_1}} 

All experiments are held on Intel(R) Xeon(R) Platinum 8180 CPU @ 2.50GHz, except for TuRBO method, a NVIDIA A100 is used. No tuning is applied for all optimizers including \algo.

\subsection{Mujoco Locomotion Task} \label{appendix_mujoco}

For Mujoco locomotion benchmarks, we refer to ARS\footnote{\url{https://github.com/modestyachts/ARS}} to model the task as a sampling problem, rollout is set to $1$ for simplicity. Full optimization procedure for all 5 tasks are provided in \ref{fig:mujoco}.

\subsection{Rover Planning Task} \label{appendix_rover}

\begin{figure}[!htb]
    \centering \subfigure[Half-Cheetah(102d)]{
        \includegraphics[width = .4\linewidth]{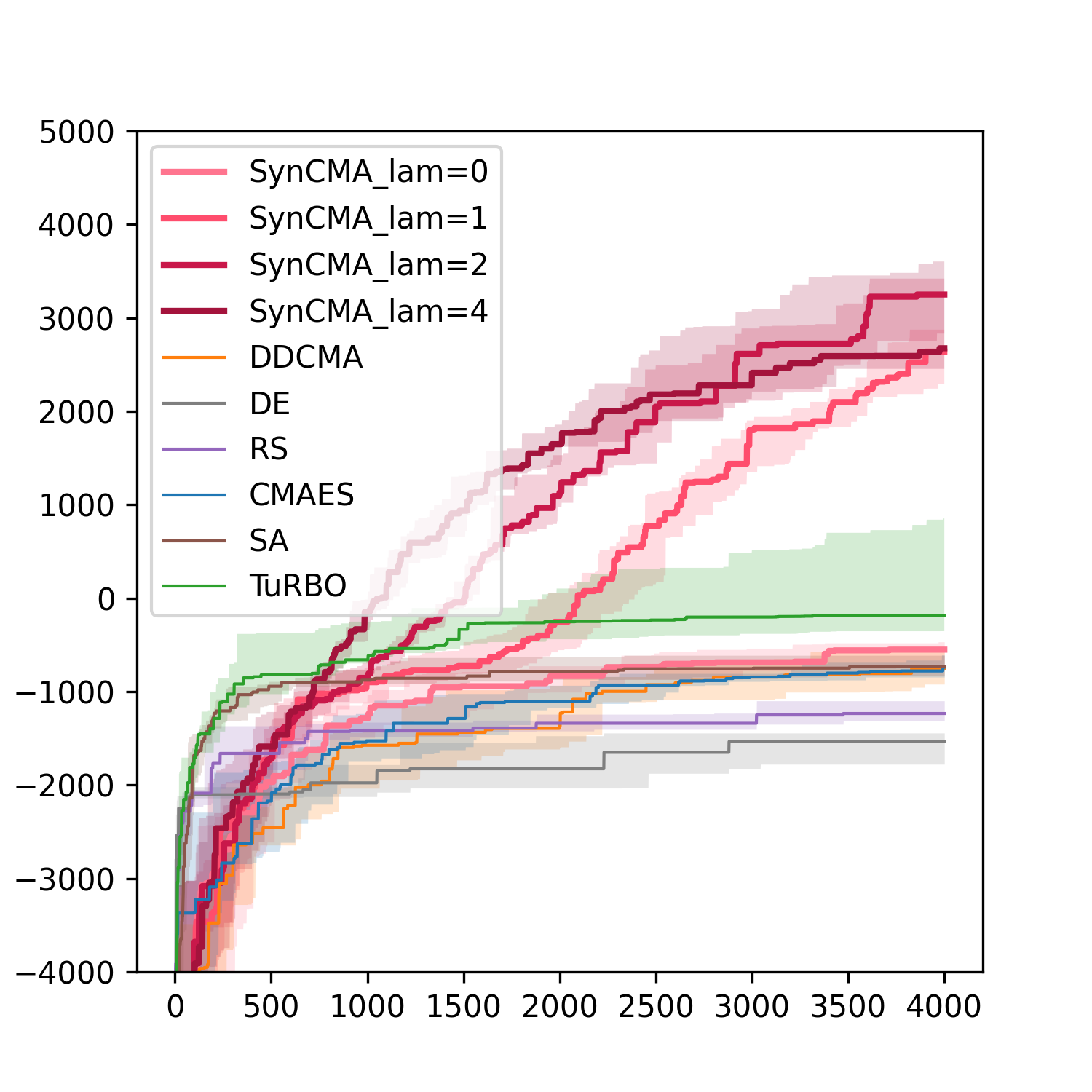}
    }
    \subfigure[Walker(102d)]{
        \includegraphics[width = .4\linewidth]{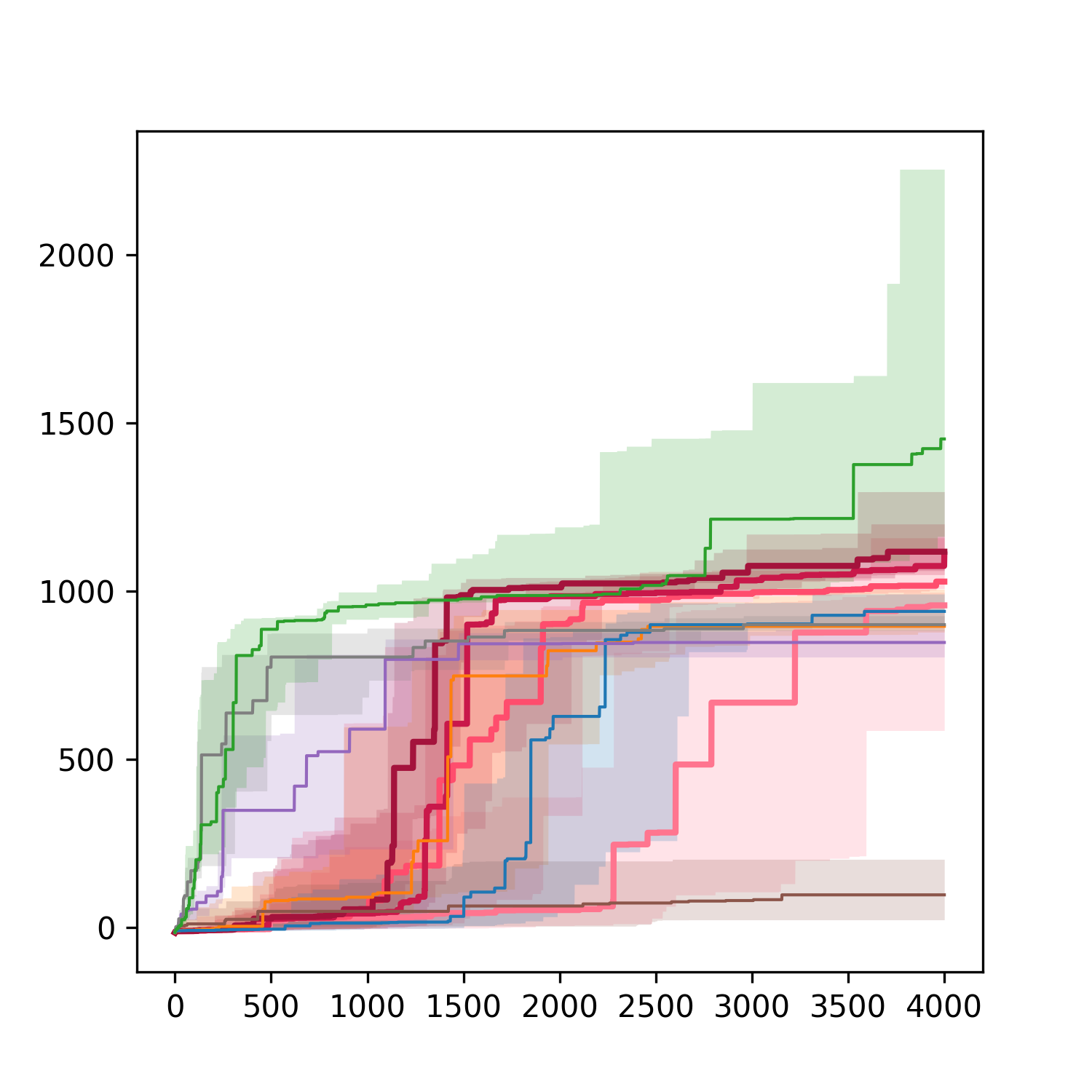}
    }
    \vfill
    \subfigure[Ant(888d)]{
        \includegraphics[width = .4\linewidth]{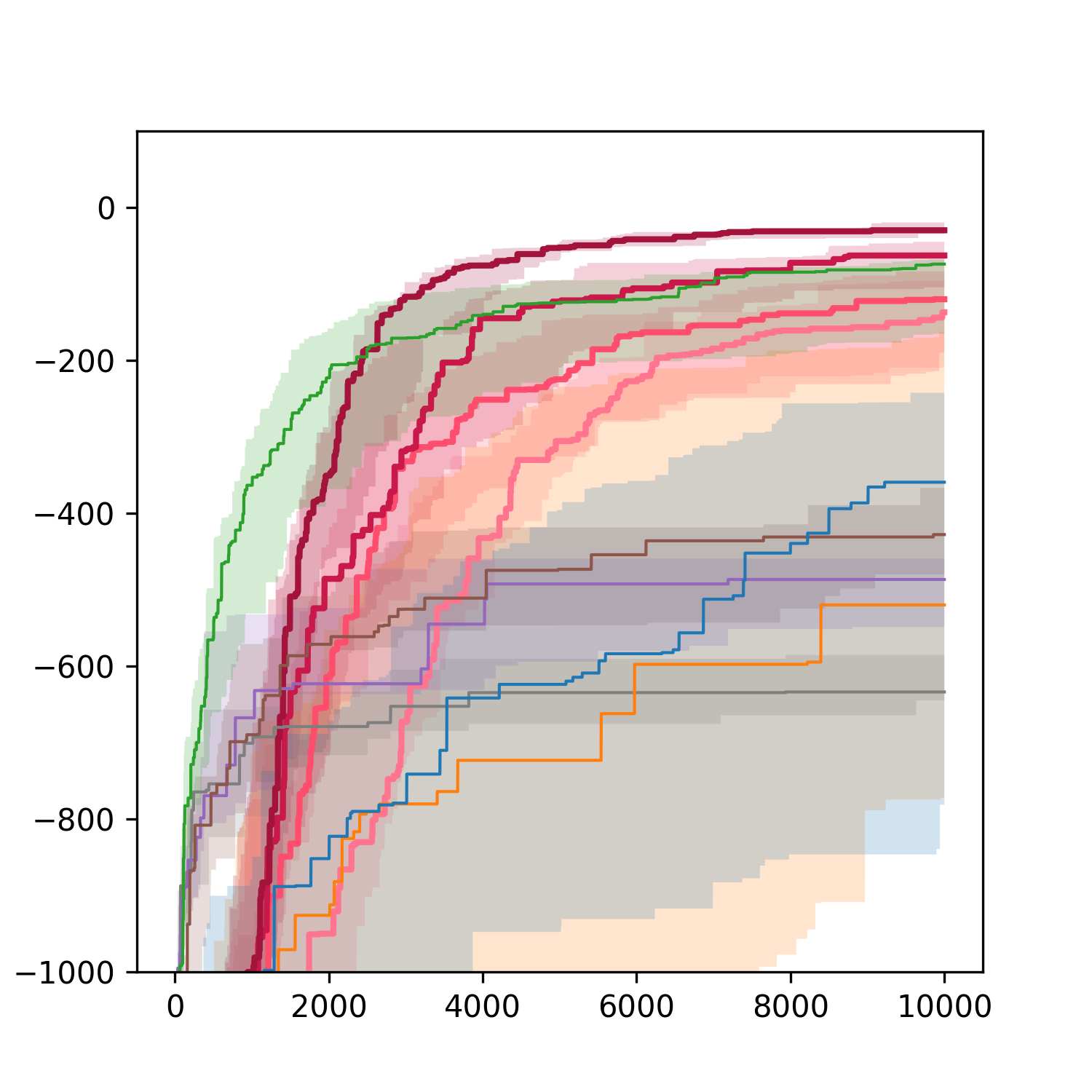}
    }
    \subfigure[Swimmer(16d)]{
        \includegraphics[width = .4\linewidth]{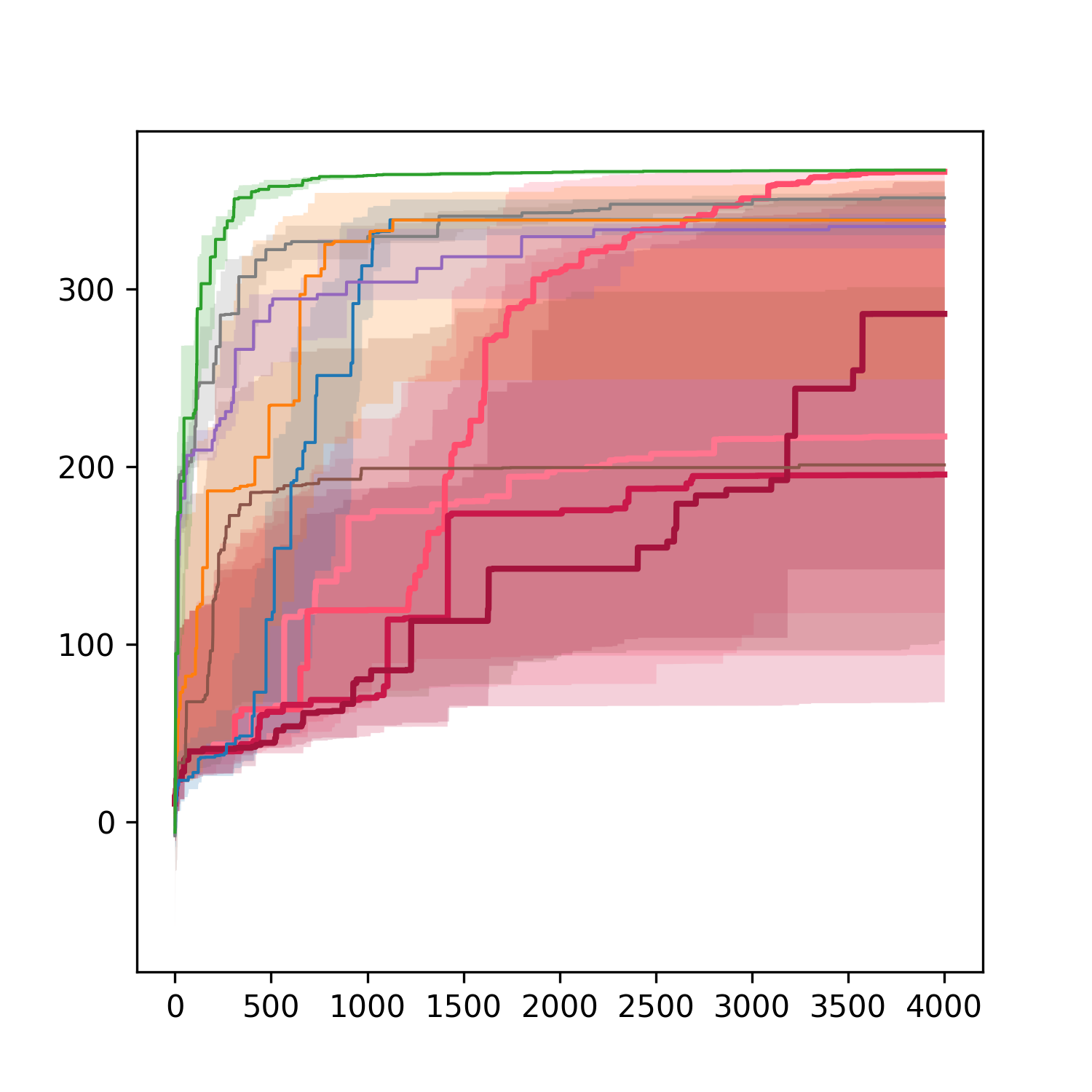}
    }
    \vfill
    \subfigure[Hopper(33d)]{
        \includegraphics[width = .4\linewidth]{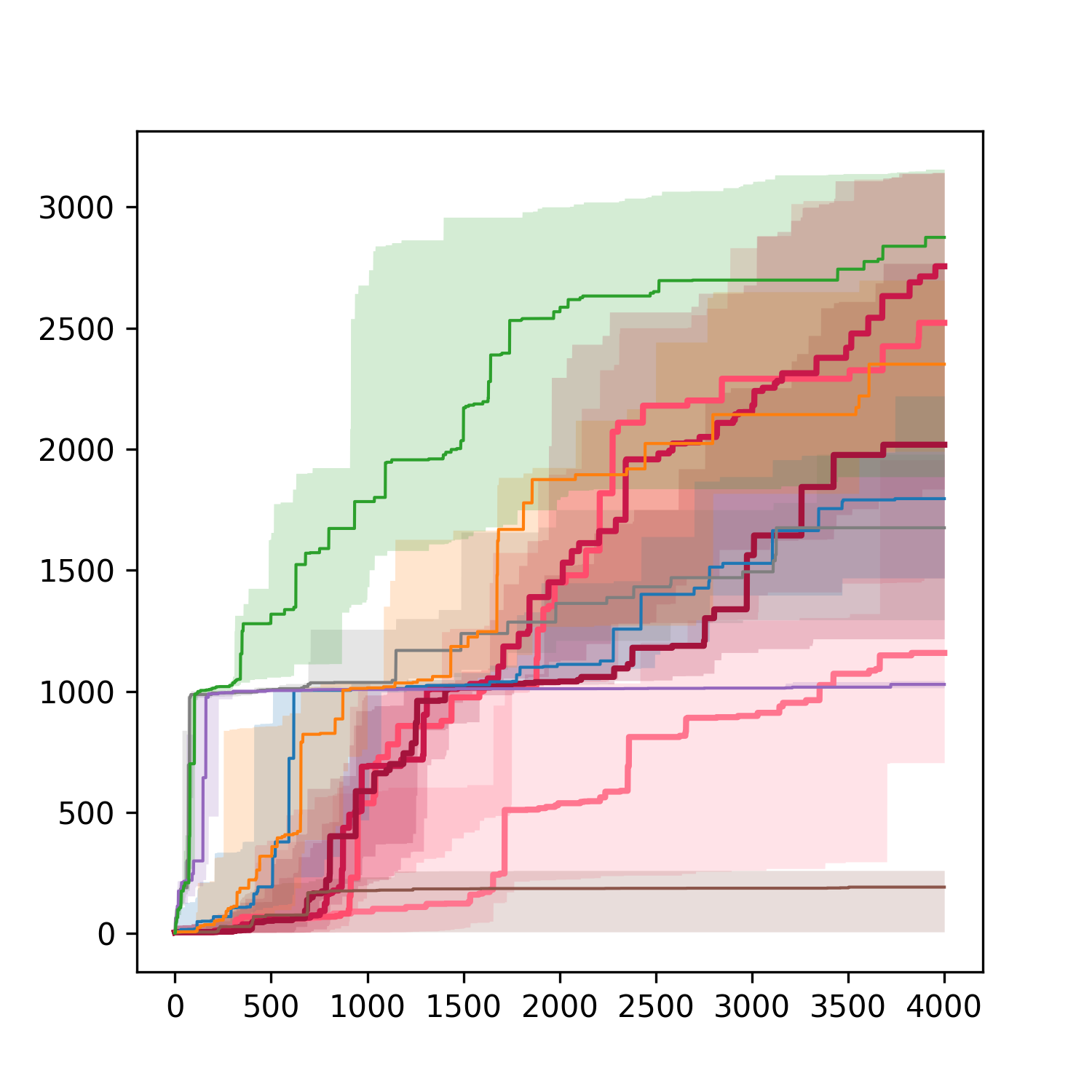}
    }
    \subfigure[Rover planning(60d)]{
        \includegraphics[width = .4\linewidth]{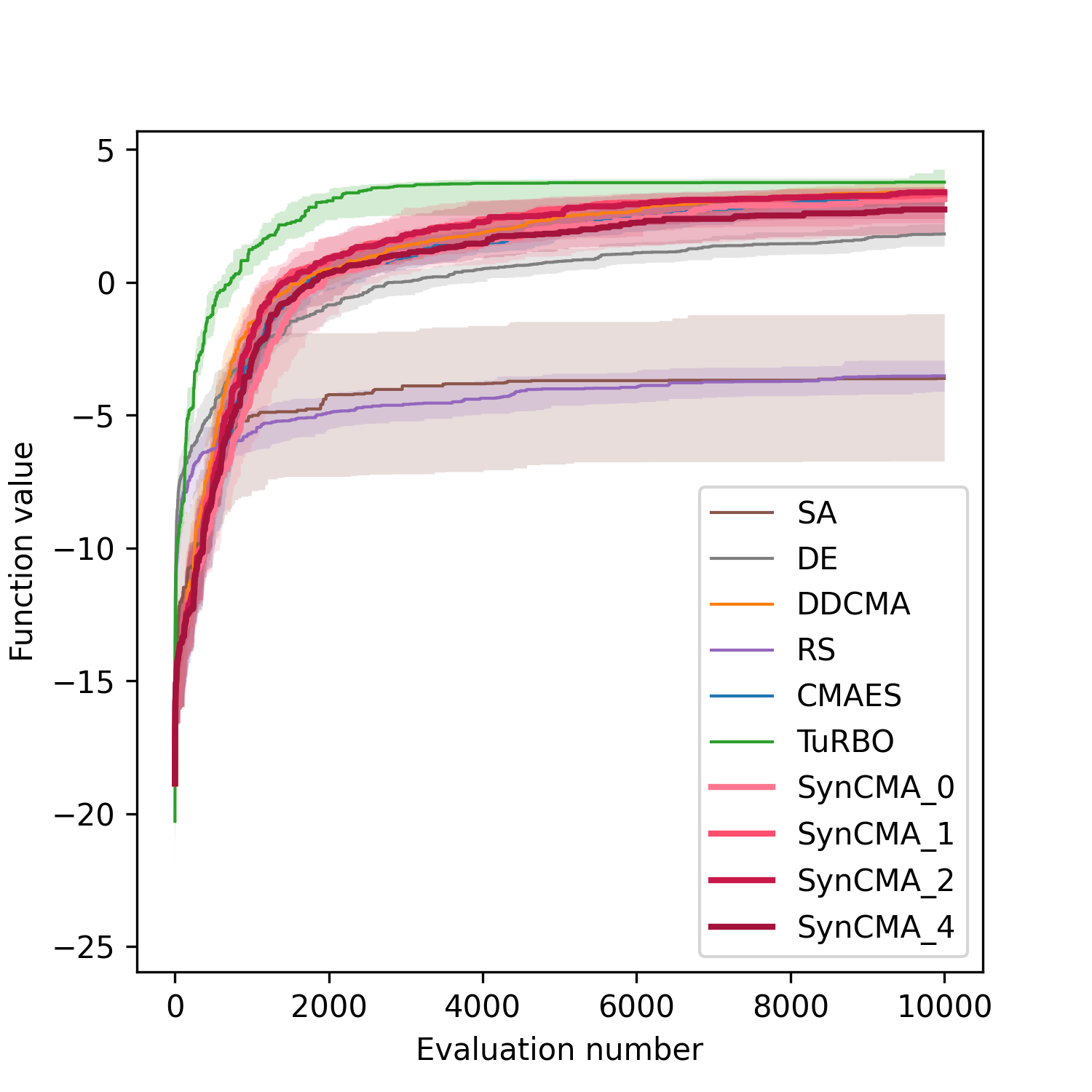}
        \label{fig:rover}
    }
    \caption{Optimization performance on 5 Mujoco locomotion tasks and the rover planing task}
    \label{fig:mujoco}
\end{figure}

For rover planning task, we refer to nevergrad\footnote{\url{https://github.com/facebookresearch/nevergrad}} to test. Optimization procedure is provided in \ref{fig:rover}.

\subsection{Synthetic Functions} \label{appendix_syn}

\begin{figure}[!htb]
    \centering
    \subfigure[Sphere]{
        \includegraphics[width = .25\linewidth]{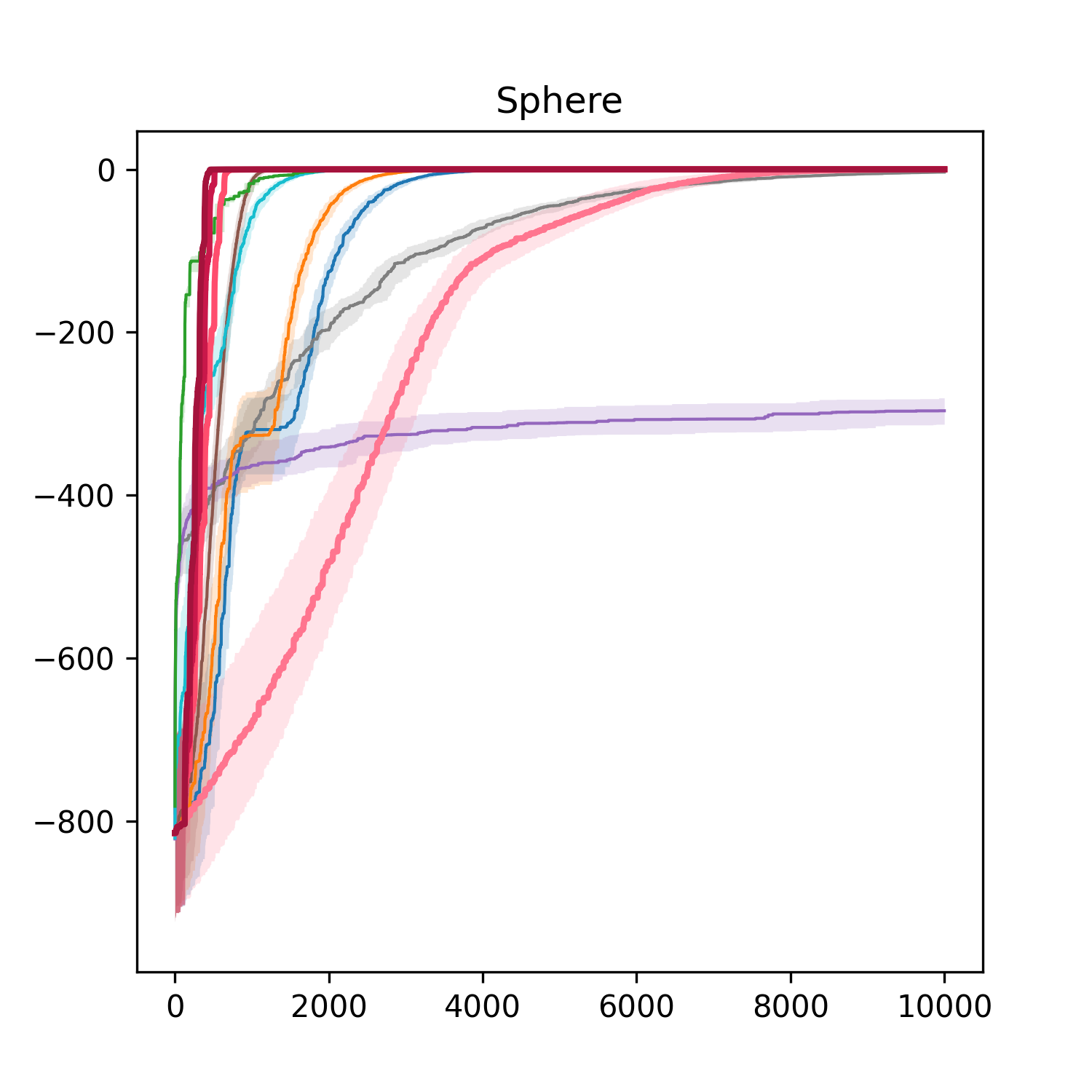}
    }
    \subfigure[Bohachevsky]{
        \includegraphics[width = .25\linewidth]{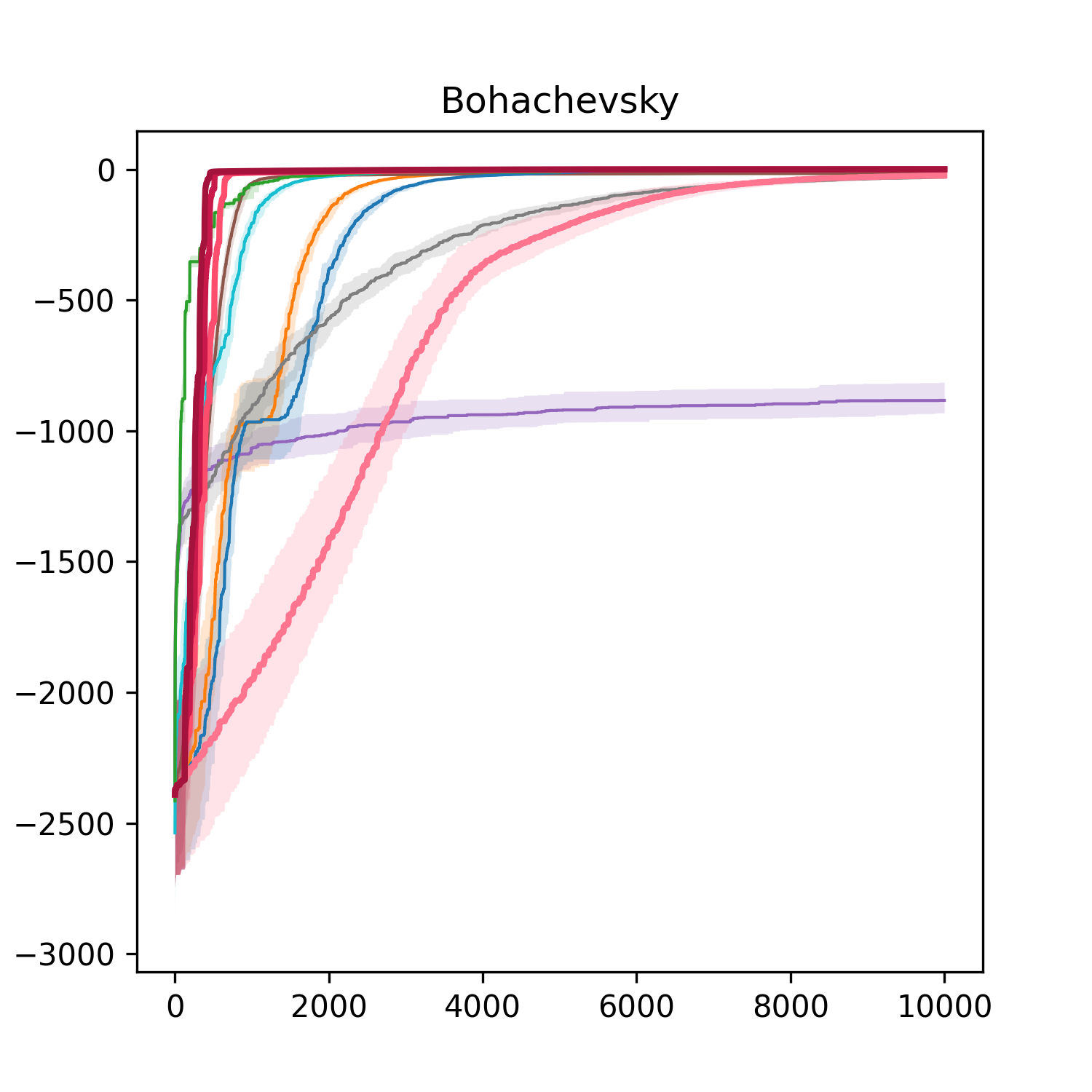}
    }
    \subfigure[LevyMontalvo]{
        \includegraphics[width = .25\linewidth]{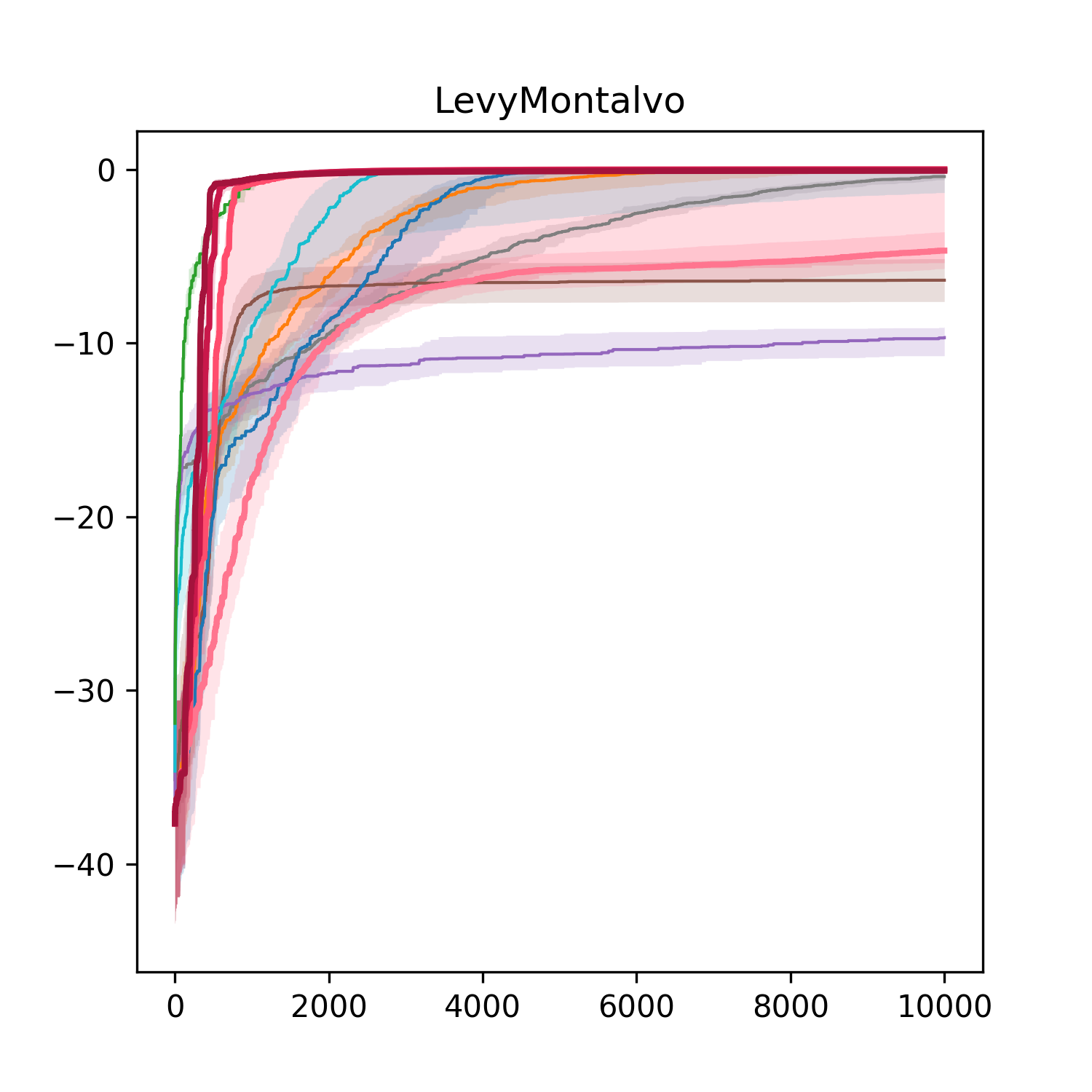}
    }
    \vfill
    \subfigure[Rastrigin]{
        \includegraphics[width = .25\linewidth]{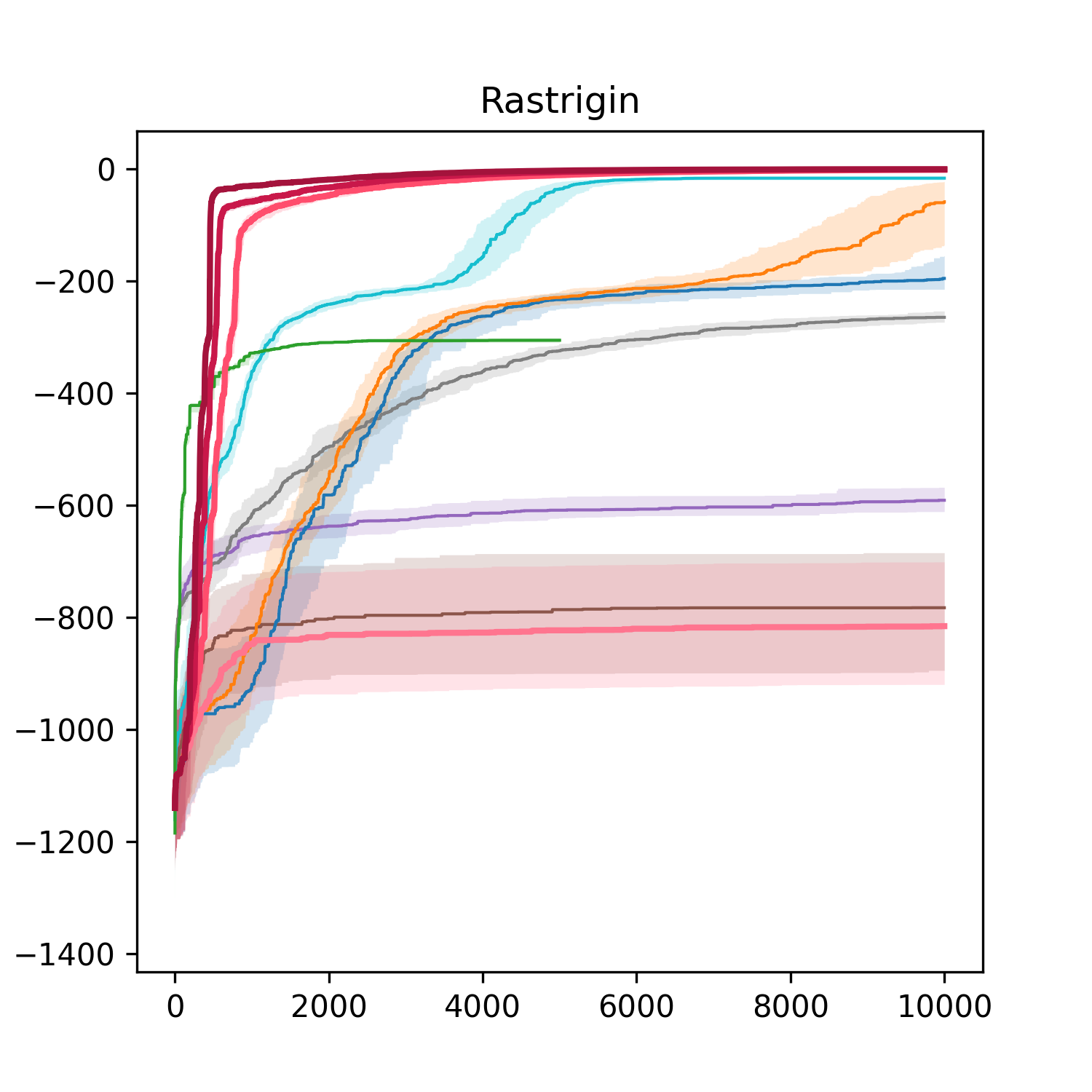}
    }
    \subfigure[Ackley]{
        \includegraphics[width = .25\linewidth]{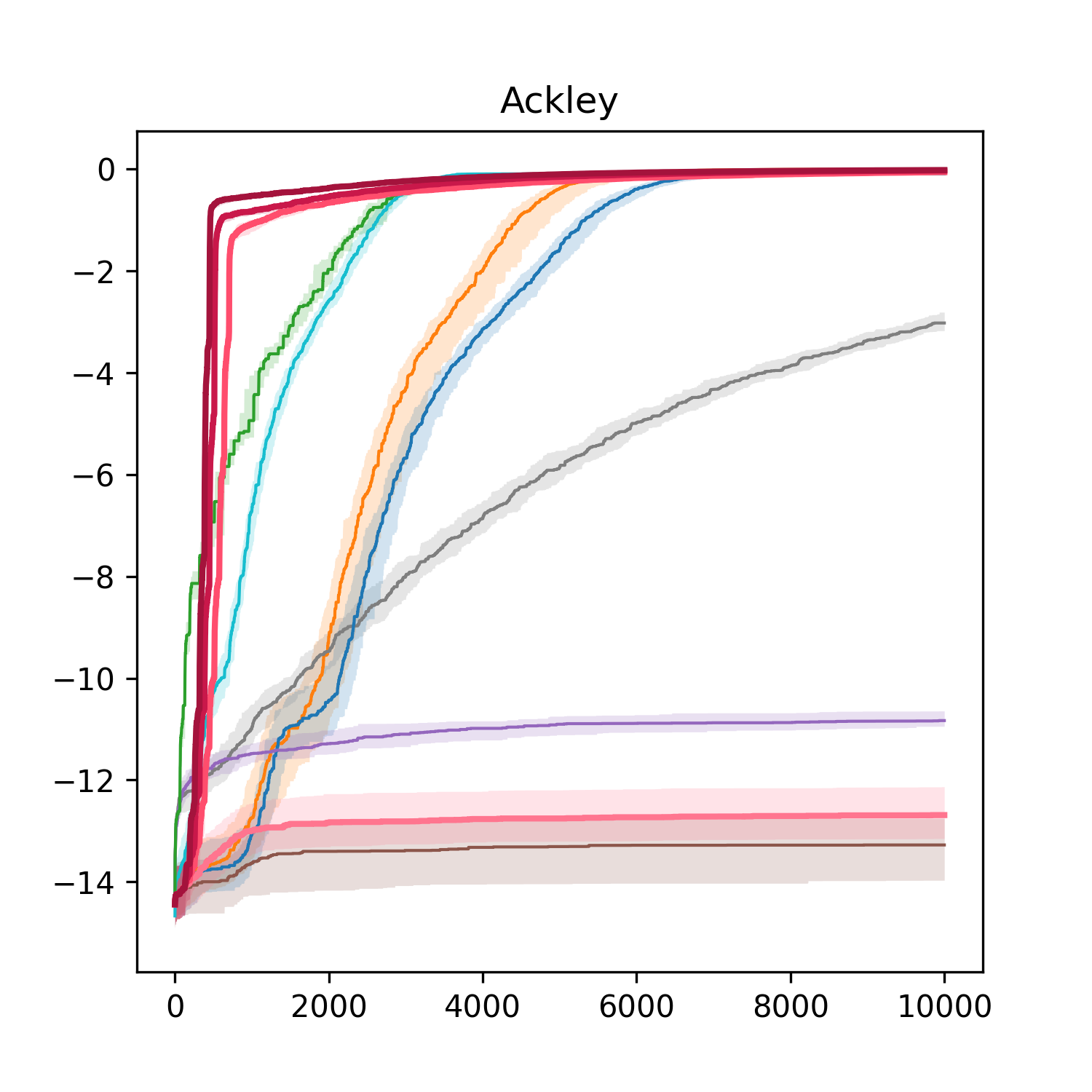}
    }
    \subfigure[Schaffer]{
        \includegraphics[width = .25\linewidth]{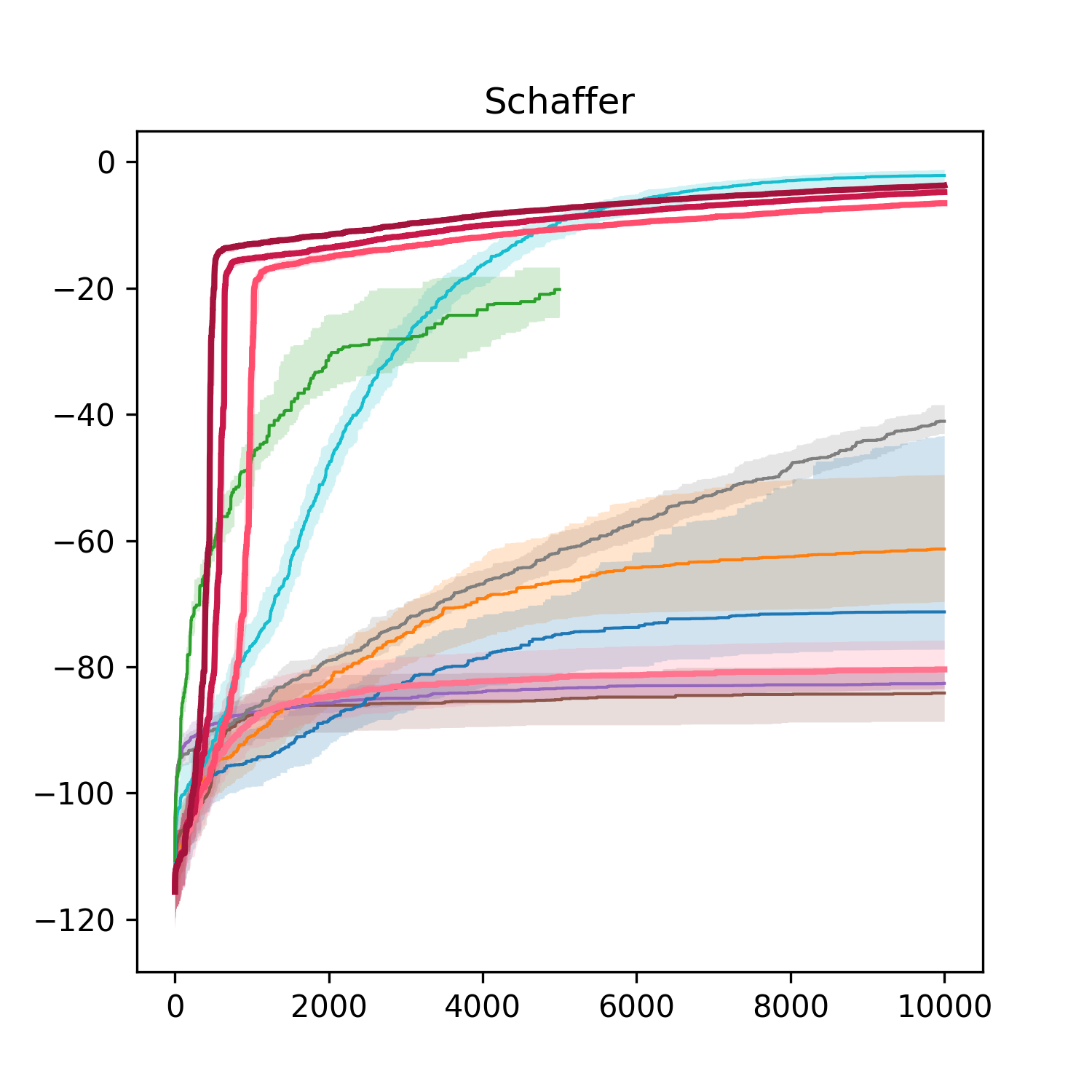}
    }
    \vfill
    \subfigure[Discus]{
        \includegraphics[width = .25\linewidth]{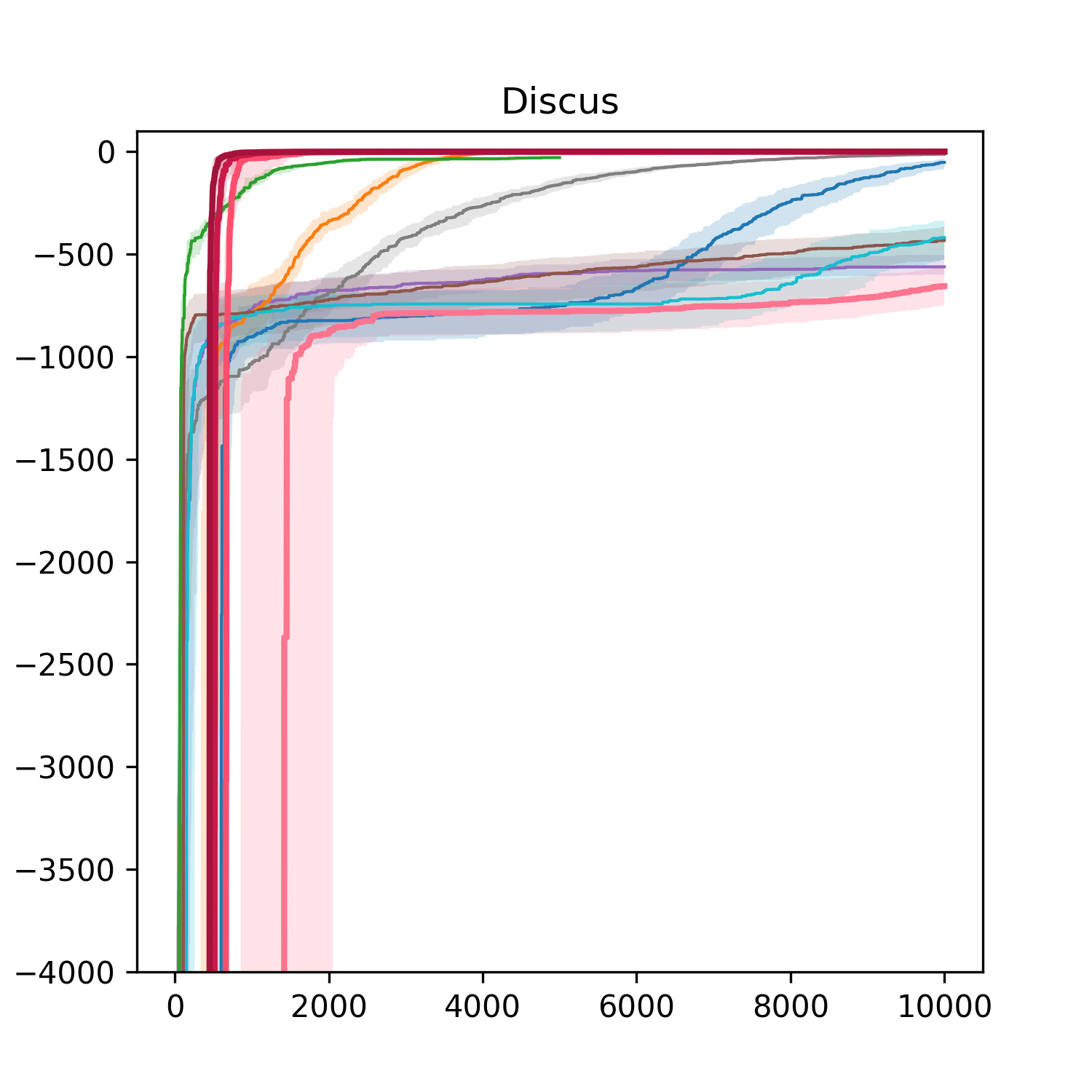}
    }
    \subfigure[Schwefel]{
        \includegraphics[width = .25\linewidth]{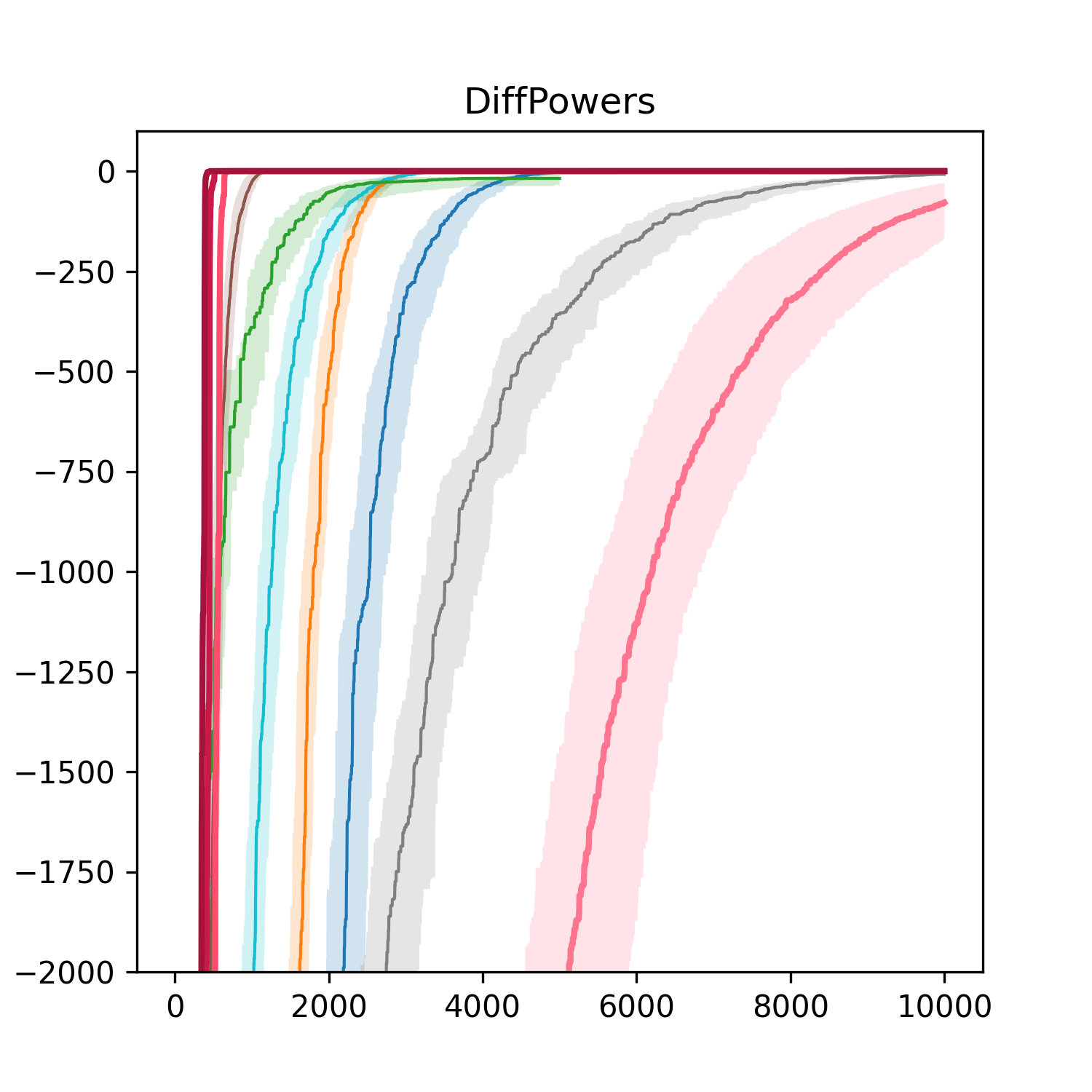}
    }
    \subfigure[DiffPowers]{
        \includegraphics[width = .25\linewidth]{./figs/synthetic/bs_DiffPowers_32D_64_noLegend.png}
    }
    \vfill
    \subfigure[Rosenbrock]{
        \includegraphics[width = .25\linewidth]{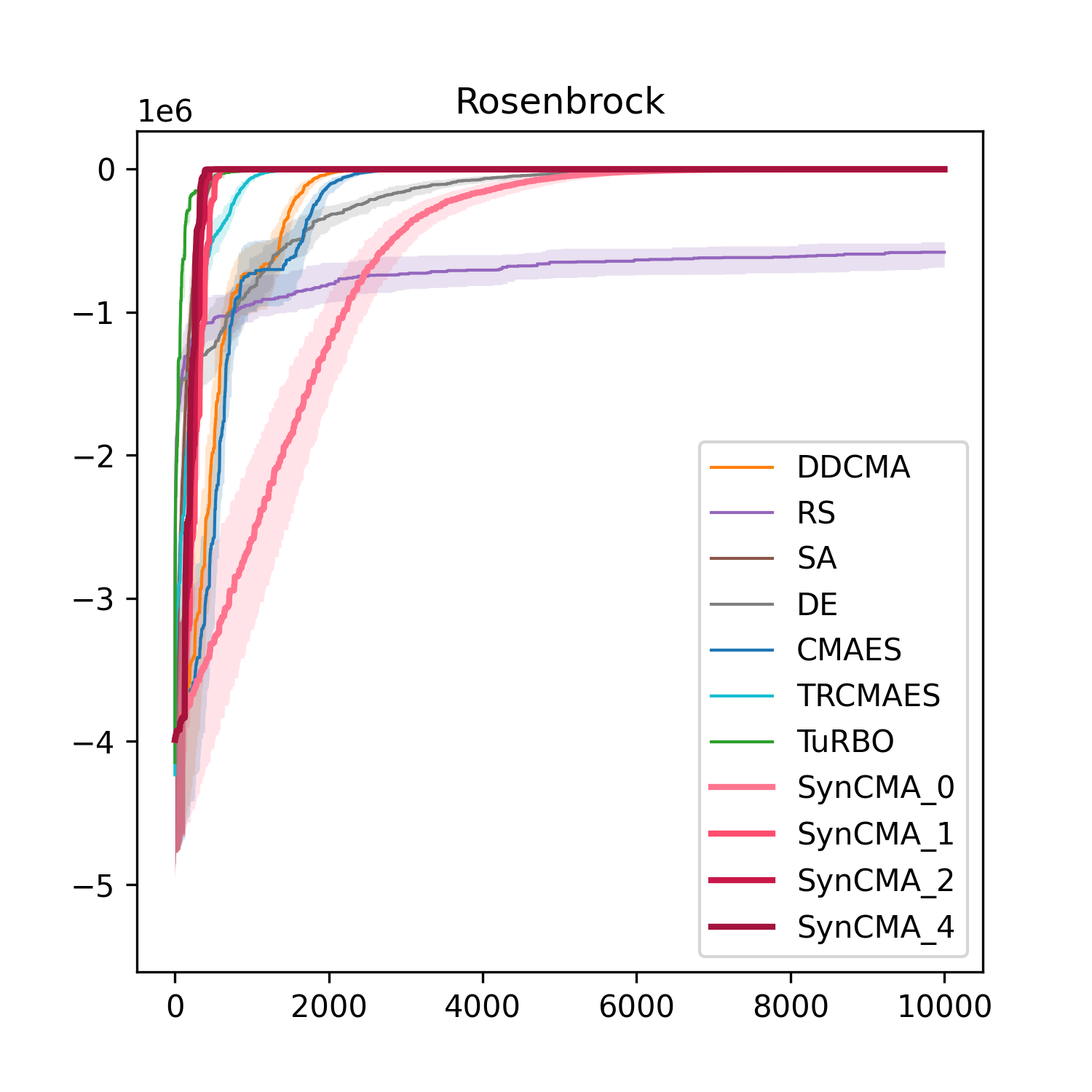}
    }
    \caption{Optimization procedure in 10 tests function with dimension $n = 32$ over 100 trails with 10000 evaluations.}
    \vspace{10pt}
    \label{fig:syn_32}
\end{figure}

\begin{figure}[!htb]
        \centering\subfigure[Sphere]{
        \includegraphics[width = .25\linewidth]{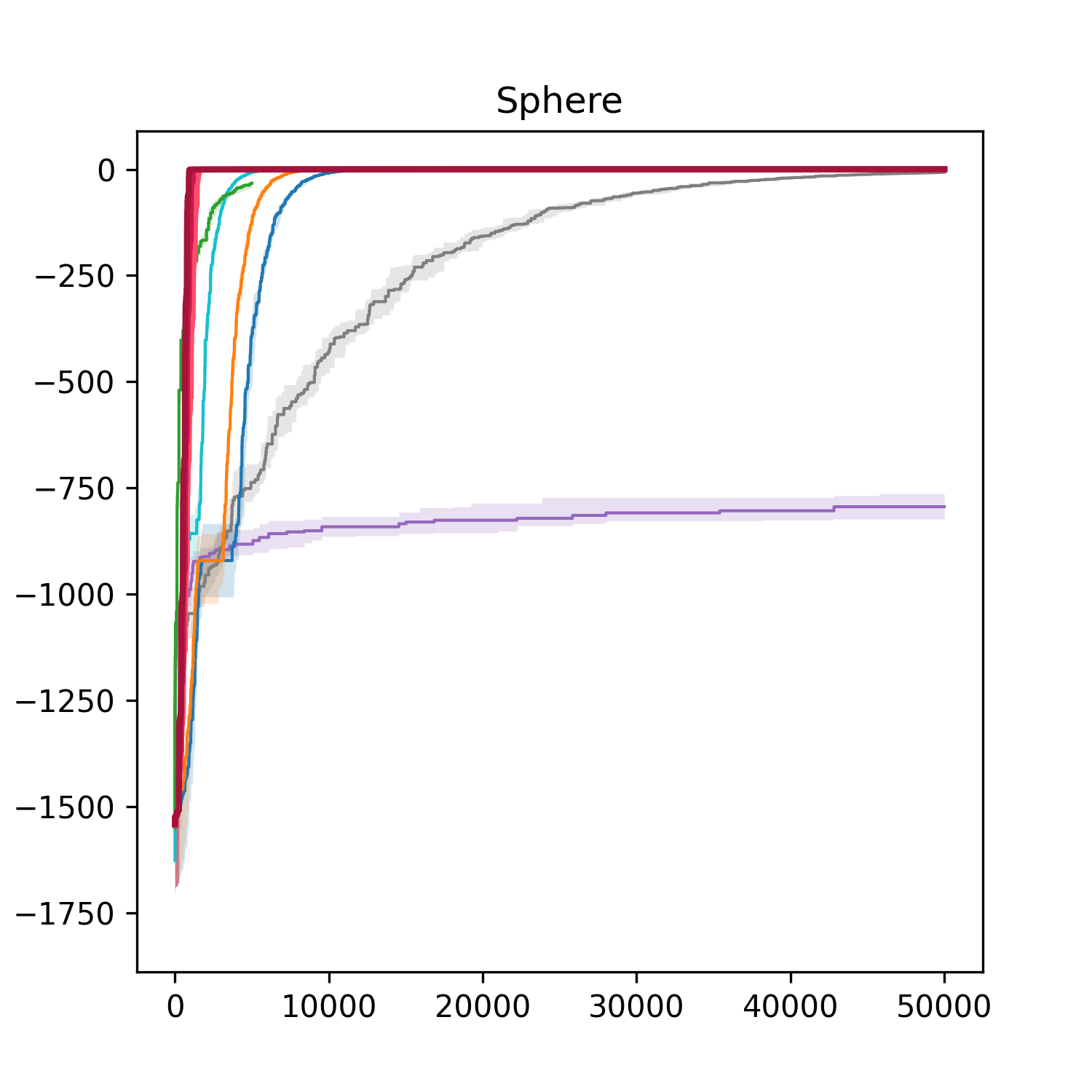}
    }
    \subfigure[Bohachevsky]{
        \includegraphics[width = .25\linewidth]{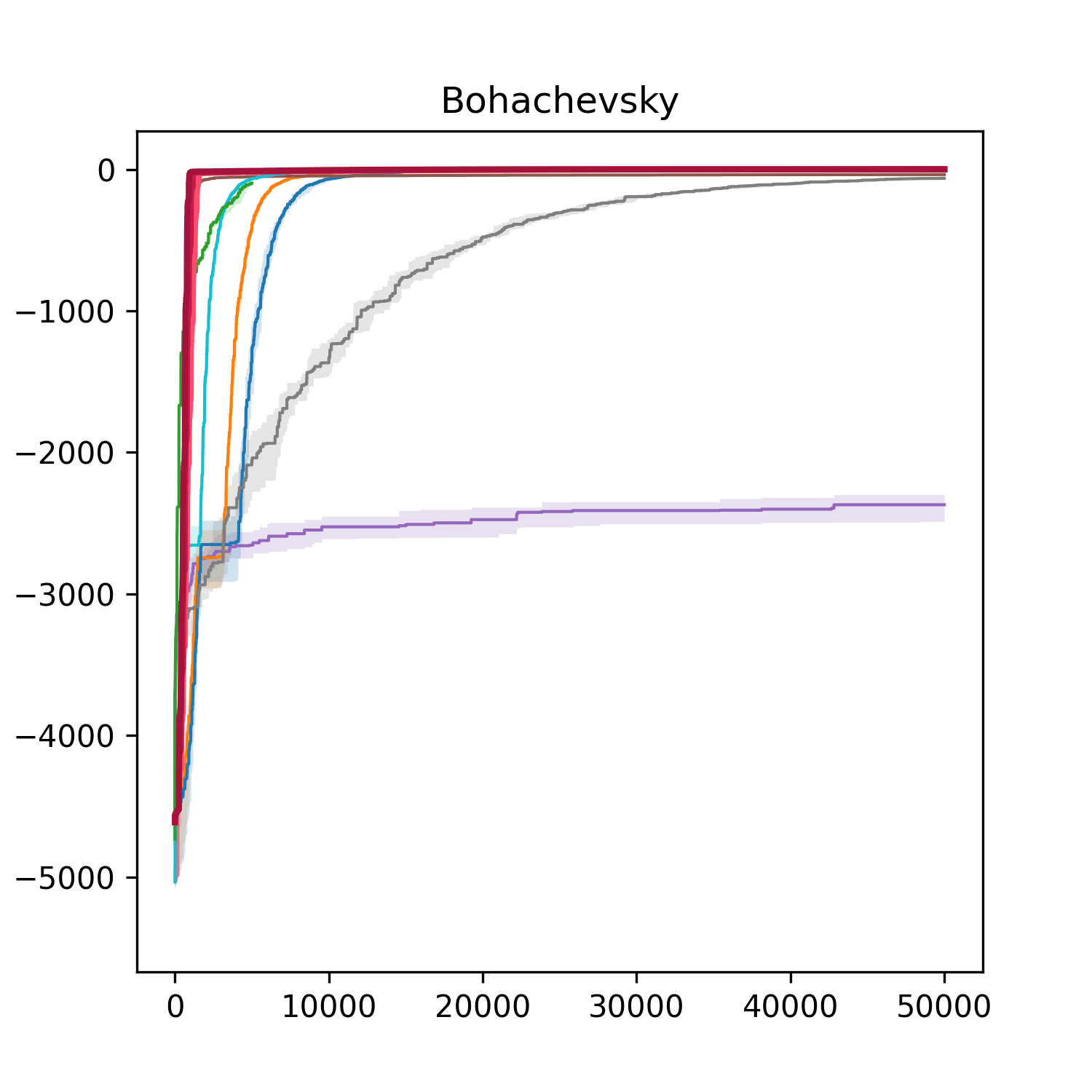}
    }
    \subfigure[LevyMontalvo]{
        \includegraphics[width = .25\linewidth]{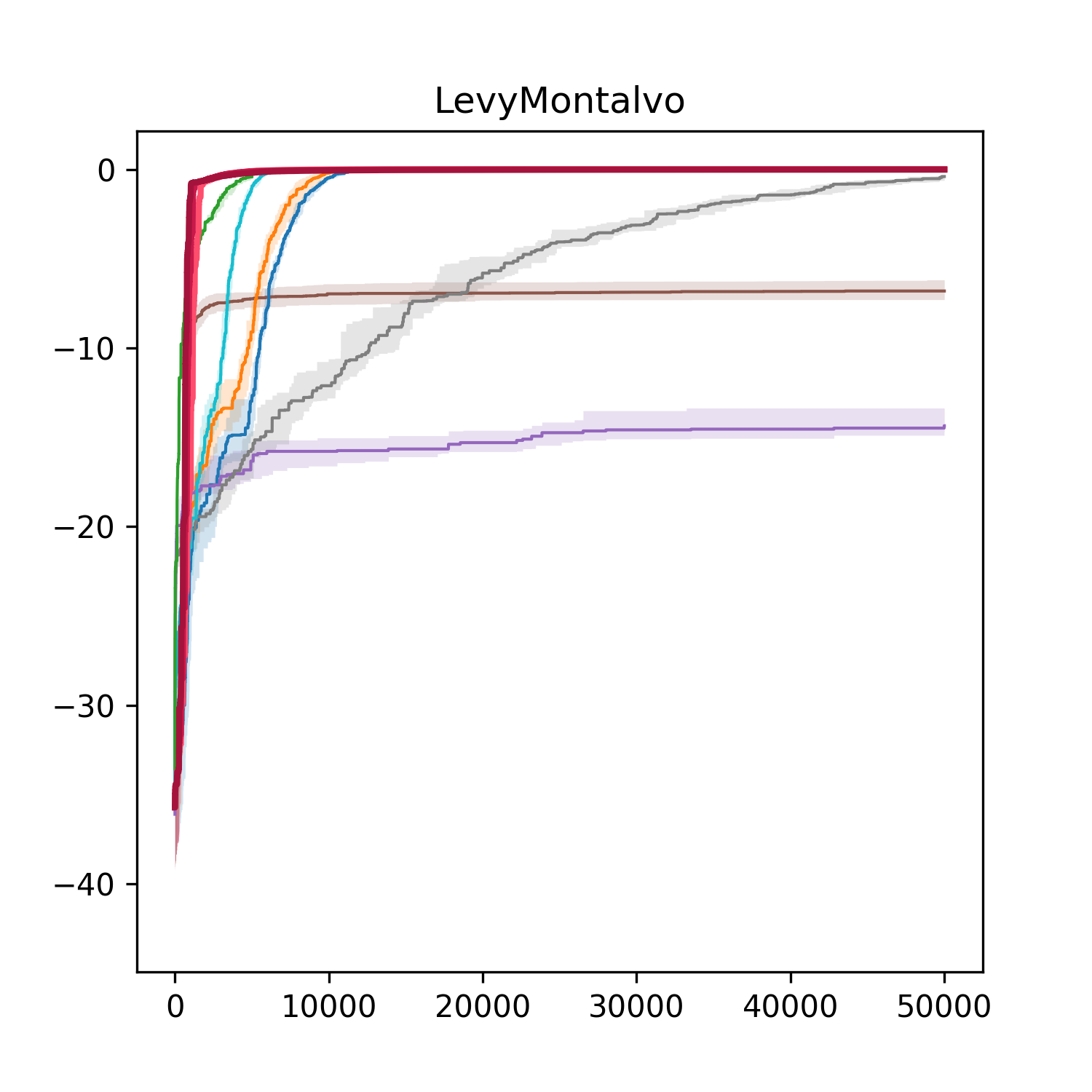}
    }
    \vfill
    \subfigure[Rastrigin]{
        \includegraphics[width = .25\linewidth]{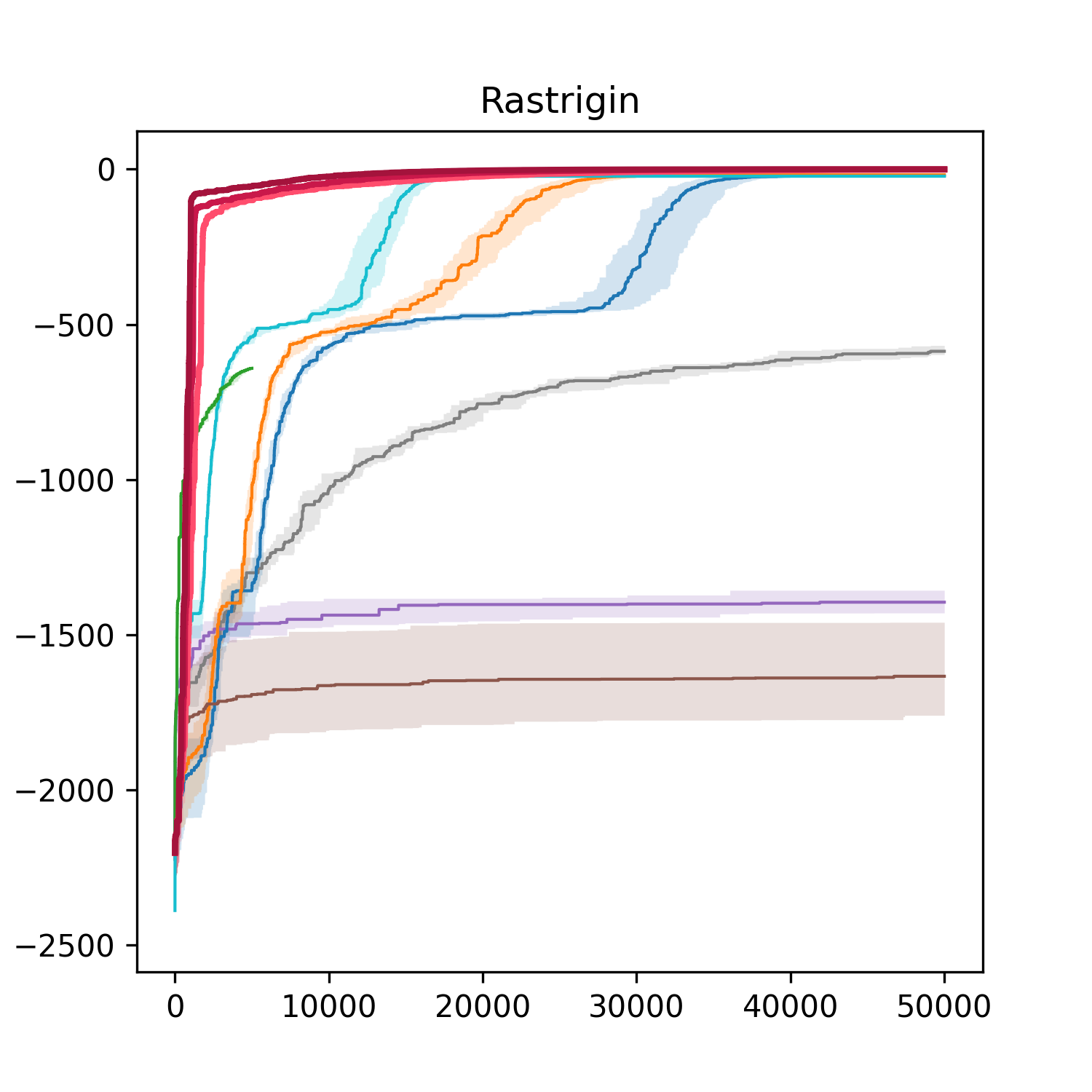}
    }
    \subfigure[Ackley]{
        \includegraphics[width = .25\linewidth]{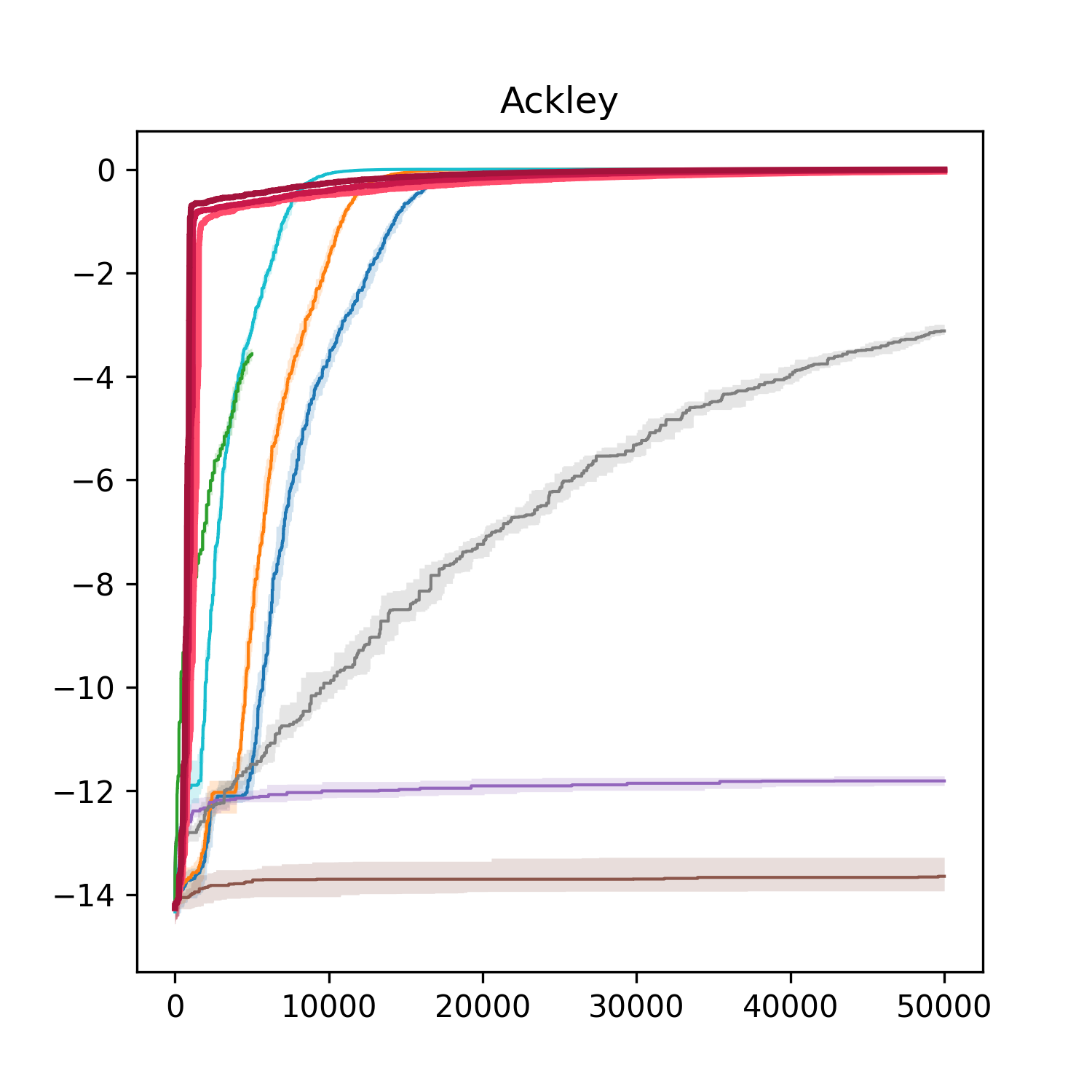}
    }
    \subfigure[Schaffer]{
        \includegraphics[width = .25\linewidth]{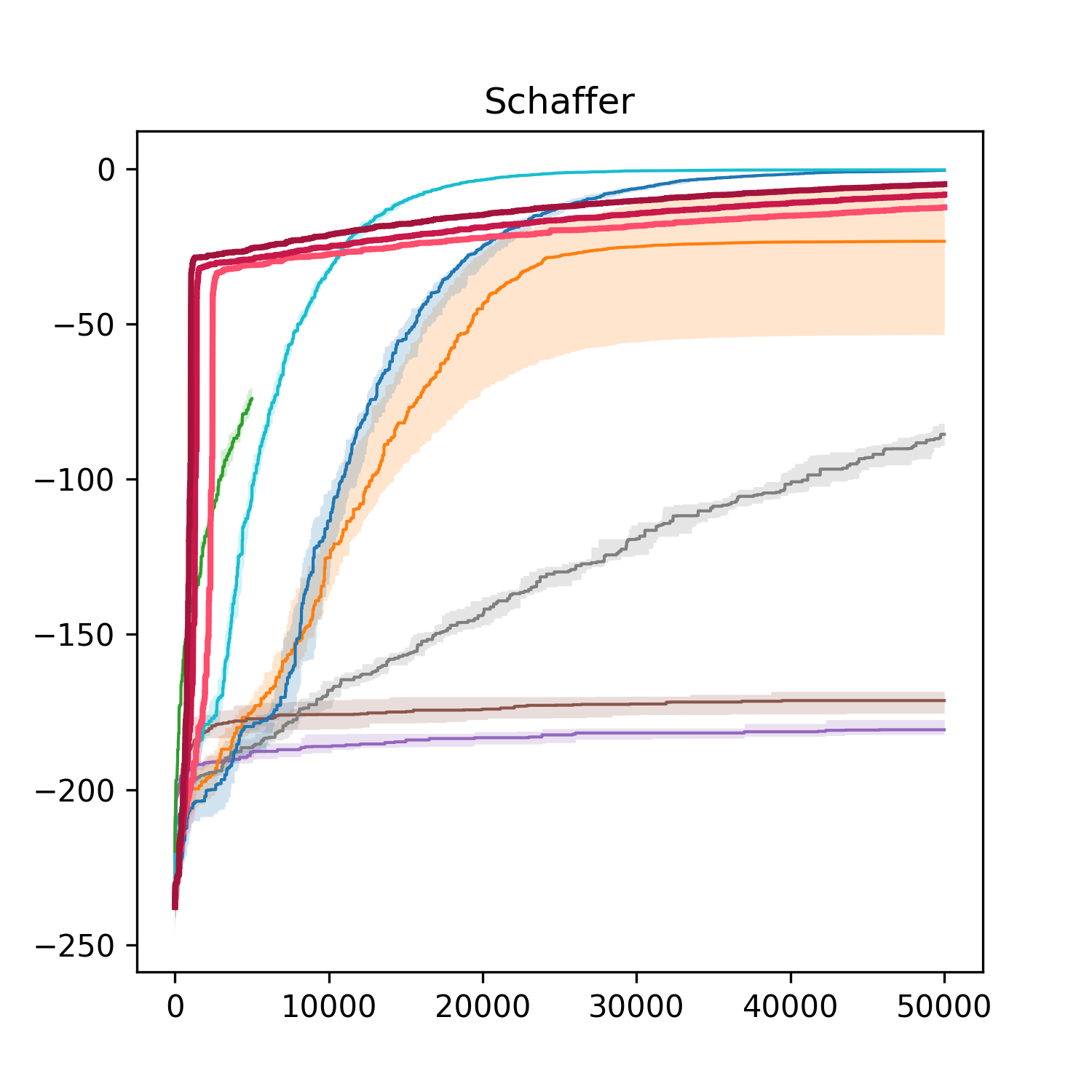}
    }
    \vfill
    \subfigure[Discus]{
        \includegraphics[width = .25\linewidth]{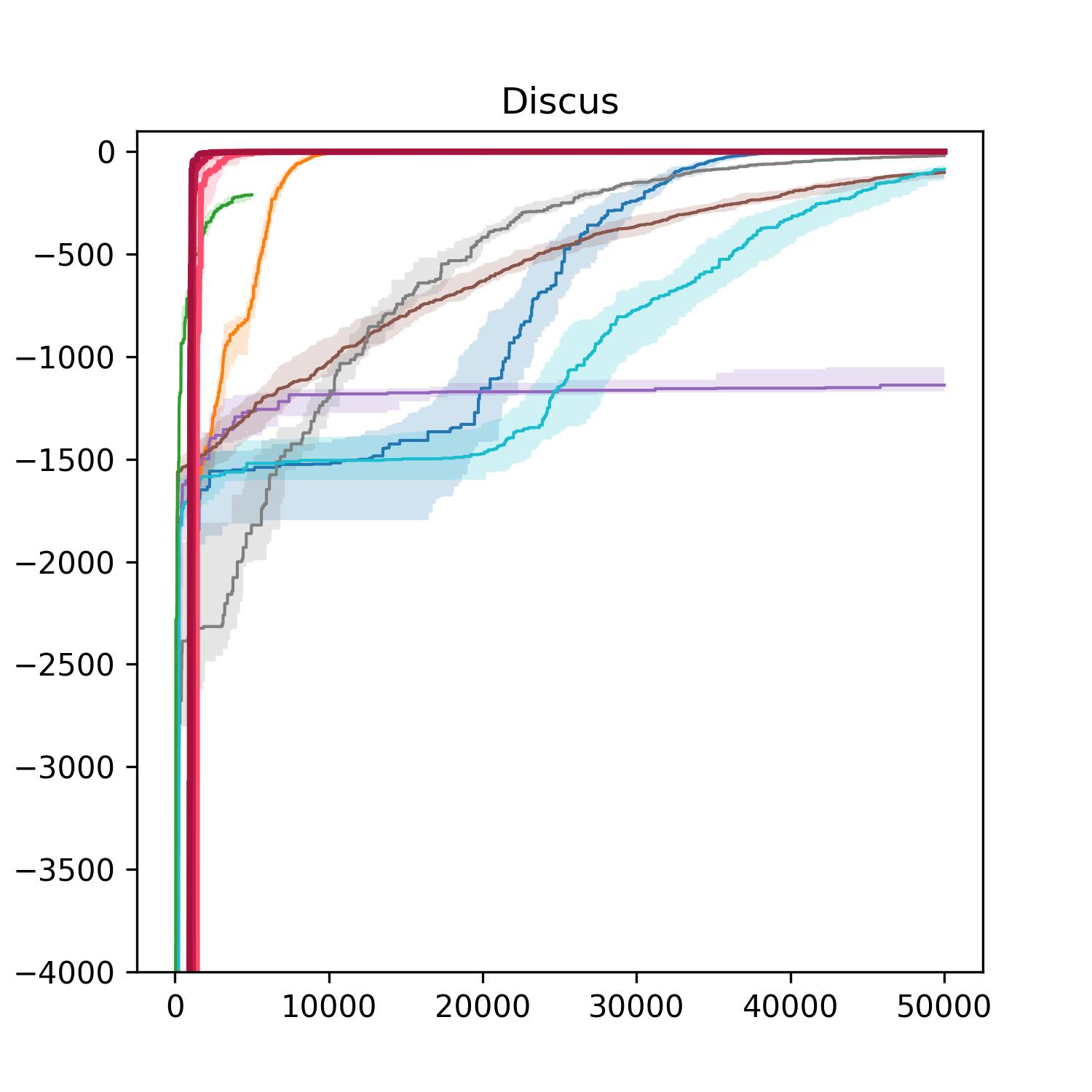}
    }
    \subfigure[Schwefel]{
        \includegraphics[width = .25\linewidth]{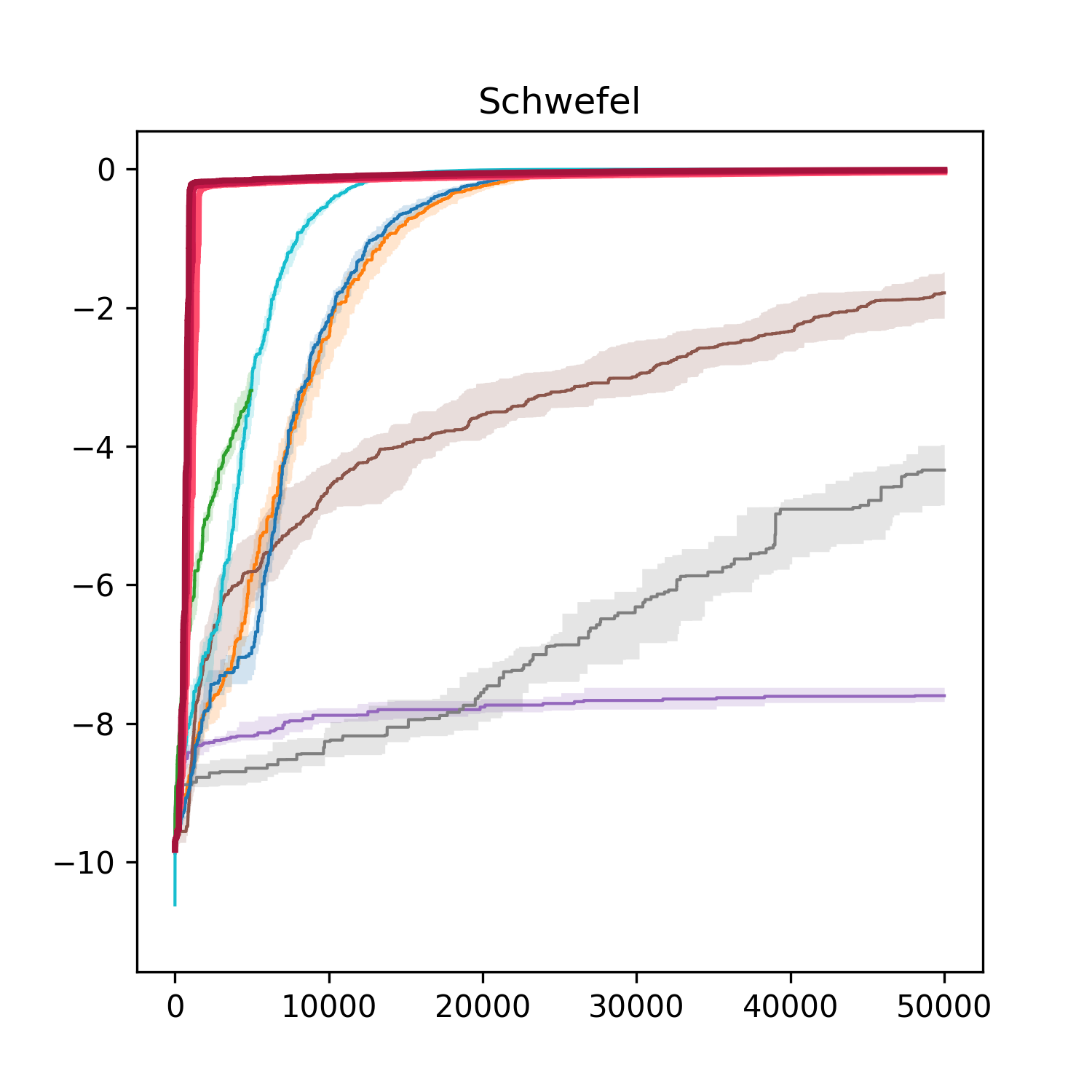}
    }
    \subfigure[DiffPowers]{
        \includegraphics[width = .25\linewidth]{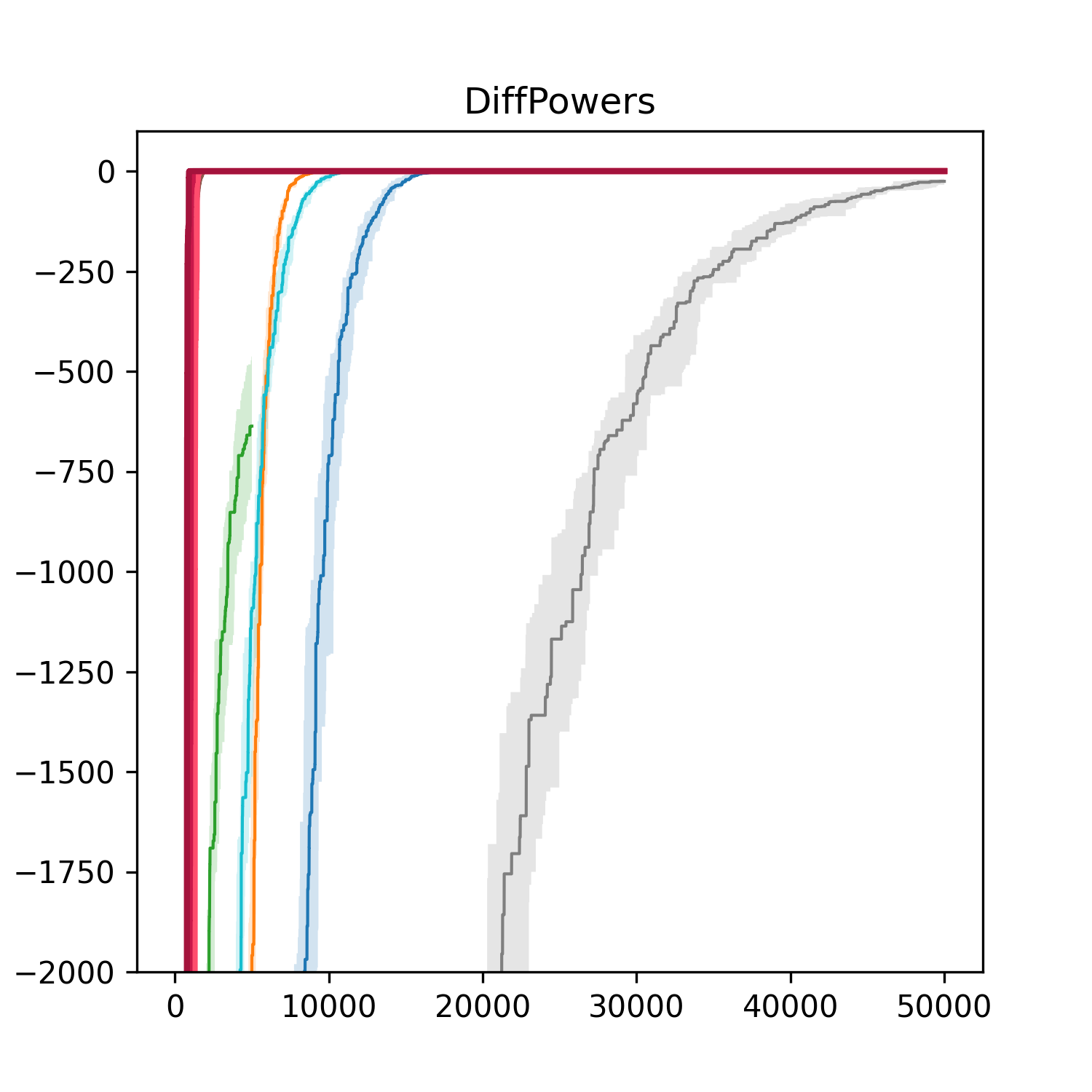}
    }
    \vfill
    \subfigure[Rosenbrock]{
        \includegraphics[width = .25\linewidth]{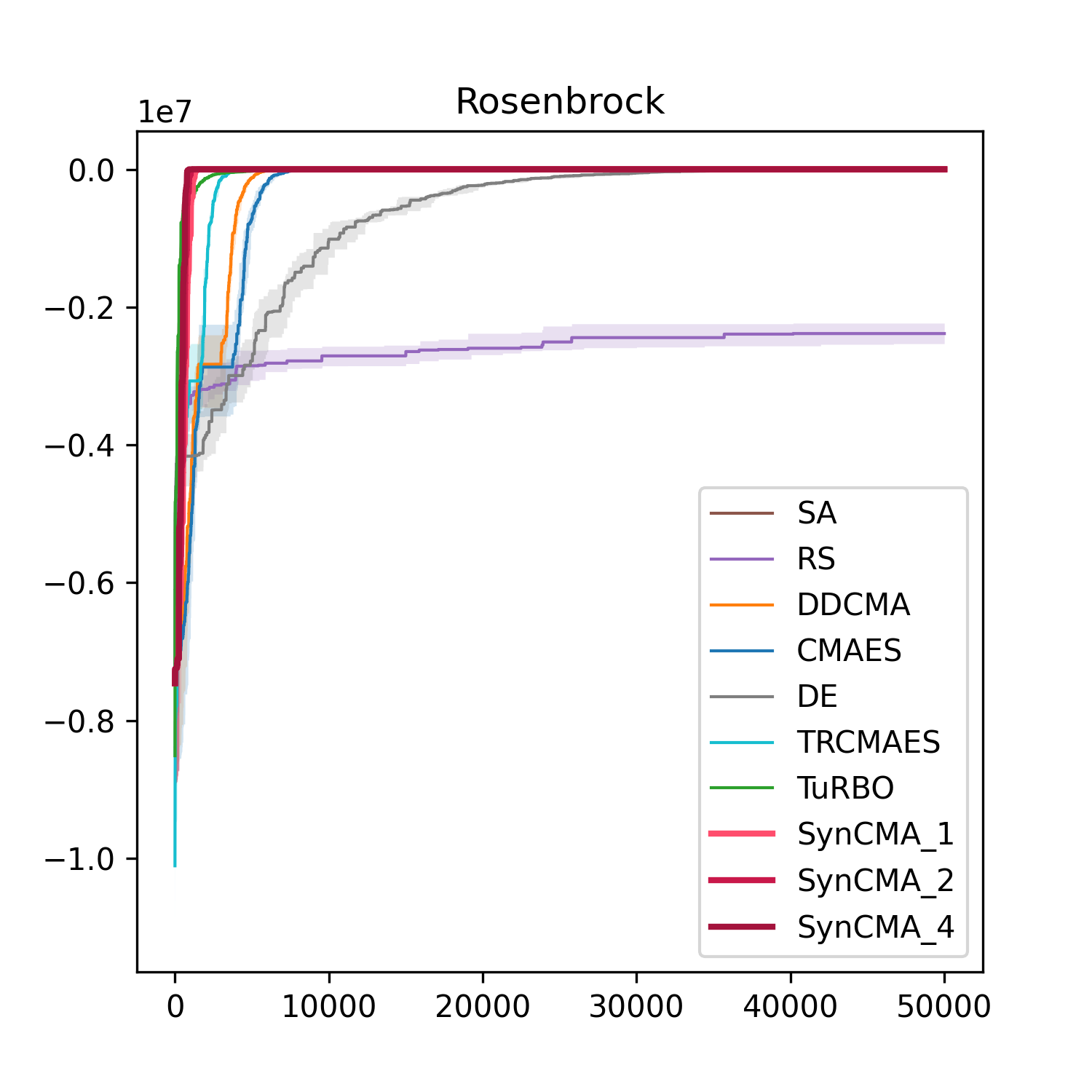}
    }
    
    \caption{Optimization procedure in 10 tests function with dimension $n = 64$ over 100 trails with 50000 evaluations.}
    \vspace{10pt}
    \label{fig:syn_64}
\end{figure}

\begin{figure}[!htb]
    \centering
    \subfigure[Sphere]{
        \includegraphics[width = .25\linewidth]{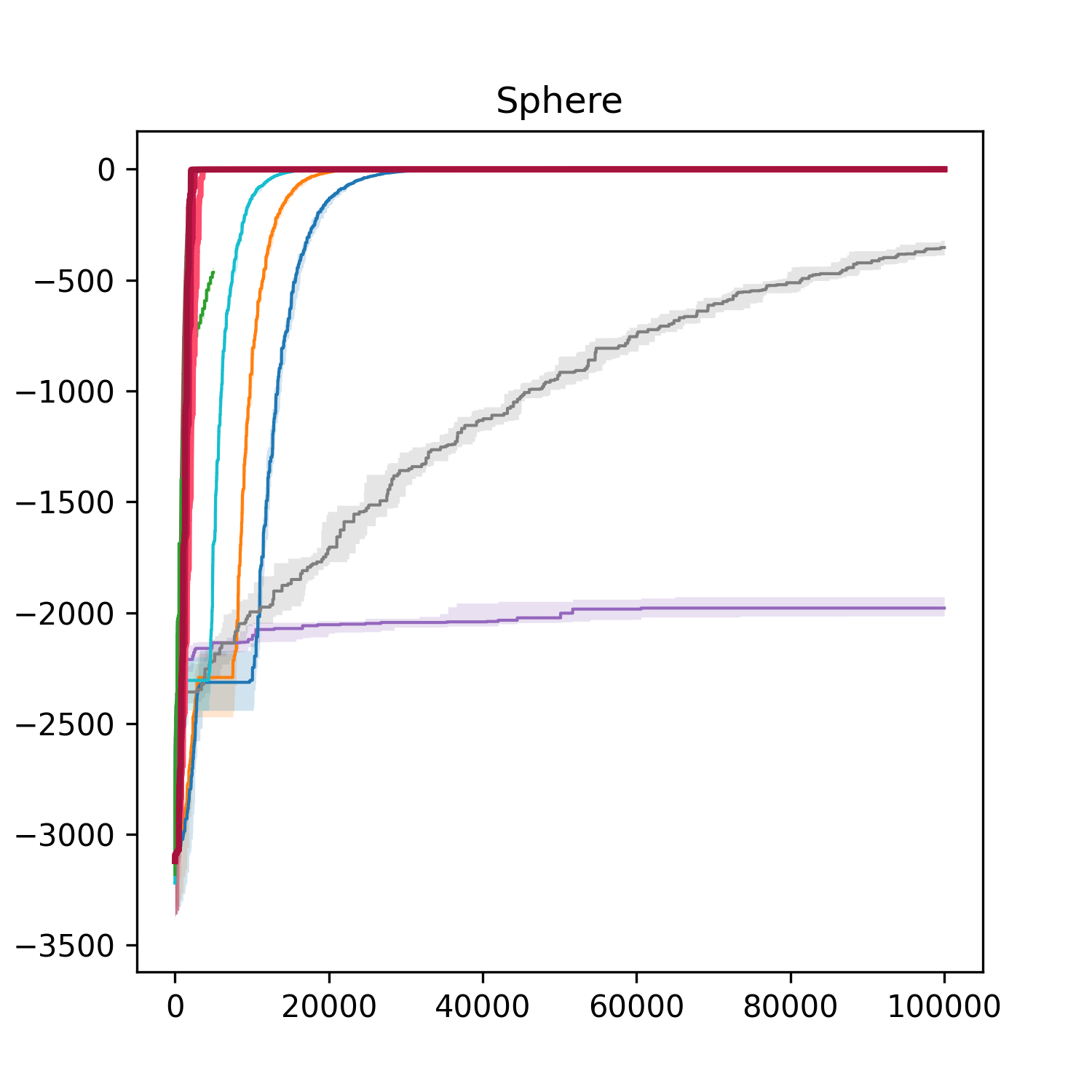}
    }
    \subfigure[Bohachevsky]{
        \includegraphics[width = .25\linewidth]{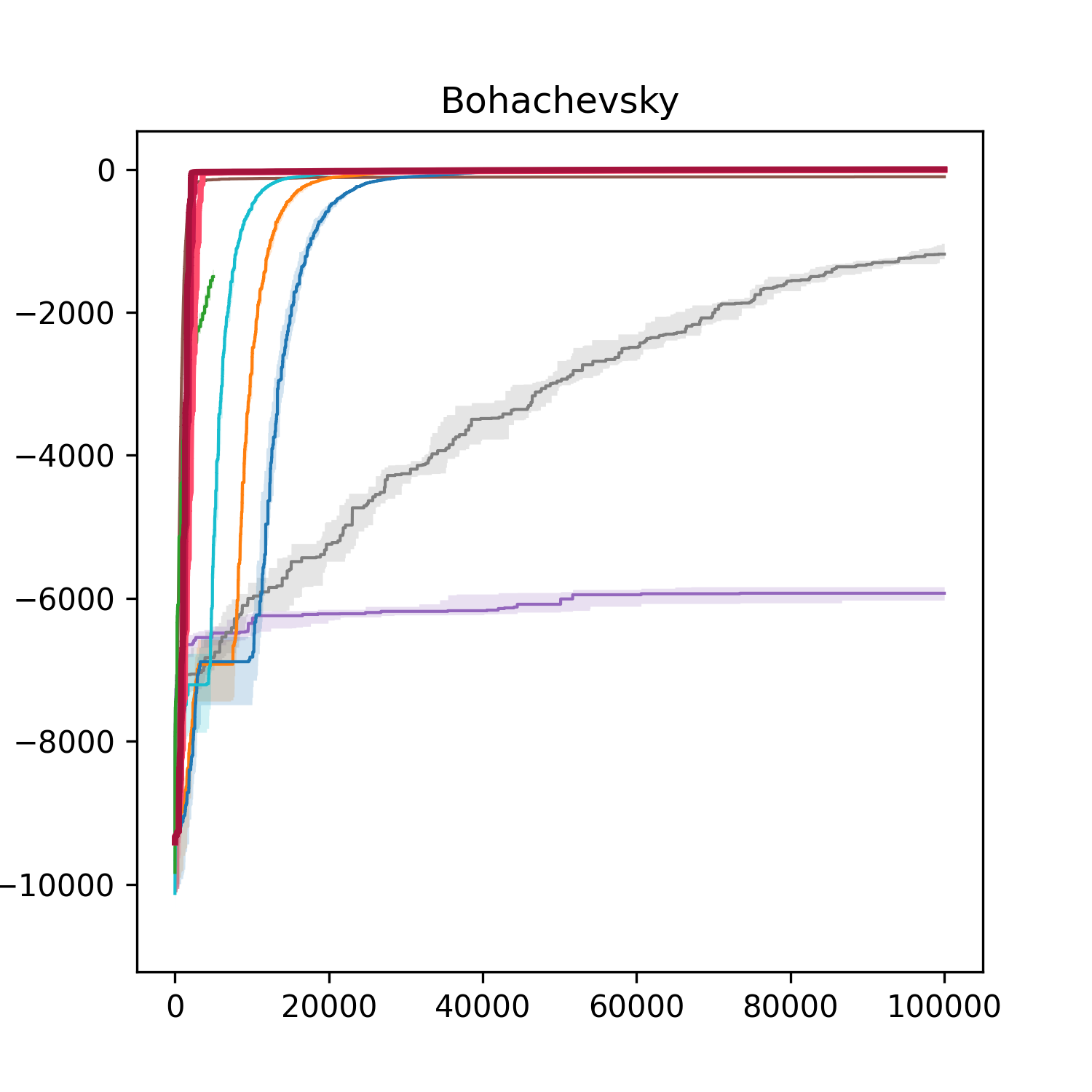}
    }
    \subfigure[LevyMontalvo]{
        \includegraphics[width = .25\linewidth]{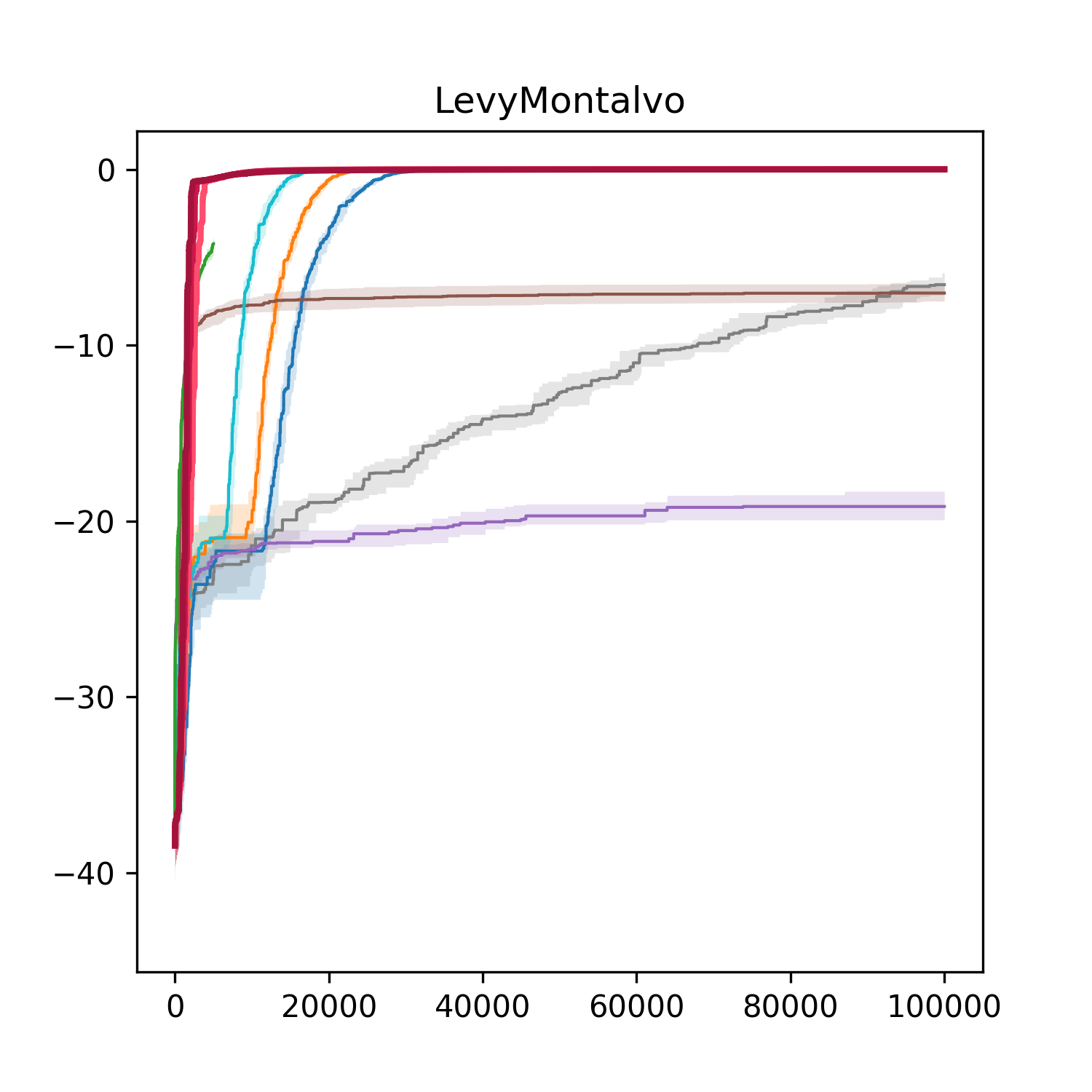}
    }
    \vfill
    \subfigure[Rastrigin]{
        \includegraphics[width = .25\linewidth]{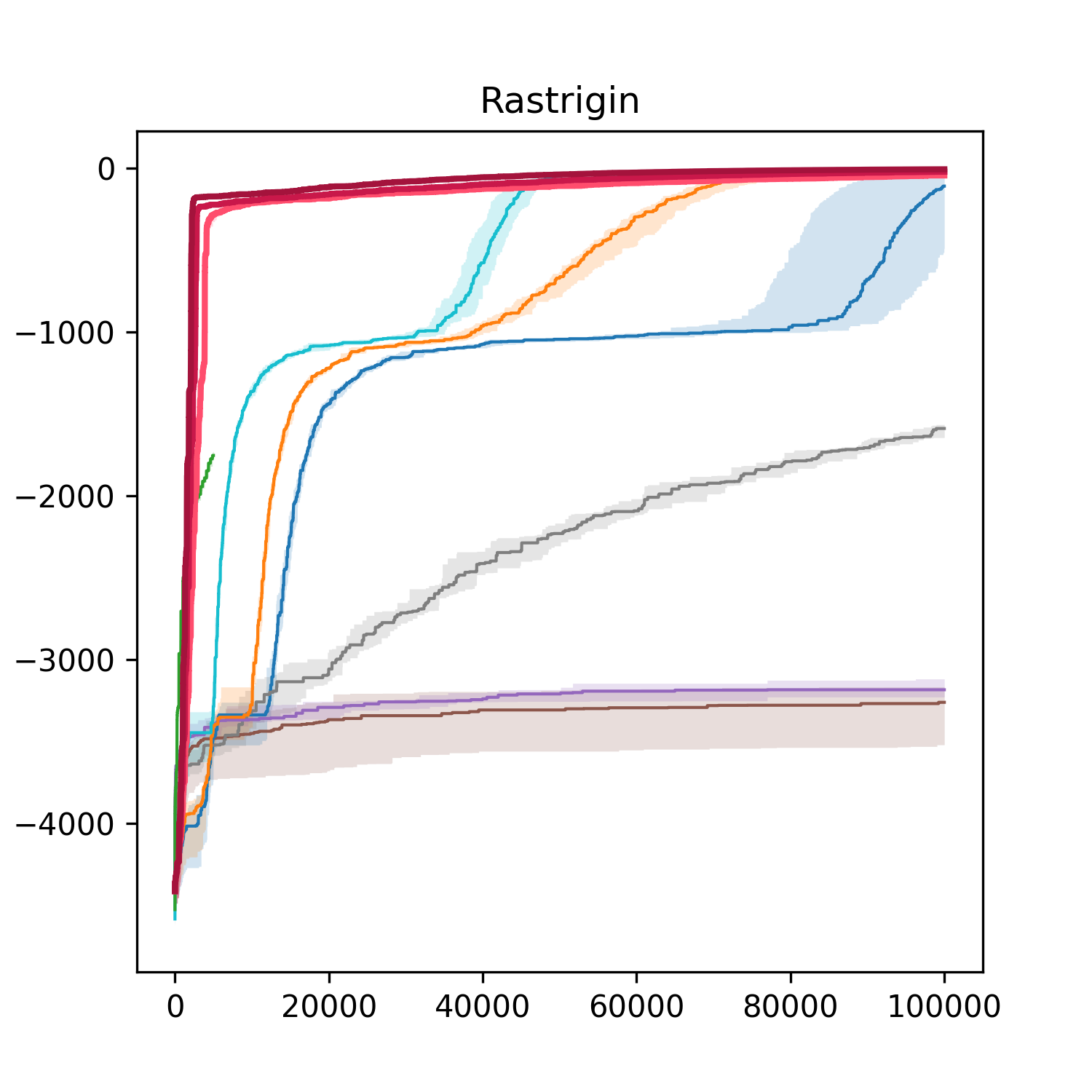}
    }
    \subfigure[Ackley]{
        \includegraphics[width = .25\linewidth]{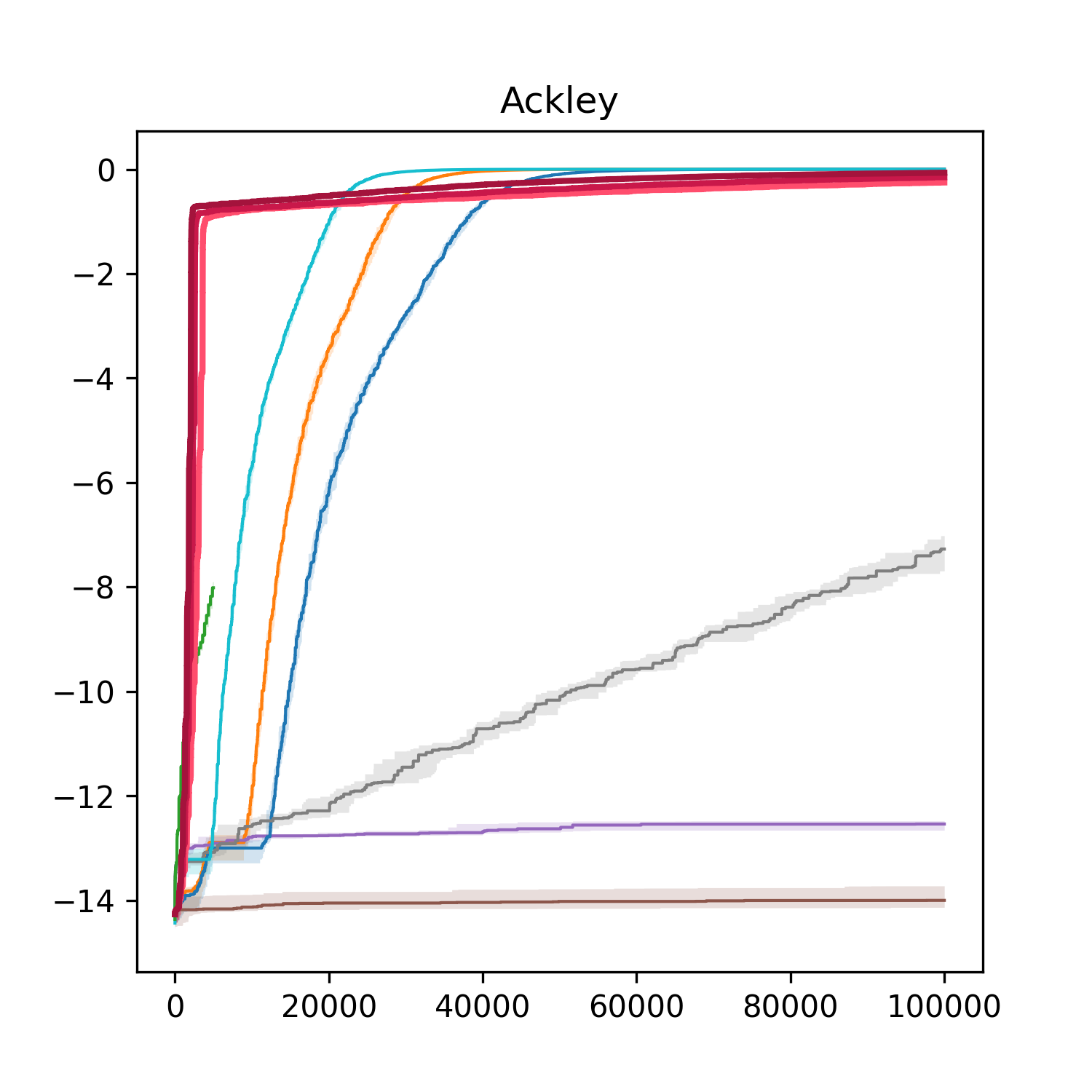}
    }
    \subfigure[Schaffer]{
        \includegraphics[width = .25\linewidth]{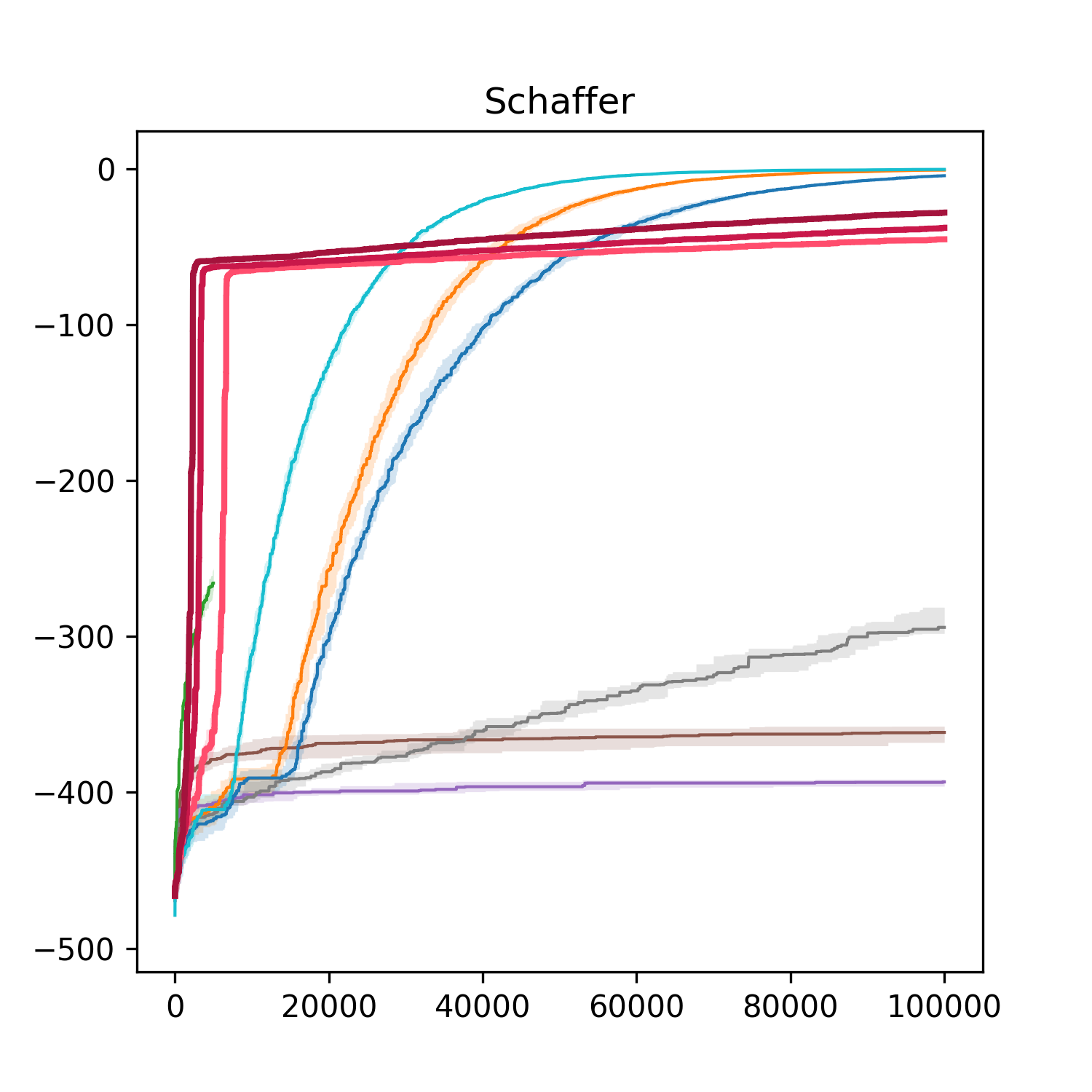}
    }
    \vfill
    \subfigure[Discus]{
        \includegraphics[width = .25\linewidth]{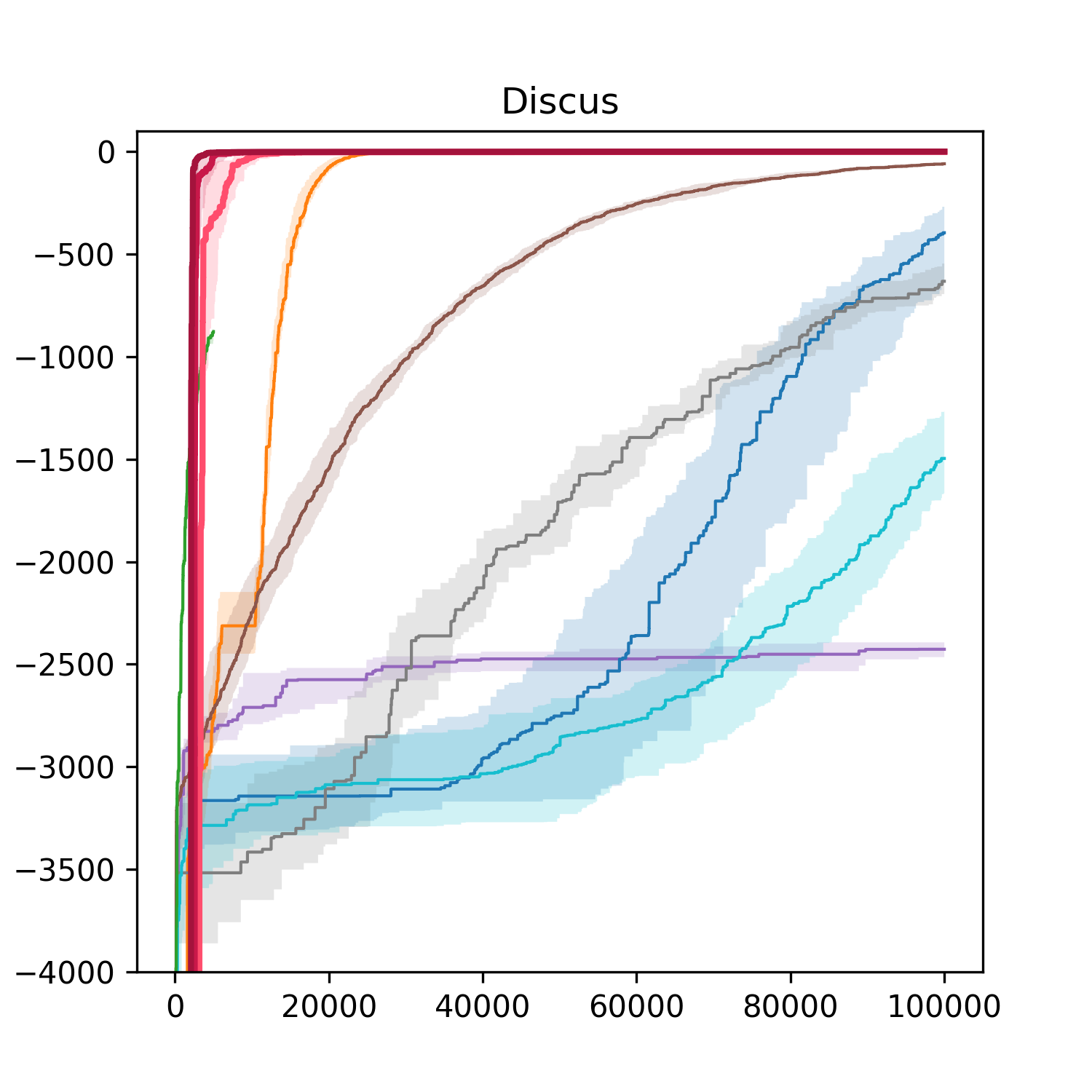}
    }
    \subfigure[Schwefel]{
        \includegraphics[width = .25\linewidth]{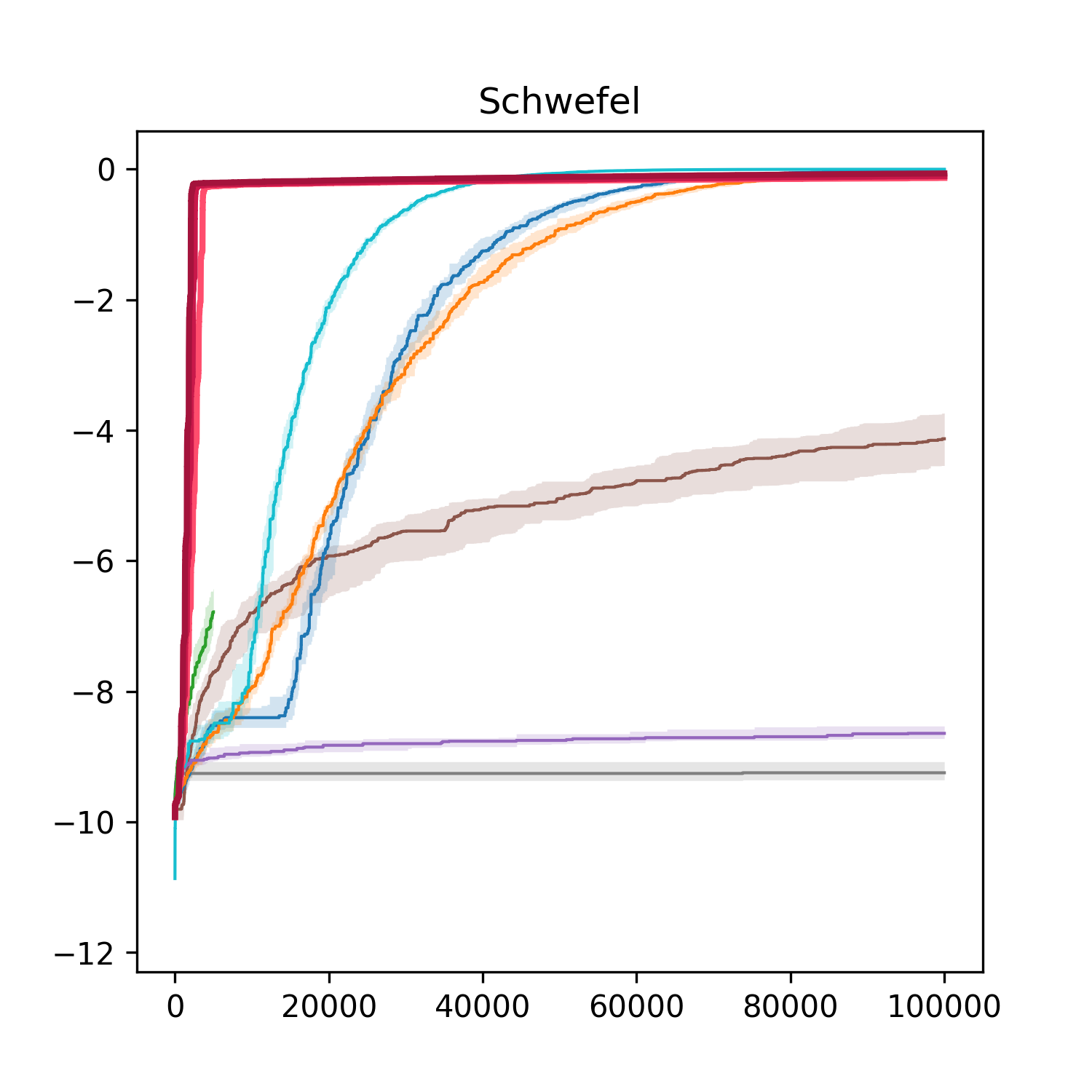}
    }
    \subfigure[DiffPowers]{
        \includegraphics[width = .25\linewidth]{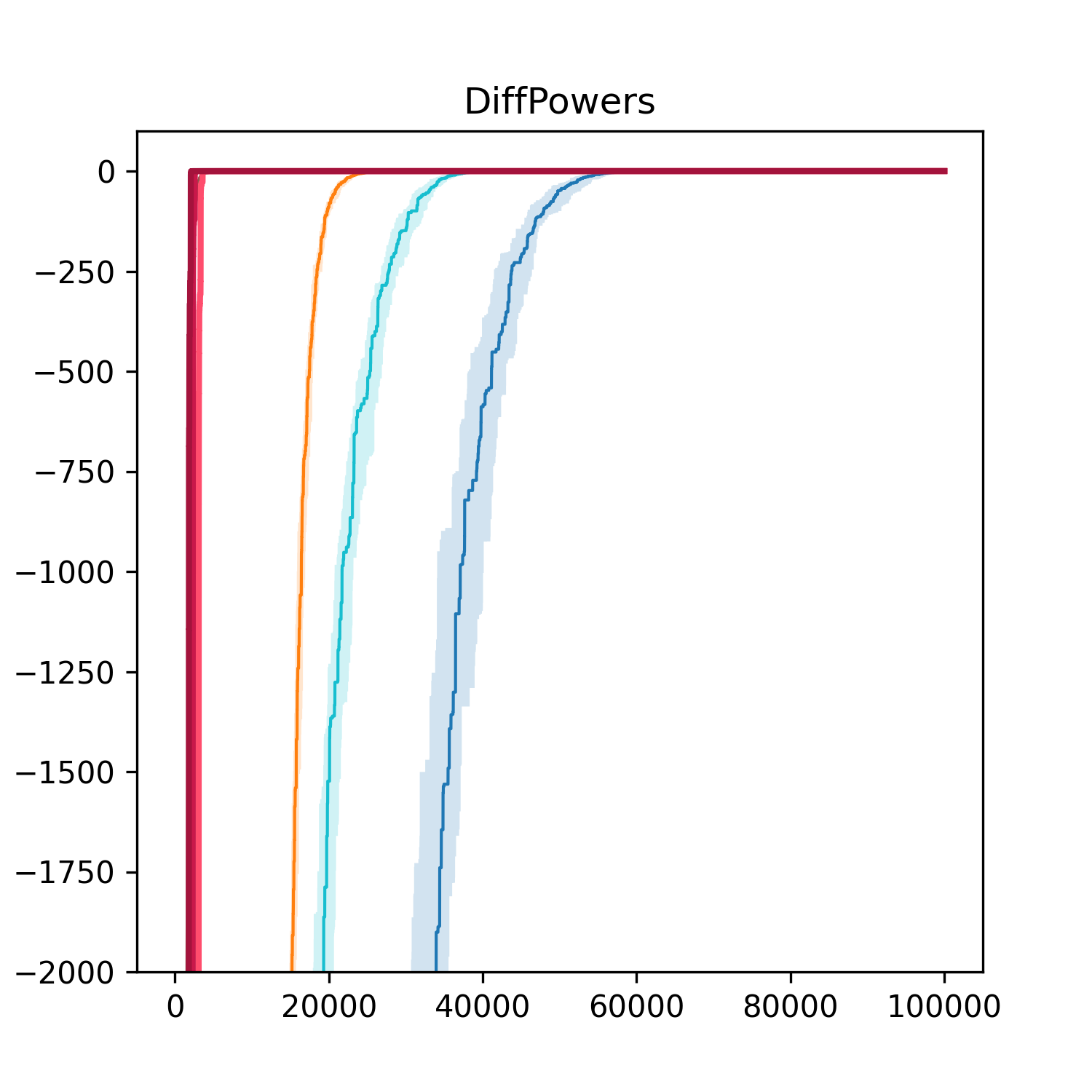}
    }
    \vfill
    \subfigure[Rosenbrock]{
        \includegraphics[width = .25\linewidth]{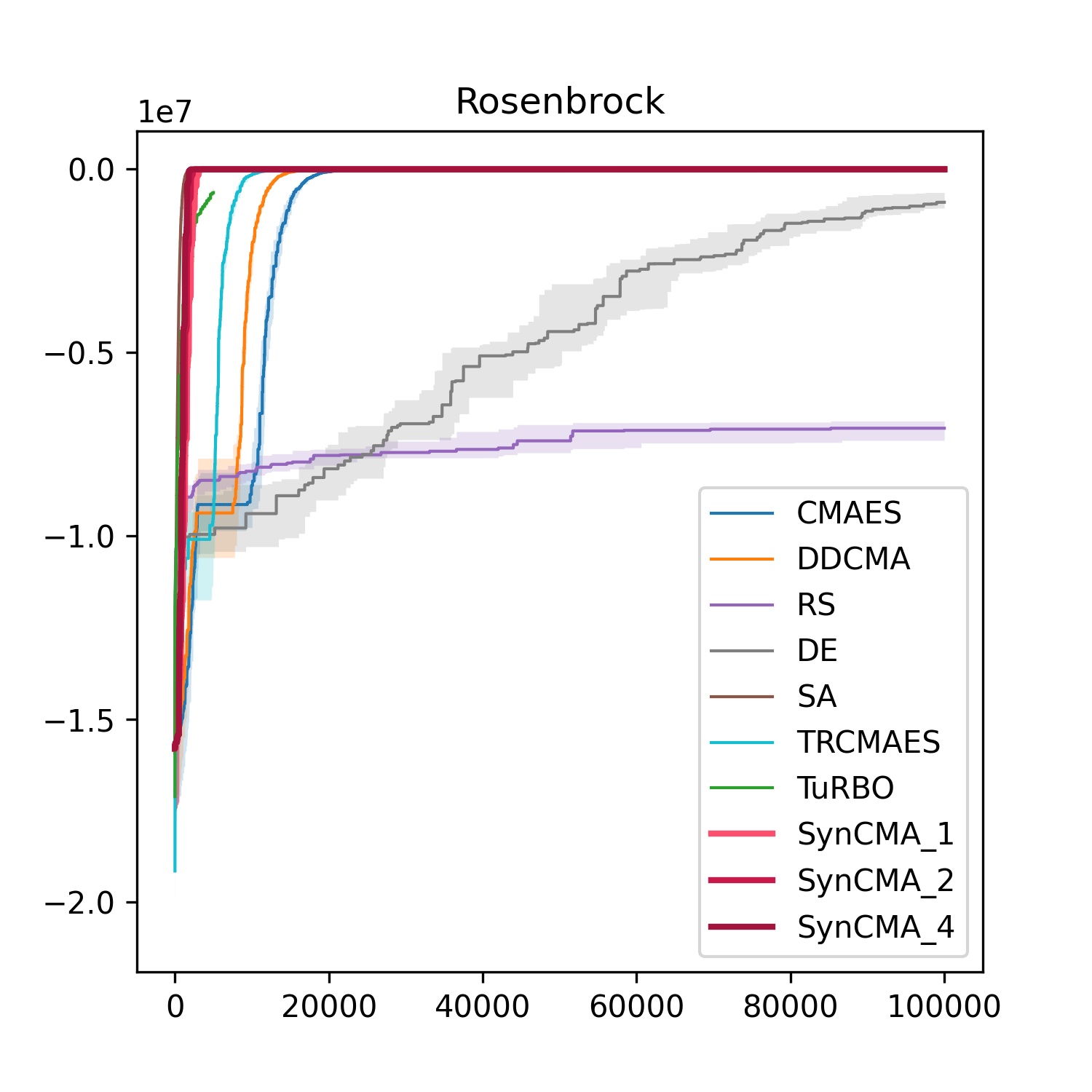}
    }
    
    \caption{Optimization procedure in 10 tests function with dimension $n = 128$ over 100 trails with 100000 evaluations.}
    \vspace{10pt}
    \label{fig:syn_128}
\end{figure}

For synthetic functions, we directly use the benchmarks provided by PyPop7, with formulation shown in Table \ref{tab:syn_form}. Full contents of near optimal performance and the average efficiency rank, i.e. the first time hitting time into 0.5 value, for CMA optimizers under the same budget are provided in tables \ref{tab:dim32} and \ref{tab:dim128}, other optimizers are excluded as their poor performance and linewidth limit. The optimization procedure of all 10 functions for \algo~with different $\lambda_0$ and other baselines are provided in figures \ref{fig:syn_32}, \ref{fig:syn_64}, \ref{fig:syn_128}.

It can be seen that as dimension increasing, the average performance of \algo~is getting worse. This align with the observation on dimensionality in ablation study that higher dimension generally yields higher $\lambda_0$. 

\begin{table}[!hbpt]
    \caption{10 different synthetic functions. Selected from classic test function for global optimization, including different rugged characteristics like multi-modality, high condition number and different optima landscape. The optima all scale to $x^* = 0, f(x^*) = 0$. For the brevity sake, $w_i = 1 + \frac{x_i + 1}{4}$ in LevyMontalvo function, $z_i = x_i^2 + x_{i + 1}^2$ in Schaffer function.}
    \vspace{1em}
    \centering
    \begin{tabular}[!hbpt]{cc}
        \toprule
        Name & Expression\\
        \midrule
        Sphere & $\sum_i x_i^2$\\
        Discus & $10^6x_0 + \sum_{i \ge 1} x_i$\\
        Schwefel & $\sum_i x_i + \prod_i x_i$\\
        DiffPowers & $\sum_{i} x^{2 + \nicefrac{4i}{n}}$\\
        Bohachevsky & $\sum_{i < n - 1} 0.7 + x_i^2 + 2 * x_{i+1}^2 - 0.3 \cos(3\pi x_i) - 0.4\cos(4\pi x_{i + 1})$\\
        LevyMontalvo & $\frac{\pi}{n}(10 \sin^2(\pi w_0) + (w_{n - 1} - 1)^2 + \sum_{i < n - 1} (w_i - 1)^2(1 + 10\sin^2(\pi w_{i + 1})))$\\
        Rastrigin & $10n + \sum_i x_i^2 - 10\cos(2\pi x_i)$\\
        Ackley & $-20\exp(-0.2(\frac{1}{n}\sum_i x_i^2)^{0.5}) - \exp(\frac{1}{n}\sum_i \cos(2\pi x_i)) + 20 + e$\\
        Schaffer & $\sum_{i < n - 1} z_i^{0.25} * (\sin^2(50 * z_i^{0.1}) + 1)$\\
        Rosenbrock & $100 \sum_{i \ge 1} (x_i - x_{i - 1}^2)^2 + \sum_{i < n - 1} (x_i - 1)^2$\\
        \bottomrule
    \end{tabular}
    \vspace{1em}
    
    \label{tab:syn_form}
\end{table}

\subsection{Ablation Study} \label{appendix_ablation}

All the data provided in appendix actually contains different version of \algo~with $\lambda_0 = \{0, 1, 2, 4\}$, so they naturally consist the ablation study. The three observations can be verified thereby. Through every experiment on each $\lambda_0$, we monitor the behavior of \algo~and do not find any degeneration happen.
\begin{table}[th]
\caption{Near optimal performance of CMA optimizers on 32d synthetic functions(lower is better). We show the median best performance of 10,000 evaluations over 100 trials. Numbers in brackets indicates the evaluation number need for algorithms to achieve better than 0.5.}
    \label{tab:dim32}
\resizebox{\textwidth}{!}{%
\begin{tabular}{cccccccccc}
        \toprule 
        Function & DD-CMA & CMA-ES & TR-CMA-ES & SynCMA\_1 & SynCMA\_2 & SynCMA\_4 \\
        \midrule
        Sphere & 0.0(3665) & 0.0(4612) & 0.1(2405) & 0.0(855) & 0.0(555) & 0.0(455) \\ 
        Discus & 0.0(5223) & 52.8 & 419.6 & 0.1(6363) & 0.0(4437) & 0.0(2568) \\ 
        Schwefel & 0.0(5325) & 0.0(5124) & 0.1(3074) & 0.0(704) & 0.0(519) & 0.0(449) \\ 
        DiffPowers & 0.0(3703) & 0.0(5585) & 0.1(3720) & 0.0(704) & 0.0(519) & 0.0(450) \\ 
        Bohachevsky & 0.0(5420) & 0.0(7115) & 0.4(4428) & 0.2(8163) & 0.1(6340) & 0.0(5106) \\ 
        LevyMontalvo & 0.0(5153) & 0.0(4030) & 0.1(2506) & 0.0(1350) & 0.0(1092) & 0.1(1094) \\ 
        Rastrigin & 58.1 & 194.8 & 15.9 & 1.0 & 0.2(8758) & 0.1(7372) \\ 
        Ackley & 0.0(4849) & 0.0(5850) & 0.1(2946) & 0.1(2733) & 0.0(2133) & 0.0(1134) \\ 
        Schaffer & 61.3 & 71.3 & 2.2 & 6.6 & 4.8 & 3.8 \\ 
        Rosenbrock & 28.0 & 29.9 & 0.1(7251) & 29.3 & 29.2 & 29.6 \\ 
        \midrule
        \textbf{Ave\_Rank} & 5 & 6 & 4 & 3 & 2 & 1\\
        \bottomrule
        
    \end{tabular}
  
}
\end{table}
    
\begin{table}[th]
\caption{Near optimal performance of CMA optimizers on 64d synthetic functions(lower is better). We show the median best performance of 50,000 evaluations over 20 trials. Numbers in brackets indicates the evaluation number need for algorithms to achieve better than 0.5. }
    \label{tab:dim64}
\resizebox{\textwidth}{!}{%
\begin{tabular}{cccccccccc}
        \toprule 
        Function & DD-CMA & CMA-ES & TR-CMA-ES & SynCMA\_1 & SynCMA\_2 & SynCMA\_4 \\
        \midrule
        Sphere & 0.0(10047) & 0.0(13825) & 0.0(7185) & 0.0(4691) & 0.0(3938) & 0.0(1047) \\ 
        Discus & 0.0(12748) & 0.0(43265) & 85.9 & 0.0(27662) & 0.0(18820) & 0.0(10484) \\ 
        Schwefel & 0.0(16820) & 0.0(16134) & 0.0(9905) & 0.0(1548) & 0.0(1157) & 0.0(929) \\ 
        DiffPowers & 0.0(10164) & 0.0(18503) & 0.0(12040) & 0.0(1548) & 0.0(1158) & 0.0(937) \\ 
        Bohachevsky & 0.0(14605) & 0.0(20369) & 0.0(11865) & 0.3(44088) & 0.0(32247) & 0.0(23314) \\ 
        LevyMontalvo & 0.0(9087) & 0.0(9901) & 0.0(5515) & 0.0(2522) & 0.0(2318) & 0.0(2486) \\ 
        Rastrigin & 17.9 & 21.4 & 22.4 & 1.3 & 0.2(42696) & 0.0(31565) \\ 
        Ackley & 0.0(11757) & 0.0(15620) & 0.0(7869) & 0.0(9800) & 0.0(7567) & 0.0(4549) \\ 
        Schaffer & 23.3 & 0.4(49400) & 0.3(32701) & 12.3 & 8.2 & 4.9 \\ 
        Rosenbrock & 53.8 & 57.8 & 0.0(21543) & 59.8 & 60.1 & 60.9 \\ 
        \midrule
        \textbf{Ave\_Rank} & 5 & 6 & 3 & 4 & 2 & 1\\
        \bottomrule
        
    \end{tabular}
  
  }
\end{table}

\begin{table}[th]
\caption{Near optimal performance of CMA optimizers on 128d synthetic functions(lower is better). We show the median best performance of 100,000 evaluations over 20 trials. Numbers in brackets indicates the evaluation number need for algorithms to achieve better than 0.5.}
    \label{tab:dim128}
\resizebox{\textwidth}{!}{%
\begin{tabular}{cccccccccc}
        \toprule 
        Function & DD-CMA & CMA-ES & TR-CMA-ES & SynCMA\_1 & SynCMA\_2 & SynCMA\_4 \\
        \midrule
        Sphere & 0.0(39745) & 0.0(27751) & 0.0(21428) & 0.2(54617) & 0.1(40403) & 0.0(25428) \\ 
        Discus & 395.8 & 0.0(32544) & 1496.4 & 1.0 & 0.4(93272) & 0.1(58425) \\ 
        Schwefel & 0.0(51681) & 0.0(59947) & 0.0(31407) & 0.1(3610) & 0.1(2600) & 0.1(2085) \\ 
        DiffPowers & 0.0(62675) & 0.0(28185) & 0.0(41019) & 0.0(3626) & 0.0(2621) & 0.0(2123) \\ 
        Bohachevsky & 0.0(57138) & 0.0(39401) & 0.0(31801) & 9.8 & 4.8 & 1.2 \\ 
        LevyMontalvo & 0.0(26924) & 0.0(20332) & 0.0(14875) & 0.0(5672) & 0.0(5404) & 0.0(5344) \\ 
        Rastrigin & 107.6 & 23.0 & 33.3 & 42.6 & 19.9 & 5.0 \\ 
        Ackley & 0.0(41488) & 0.0(29714) & 0.0(21994) & 0.3(46327) & 0.2(33839) & 0.1(20330) \\ 
        Schaffer & 4.3 & 0.7 & 0.2(91494) & 45.1 & 37.7 & 27.9 \\ 
        Rosenbrock & 124.3 & 120.5 & 0.0(68532) & 146.6 & 134.8 & 127.8 \\ 
        \midrule
        \textbf{Ave\_Rank} & 5 & 3 & 2 & 6 & 4 & 1\\ 
        \bottomrule
        
    \end{tabular}
  
  }
\end{table}

\end{document}